\documentclass{article}
\PassOptionsToPackage{table}{xcolor}
\usepackage{iclr2027_conference,times}
\usepackage[T1]{fontenc}
\usepackage[utf8]{inputenc}
\usepackage{xcolor}
\usepackage{amsmath,amssymb,amsfonts,mathtools,bm}
\usepackage{graphicx,booktabs,multirow,array,tabularx,adjustbox,wrapfig,placeins}
\usepackage{algorithm,algorithmic}
\usepackage{enumitem,makecell,xspace,microtype}
\usepackage{hyperref,url}
\usepackage{cleveref}
\makeatletter
\newcommand{\intablecaption}[1]{\def\@captype{table}\caption{#1}}
\newcommand{\intextcaption}[1]{\def\@captype{figure}\caption{#1}}
\makeatother
\newcolumntype{L}{>{\raggedright\arraybackslash}X}
\newcommand{\method}{\textsc{MS-Point}\xspace}

\newcommand{\loss}{\mathcal{L}}
\newcommand{\setL}{\mathcal{D}}

\newcommand{\setS}{\mathcal{S}}

\makeatletter
\newcommand{\finishwrap}{\par\ifnum\c@WF@wrappedlines>1\relax\vspace{\dimexpr\c@WF@wrappedlines\baselineskip-\baselineskip\relax}\fi\WFclear}
\makeatother
\newcommand{\wrapreserve}[1]{\finishwrap\begingroup\dimen0=\pagegoal\advance\dimen0 by-\pagetotal\ifdim\dimen0<#1\relax\newpage\fi\endgroup}
\iclrfinalcopy
\title{\Large MS-Point: Multi-Modal Hierarchical Spectral\\Learning for Point Cloud Understanding}
\author{Hai-tao Yu$^{1}$ \qquad Jun Xia$^{1,2,*}$\\[3pt]
\normalfont\small $^{1}$Information Hub, The Hong Kong University of Science and Technology (Guangzhou),\\
\normalfont\small Guangzhou, China\\
\normalfont\small $^{2}$The Hong Kong University of Science and Technology, Hong Kong SAR, China\\[2pt]
\normalfont\small \href{mailto:hyu382@connect.hkust-gz.edu.cn}{hyu382@connect.hkust-gz.edu.cn}\quad
\href{mailto:junxia@hkust-gz.edu.cn}{junxia@hkust-gz.edu.cn}\\
\normalfont\small $^{*}$Corresponding author}
\hypersetup{hidelinks,pdftitle={MS-Point: Multi-Modal Hierarchical Spectral Learning for Point Cloud Understanding},pdfauthor={Hai-tao Yu and Jun Xia}}
\makeatletter
\let\VnineCaption\caption
\let\VnineAtCaption\@caption
\let\VnineMakeCaption\@makecaption
\makeatother
\usepackage[font=small,labelfont=bf]{caption}
\definecolor{RevisionRedColor}{RGB}{0,0,0}
\DeclareRobustCommand{\RevisionRed}[1]{\color{RevisionRedColor}\relax#1\color{black}\relax}
\DeclareRobustCommand{\RevisionBlack}[1]{\color{black}\relax#1\color{black}\relax}
\newcommand{\UseAppendixRevisionStyle}{\renewcommand{\RevisionBlack}[1]{\leavevmode{\color{black}##1}}\renewcommand{\RevisionRed}[1]{\leavevmode{\color{RevisionRedColor}##1}}}

\begin{document}

\maketitle
\lhead{Preprint}

\begin{abstract}
\RevisionRed{Point--image alignment learns useful object representations, but different local feature patterns can produce the same pooled descriptor and receive the same alignment signal.} \RevisionRed{We introduce \method, a cross-modal pretraining framework that directly supervises intermediate point features and their neighborhood responses.} \RevisionRed{At selected encoder stages, we compute neighborhood averages and residuals before pooling and align their branch descriptors with image features.} \RevisionBlack{Branch-level objectives align each representation with its visual target, while cross-level consistency and adaptive fusion coordinate information across branches.} A common interface of intermediate features and spatial neighborhoods supports both graph encoders and point Transformers. Evaluations across object classification, few-shot adaptation, dense prediction, and low-label scene learning demonstrate the transfer value of this design. \RevisionBlack{Same-backbone comparisons} and capacity- and computation-matched controls connect these improvements to the structure of supervision. \RevisionBlack{Together, our findings support directly supervising intermediate features and neighborhood variation to learn transferable, geometry-aware representations across modalities.}
\end{abstract}

\begingroup
\section{Introduction}
\label{sec:intro}
A point cloud and an image observe the same object through different sensing processes. \RevisionBlack{Matching their representations provides supervision for a 3D encoder that must transfer to tasks with limited labels.} CrossPoint learns this agreement through object-level contrast; MM-Point enriches it with multiple views and projection spaces~\citep{afham2022crosspoint,yu2024mmpoint}. \RevisionBlack{How directly does object-level alignment constrain the features of neighboring points?}

With object-level alignment, the visual target is assigned to the final descriptor. \RevisionRed{Gradients reach the intermediate layers through this descriptor, so their features receive visual supervision only after pooling.} Different arrangements of neighboring features can yield the same pooled descriptor. Any projection of that descriptor then gives the same output for both arrangements. This motivates computing neighborhood responses before pooling and giving each branch its own visual target.

The distinction matters most when the downstream task changes. \RevisionRed{Recognizing a chair and separating its legs from its seat require different features. Part segmentation must retain distinctions between nearby regions even when they yield the same object-level descriptor.} \RevisionBlack{We therefore seek intermediate features that remain useful when the downstream task requires more than object recognition.} 

\RevisionBlack{A point encoder already computes features at several depths. We use these features as additional inputs to cross-modal learning. At selected depths, we also compute the neighborhood average and each point's deviation from it. \RevisionRed{Pooling and aligning these responses in separate branches lets image supervision act on local feature patterns as well as the final object descriptor.}}

\begin{figure}[!tbp]
\centering
\begin{minipage}{1.0\linewidth}
\centering
\includegraphics[width=\linewidth]{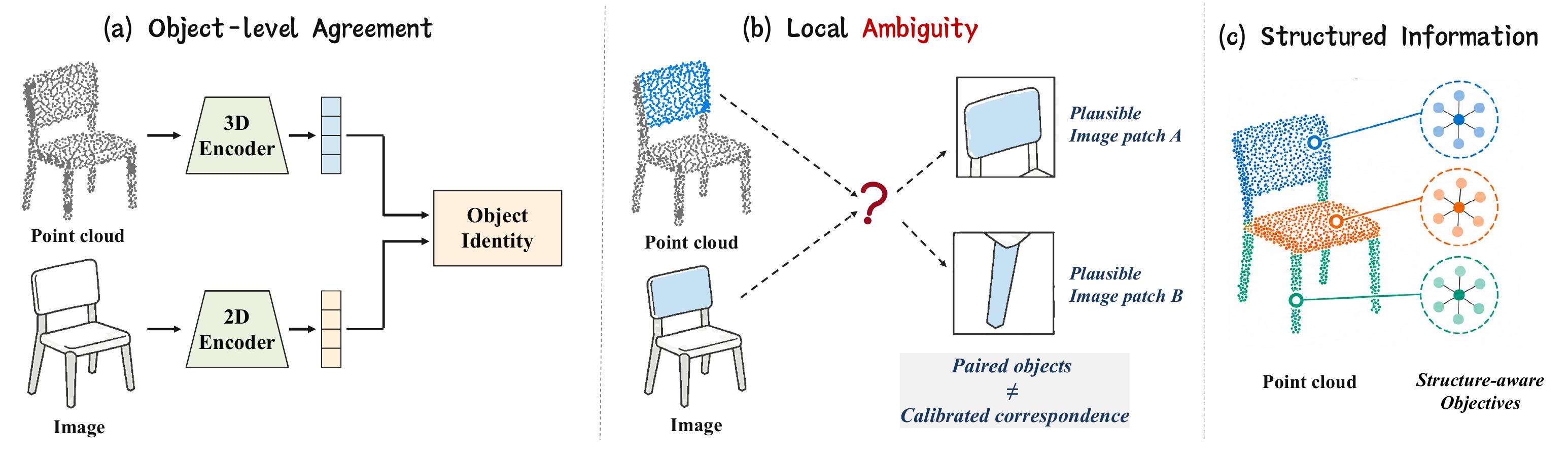}
\caption{\textbf{From object agreement to structural supervision.} Object identity leaves ambiguous local matches; intermediate states and neighborhood responses expose additional structure before pooling.}
\label{fig:intro}
\end{minipage}
\end{figure}

\RevisionBlack{We introduce \method to supervise selected encoder stages and the neighborhood variation within their features.} \RevisionBlack{Selected encoder states receive direct visual targets, while complementary neighborhood-consensus and deviation responses describe the neighborhood average and the residual from it. \RevisionRed{Encoder depth determines where we apply supervision, while graph frequency describes how features vary across neighboring nodes within a stage.}} \RevisionBlack{Semantic-depth matching pairs these branches with image features}; adaptive fusion learns their contribution to the object representation. The same interface operates on graph features and Transformer tokens without a reconstruction decoder.

\RevisionBlack{This formulation gives intermediate features and neighborhood responses their own visual targets, in addition to supervising the final object representation.} Hierarchical masking, image-guided reconstruction, and local distillation provide established routes to structure-aware learning~\citep{zhang2022pointm2ae,zhang2023i2p,sautier2022slidr}. \RevisionBlack{MS-Point aligns intermediate point features and their neighborhood responses with image features through branch-specific contrastive losses.} A complete object and a single image share only partial information, so visibility-aware local alignment further extends this principle to dense prediction.

Our contributions are:
\begin{itemize}[leftmargin=1.4em,itemsep=2pt,topsep=2pt]
\item \RevisionRed{A cross-modal objective that supervises intermediate point features and their neighborhood responses through separate visual targets.}
\item Complementary graph responses, semantic-depth targets, and joint branch/fusion objectives that transfer across graph encoders and point Transformers.
\item Evidence from classification, few-shot adaptation, segmentation, and low-label scenes, supported by same-backbone and capacity- and computation-matched interventions.
\end{itemize}

\par\endgroup
\begingroup
\section{Related Work}
\label{sec:related}
\paragraph{Hierarchical representations and prediction targets.}
PointNet++ aggregates nested neighborhoods, while DGCNN constructs dynamic feature-space relations~\citep{qi2017pointnetplusplus,wang2019dynamic}. Thus, locality and hierarchy are already present in many point encoders. Point-BERT and Point-MAE introduce masked point-token learning, and Point-M2AE extends it across scales~\citep{yu2022pointbert,pang2022masked,zhang2022pointm2ae}. These methods make clear that the prediction target, not only the backbone, shapes the learned representation.

\paragraph{Beyond a single reconstruction objective.}
PointGPT studies autoregressive point-patch generation~\citep{chen2023pointgpt}; PointDif uses diffusion-based denoising~\citep{zheng2024pointdif}. PCP-MAE examines the shortcut introduced by supplying masked patch centers to a decoder and instead predicts centers~\citep{zhang2024pcpmae}. \RevisionBlack{This connects to our motivation: the pretraining objective determines which features receive direct supervision.} TPM expands masked learning through multiple masks~\citep{liu2025tpm}, while Point-PQAE uses cross reconstruction between decoupled views~\citep{zhang2025pointpqae}.

\paragraph{Point--image supervision.}
CrossPoint and MM-Point provide the closest object-level contrastive references~\citep{afham2022crosspoint,yu2024mmpoint}. I2P-MAE and Inter-MAE exploit images within masked learning, and ReCon combines generative and contrastive signals~\citep{zhang2023i2p,liu2024intermae,qi2023recon}. \RevisionBlack{These methods establish visual guidance and multi-level learning as useful approaches to 3D pretraining.} \RevisionRed{MS-Point adds separate visual alignment objectives for intermediate point features and for neighborhood responses computed before pooling.}

OpenShape and ULIP-2 extend multimodal 3D learning through larger paired collections and richer visual supervision~\citep{liu2023openshape,xue2024ulip2}. We study a complementary question under a fixed view budget: how to use the structure already present inside the point encoder.

\par\endgroup
\begingroup
\section{Method}
\label{sec:method}
\subsection{\RevisionBlack{Intermediate Features and Branch Descriptors}}
\label{sec:overview}\label{sec:3dhier}
\RevisionBlack{Given a point cloud \RevisionRed{$X_{\rm cloud}$} and its paired image \RevisionRed{$I_{\rm img}$}, we learn an encoder whose intermediate states support transfer beyond object matching.} \RevisionRed{Let $H^{(l)}\in\mathbb R^{N_l\times C_l}$ denote a spatial feature map with $N_l$ nodes and $C_l$ channels at depth $l$, and let $\Phi$ be pooling and $q$ a projection head.} A global contrastive target observes $q(\Phi(H^{(l)}))$. \RevisionBlack{Different spatial arrangements of point features can produce the same pooled descriptor:}
\begin{equation}
H\ne H',\quad\Phi(H)=\Phi(H')\quad\Longrightarrow\quad q(\Phi(H))=q(\Phi(H')).
\label{eq:v10_observability}
\end{equation}
\RevisionBlack{The target cannot distinguish these organizations through this readout, although its loss still updates the intermediate layers.} \RevisionRed{MS-Point extracts intermediate features and computes their neighborhood responses before pooling, then aligns the resulting branch descriptors with image features} (Figure~\ref{fig:framework}).

\begin{figure}[!tbp]
\centering
\begin{minipage}{1.0\linewidth}
\centering
\includegraphics[width=\linewidth]{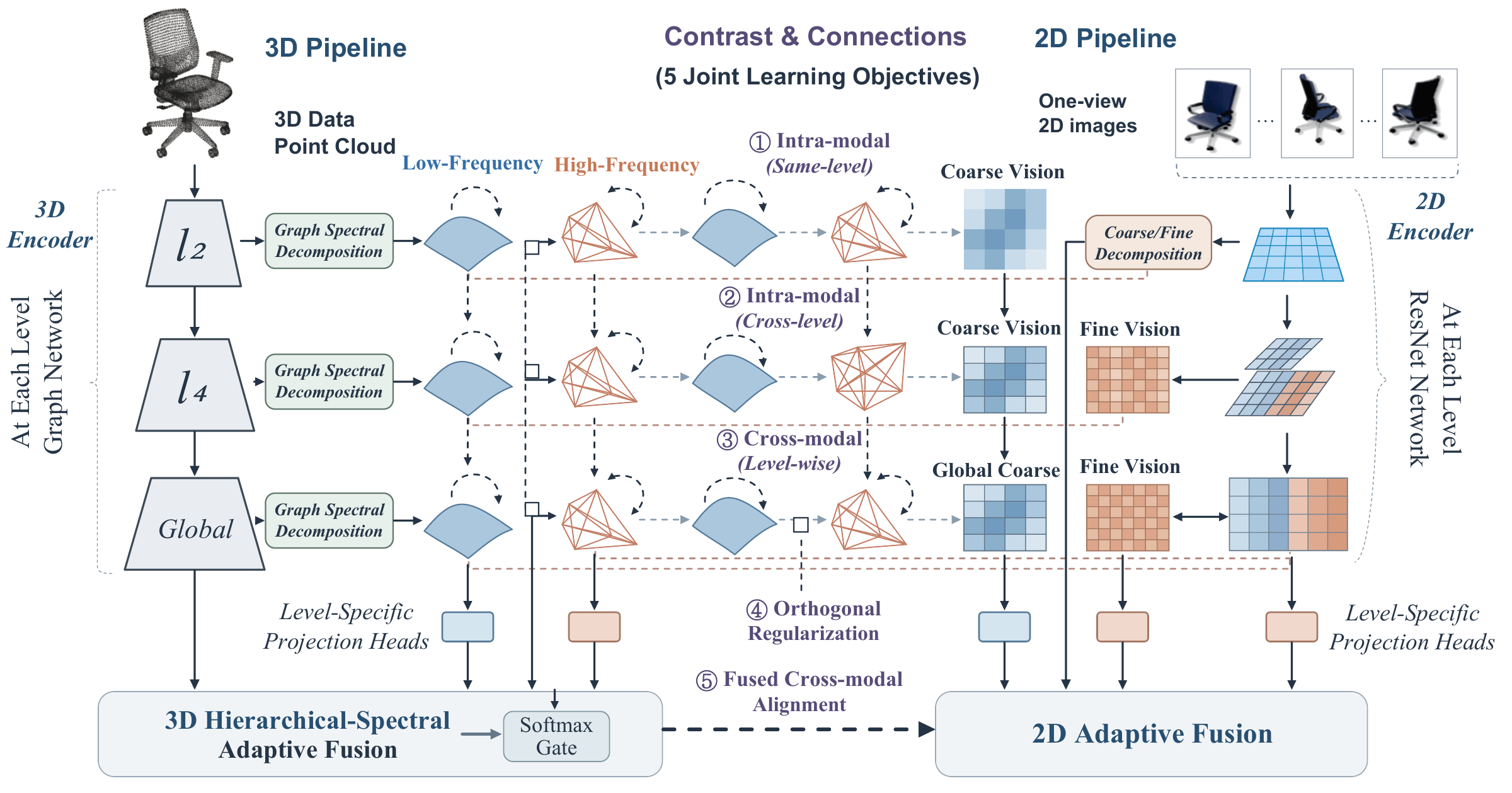}
\caption{\textbf{Structured point--image pretraining.} Intermediate states and complementary responses receive visual targets before adaptive fusion. First-order graph filters extract the neighborhood mean and residual. The diversity loss penalizes squared cosine similarity between their projected features.}
\label{fig:framework}
\end{minipage}
\vspace{-3pt}
\end{figure}

We describe graph and attention encoders through a common hierarchy $\{H^{(l)}\}_{l\in\mathcal D}$. \RevisionBlack{Features at different depths aggregate neighborhood information through different numbers of encoder blocks.} DGCNN provides evolving point-feature maps~\citep{wang2019dynamic}; point Transformers provide spatially indexed patch tokens~\citep{yu2022pointbert,pang2022masked,zhang2022pointm2ae}. Feature depth defines the hierarchy in both cases, while staged token pooling also changes spatial resolution. We read the global-stage map before pooling, retaining its spatial support for response construction.

\RevisionRed{We concatenate the channel-wise maximum and mean across points or tokens to obtain each branch descriptor:}
\begin{equation}
\Phi(H)=\big[\max_i H_i,\ N^{-1}\textstyle\sum_i H_i\big],\qquad
z_m=q_m\!\left(\Phi(H_m)\right).
\label{eq:v11_pool}
\end{equation}
\RevisionBlack{Here $m$ indexes a branch, $H_m$ is its feature map, and $N$ is its number of nodes. The map $q_m$ places the descriptor in a common contrastive space.} \RevisionBlack{Each selected stage receives its own visual target (Figure~\ref{fig:method}). A shared backbone forward pass provides the features for all branches.}

\begin{figure}[!tbp]
\centering
\begin{minipage}{.76\linewidth}
\centering
\includegraphics[width=\linewidth]{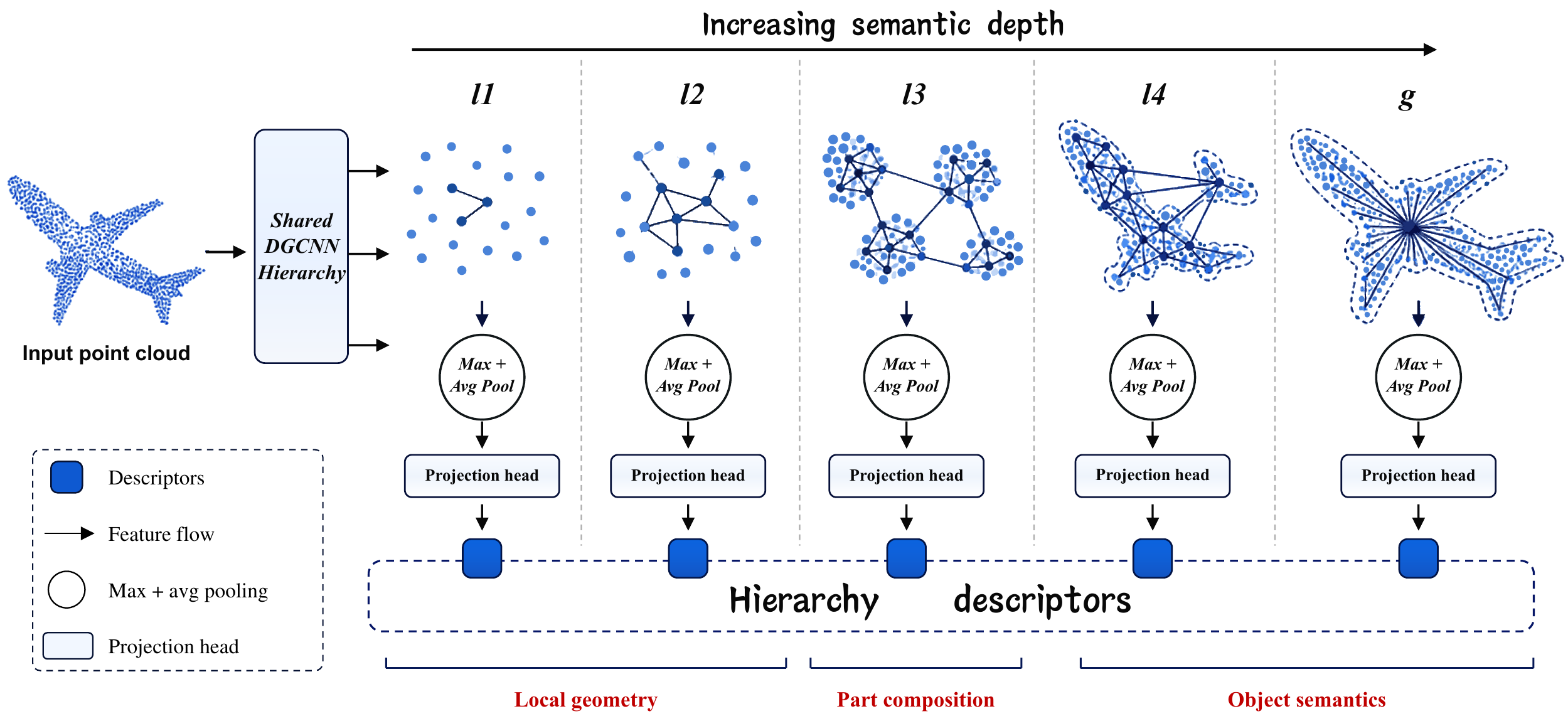}
\caption{\textbf{Supervising the feature hierarchy.} \RevisionBlack{Features from selected stages are pooled and projected through separate heads. The global-stage feature map is extracted before pooling.}\RevisionRed{ The blue squares denote the projected stage descriptors.}}
\label{fig:method}
\end{minipage}
\vspace{-3pt}
\end{figure}

\subsection{Complementary Responses on Point and Token Graphs}
\label{sec:spectral}\label{sec:transformer_main}
\RevisionRed{For a response source $l\in\mathcal S\subseteq\mathcal D$, let $W_{ij}^{(l)}\geq0$ be the weight assigned to node $j$ when aggregating features for node $i$. Define $P^{(l)}=D^{-1}W^{(l)}$, with $D_{ii}=\sum_jW_{ij}^{(l)}>0$.} We compute
\begin{equation}
H_{\rm s}^{(l)}=P^{(l)}H^{(l)},\qquad H_{\rm d}^{(l)}=(I-P^{(l)})H^{(l)},\qquad
H_{\rm s}^{(l)}+H_{\rm d}^{(l)}=H^{(l)}.
\label{eq:v10_responses}
\end{equation}
Graph features use their feature-space neighborhood relation. For MS-Point-T, token centers $c_i^{(l)}\in\mathbb R^3$ determine a geometric graph independently of attention:
\begin{equation}
P_{ij}^{(l)}=\frac{\mathbf1\{j\in\mathcal N_k(c_i^{(l)})\}}{|\mathcal N_k(c_i^{(l)})|},\qquad
\mathcal A=(\mathcal D\times\{\mathrm{base}\})\cup(\mathcal S\times\{\mathrm{s},\mathrm{d}\}).
\label{eq:v11_token_graph}
\end{equation}
This interface leaves the backbone operations unchanged. Fixed-resolution Transformers reuse token centers across selected blocks; hierarchical Transformers use the centers associated with each stage.

\begin{wrapfigure}{R}{.44\textwidth}
\vspace{-3pt}
\centering
\includegraphics[width=\linewidth]{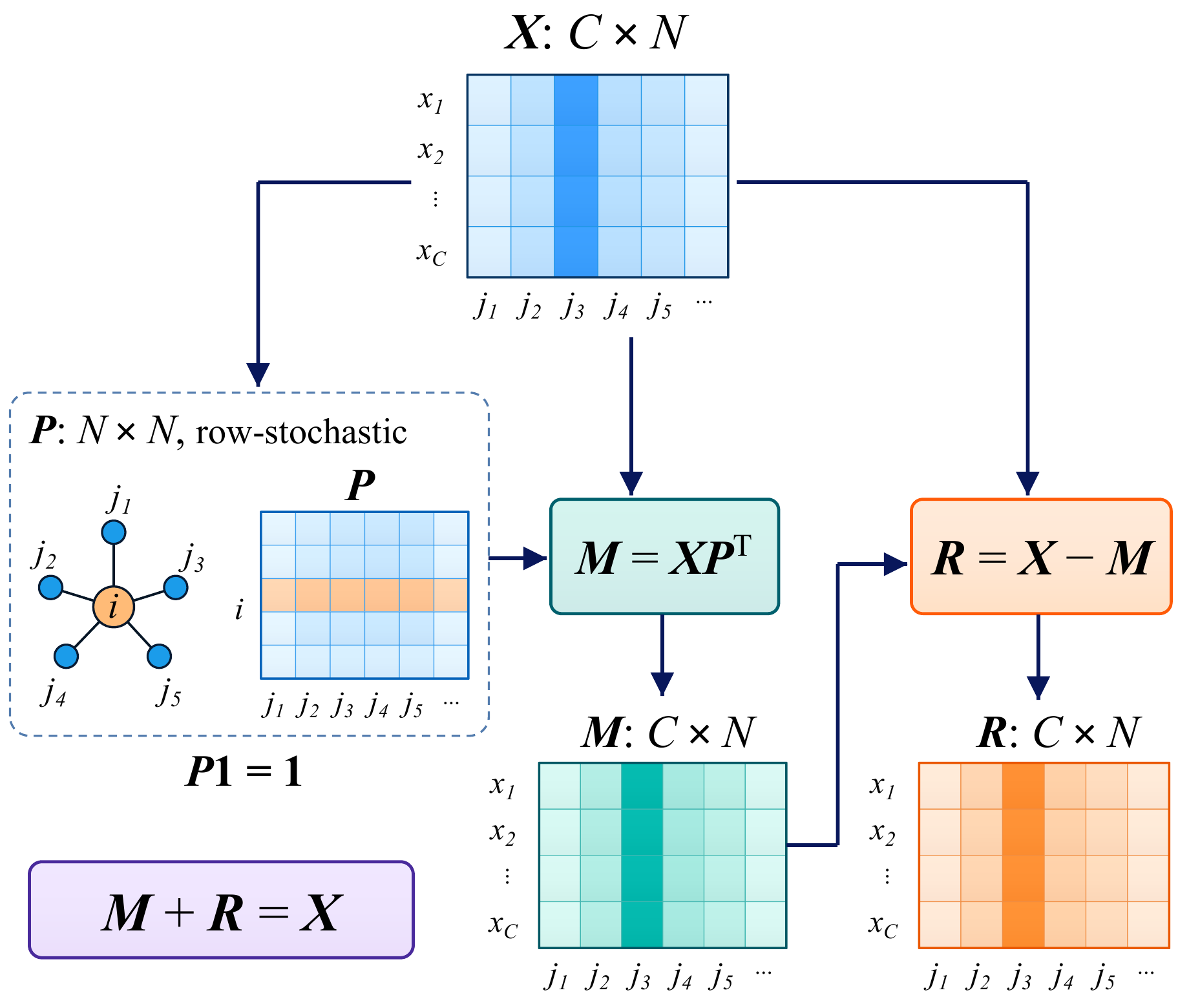}
\caption{\textbf{Responses before pooling.} In channel-first notation, $M=XP^\top$ and $R=X-M$ satisfy $M+R=X$.}
\label{fig:response_construction}
\vspace{-3pt}
\end{wrapfigure}

\RevisionRed{The consensus branch averages neighboring features, providing a local structural reference.} \RevisionRed{The deviation branch records the residual from this average, highlighting points whose features differ sharply from their neighbors.} \RevisionBlack{Both responses are computed from the same feature map and neighborhood graph, then pooled in separate branches} (Figure~\ref{fig:response_construction}). Independent heads then allow visual supervision to select useful neighborhood agreement and local contrast separately.

\RevisionBlack{\RevisionRed{For a symmetric reference graph}, normalized adjacency $S$ and Laplacian $L=I-S=U\Lambda U^\top$, with orthonormal eigenvectors $U$ and eigenvalues in $\Lambda$, give responses $SH=U(I-\Lambda)U^\top H$ and $LH=U\Lambda U^\top H$.} \RevisionBlack{They are first-order polynomial graph filters~\citep{shuman2013emerging,defferrard2016convolutional}; deviation gives greater gain to modes of larger graph variation. \RevisionRed{On $[0,1]$, the consensus gain decreases from one to zero as the deviation gain increases from zero to one.}} The directed implementation retains the complementary graph-shift pair, while the symmetric reference defines its spectral interpretation (Appendix~\ref{app:further_theory}).

\RevisionBlack{Neighborhood responses incorporate graph--feature relationships into the branch descriptors before pooling.} For a permutation $\Pi$ that preserves pooled channel statistics, $\Phi(\Pi H)=\Phi(H)$, whereas a fixed neighborhood relation generally gives $P\Pi H\ne\Pi PH$. \RevisionRed{For a fixed graph, pooling these responses can distinguish arrangements with identical base descriptors.} A joint relabeling of features and graph remains equivariant, preserving the point-set symmetry. This separates sensitivity to the relation between features and their neighbors from sensitivity to arbitrary point ordering (Appendix~\ref{sec:interface_analysis}).

The raw mean and residual sum to the original stage features. Separate adaptors and visual targets allow the branches to learn different features from these complementary inputs. Table~\ref{tab:main_controls} compares the two responses with duplicated features at matched capacity.

\subsection{\RevisionBlack{Image Targets and Adaptive Fusion}}
\label{sec:2d}\label{sec:fusion}
\RevisionBlack{The image encoder provides features from multiple stages as visual targets.} \RevisionBlack{A stage feature $F_j$ supplies a base descriptor and spatial-average/residual counterparts, with $\Gamma_{\rm base}(F)=F$, $\Gamma_{\rm s}(F)=\mathcal B(F)$ and $\Gamma_{\rm d}(F)=F-\mathcal B(F)$.} For a point branch $m=(l,b)$, the depth map $\rho(m)=(r(l),b)$ selects its visual target:
\begin{equation}
h_{\rho(m)}=q_m^{\rm img}\!\left(\Phi_{\rm img}(\Gamma_b(F_{r(l)}))\right),\qquad
u_f=\sum_{m\in\mathcal A}\alpha_m z_m,\quad z_f=q_f(u_f).
\label{eq:v11_targets_fusion}
\end{equation}
\RevisionBlack{Here $\mathcal B$ is spatial averaging, and $\Phi_{\rm img}$ aggregates spatial features into a descriptor.} Gates predict normalized nonnegative scores $\alpha_m$ from the branch embeddings, with $\sum_m\alpha_m=1$; image embeddings are fused separately to obtain $h_f$.

\begin{wrapfigure}{R}{.40\textwidth}
\vspace{-3pt}
\centering
\includegraphics[width=\linewidth]{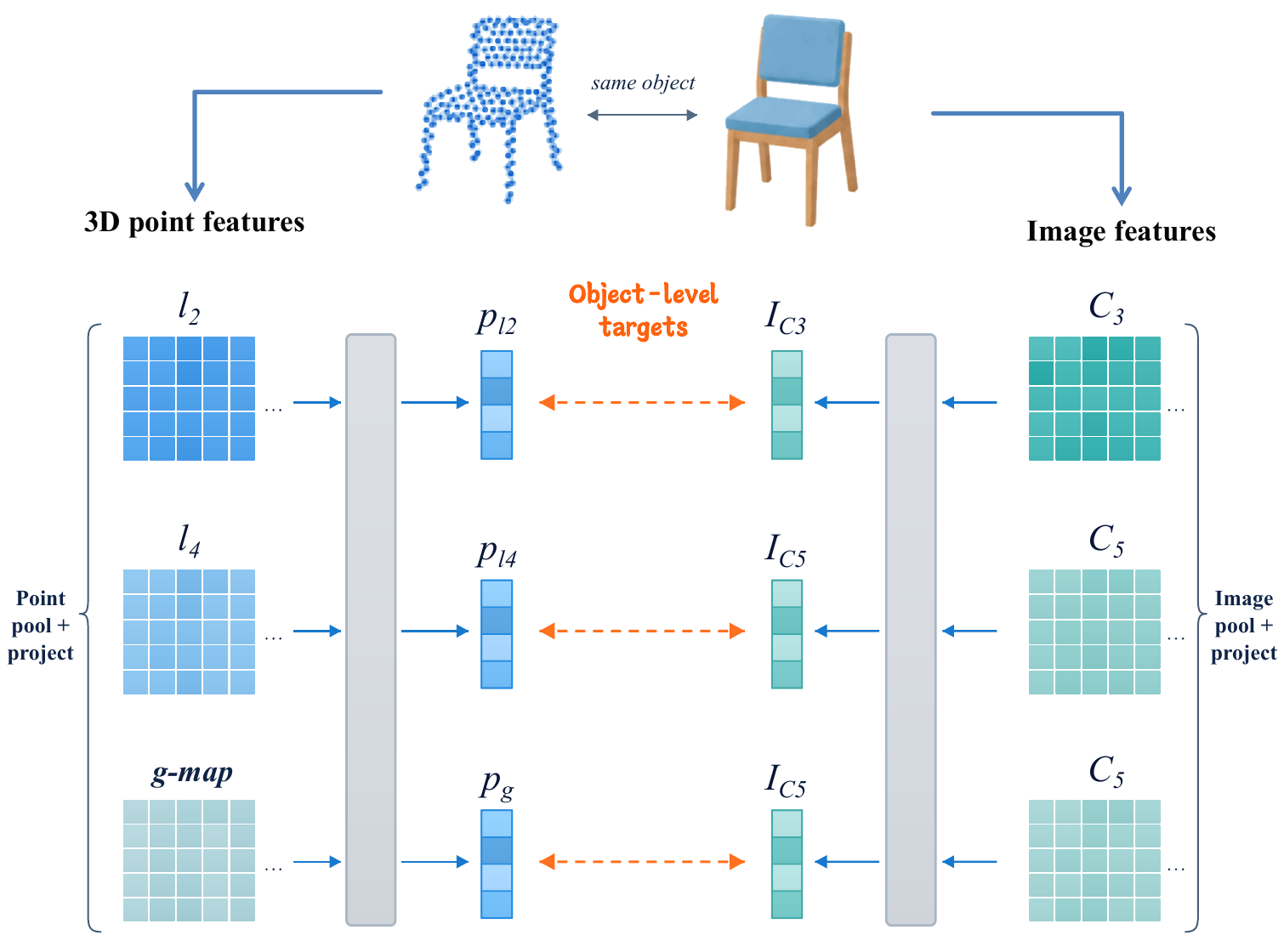}
\caption{\textbf{Semantic-depth targets.} Branch-specific projections pair intermediate and deep point states with visual anchors.}
\label{fig:stage_targets}
\vspace{-3pt}
\end{wrapfigure}

\RevisionBlack{We pair point and image stages by their semantic depth (Figure~\ref{fig:stage_targets}).} Shallower point states draw on intermediate image features, while deeper states use more abstract targets. ResNet, ConvNeXt, and spatial ViT features can supply these anchors through the same interface. Independent nonlinear heads accommodate different feature widths and let branches sharing an image source specialize in their projected objectives.

Object identity supplies the positive cross-modal relation without requiring point--pixel registration. \RevisionRed{The target map specifies which image feature supervises each branch. The fusion gates then predict branch weights for each object.} The fused objective trains their collective readout, and branch objectives retain direct supervision of every selected structural state. Full stage maps, dimensions, graph conventions, and implementation variants are specified in Appendix~\ref{app:implementation_variants}.

\subsection{Joint Consistency, Alignment, and Diversity}
\label{sec:objective}
Two point augmentations yield embeddings $z_m^{(1)},z_m^{(2)}$; the paired image supplies $h_{\rho(m)}$. Let $\ell$ denote NT-Xent~\citep{chen2020simple}, $\mathcal E$ the selected cross-level pairs, and $\bar z_m=(z_m^{(1)}+z_m^{(2)})/2$. For paired batches $a,b$ of size $B$, its directional term is
\begin{equation}
\begin{gathered}\ell_{a\to b}=-\frac1B\sum_{i=1}^{B}\log\frac{\exp(s(a_i,b_i)/\tau)}{\sum_{j\ne i}\exp(s(a_i,a_j)/\tau)+\sum_{j=1}^{B}\exp(s(a_i,b_j)/\tau)},\\
\ell(a,b)=\tfrac12(\ell_{a\to b}+\ell_{b\to a})\end{gathered},
\label{eq:v11_ntxent}
\end{equation}
where $s$ is cosine similarity and $\tau$ is temperature. Corresponding observations of the same object are positives; the remaining batch instances provide negatives. The branch map changes which representation enters this objective, while the instance discrimination rule stays shared. We define $\bar z_f$ by the same view average. Branch stability and hierarchy coordination are
\begin{align}
\mathcal L_{\rm same}&=|\mathcal A|^{-1}\sum_{m\in\mathcal A}\ell(z_m^{(1)},z_m^{(2)}),\label{eq:v11_same}\\
\mathcal L_{\rm cross}&=(2|\mathcal E|)^{-1}\sum_{(m,n)\in\mathcal E}\sum_{v=1}^{2}\ell(z_m^{(v)},z_n^{(v)}).\label{eq:v11_cross}
\end{align}
\begin{table}[!t]
\begin{minipage}[t]{0.485\linewidth}
\centering\footnotesize
\setlength{\tabcolsep}{3pt}
\caption{\RevisionRed{ModelNet40/10 accuracy (\%). $^{*}$ denotes our controlled linear-SVM evaluation with 1,024 input points and no voting. A dash indicates an unreported result.}}
\label{tab:modelnet}
\label{tab:completed_baselines}
\begin{tabularx}{\linewidth}{@{\hspace{1.5pt}}Lrr@{}}
\toprule
Method & MN40 & MN10\\
\midrule
\multicolumn{3}{l}{\emph{Supervised references}}\\
PointNet \citep{qi2017pointnet} & 89.2 & --\\
\leavevmode\RevisionRed{GPSFormer \citep{wang2024gpsformer}} & \RevisionRed{93.8} & --\\
\leavevmode\RevisionRed{PCM \citep{zhang2025pcm}} & \RevisionRed{93.4} & --\\
\midrule
\multicolumn{3}{l}{\emph{Frozen-feature linear evaluation}}\\
SwAV \citep{caron2020unsupervised} & 90.3 & 93.5\\
OcCo \citep{wang2021unsupervised} & 89.2 & 92.7\\
STRL \citep{liu2021strl} & 90.9 & --\\
Point-BERT \citep{yu2022pointbert} & 87.4 & --\\
Point-MAE \citep{pang2022masked} & 91.0 & --\\
Point-MAE$^{*}$ & 91.4 & --\\
Point-M2AE \citep{zhang2022pointm2ae} & 92.9 & --\\
CrossPoint \citep{afham2022crosspoint} & 91.2 & --\\
I2P-MAE \citep{zhang2023i2p} & 93.4 & --\\
Inter-MAE$^{*}$ \citep{liu2024intermae} & 92.1 & --\\
MM-Point \citep{yu2024mmpoint} & 92.4 & 95.4\\
MM-Point$^{*}$ & 92.5 & --\\
Point-MAE + TPM$^{*}$ \citep{liu2025tpm} & 91.8 & --\\
MS-Point (DGCNN)$^{*}$ & 93.6 & 96.5\\
\rowcolor{gray!15} MS-Point-T$^{*}$ & \textbf{94.3} & \textbf{97.2}\\
\bottomrule
\end{tabularx}
\end{minipage}\hfill
\begin{minipage}[t]{0.485\linewidth}
\centering\footnotesize
\setlength{\tabcolsep}{3pt}
\caption{\RevisionRed{ScanObjectNN accuracy (\%). Literature results retain native protocols; our controlled linear SVM uses OBJ-BG. Protocol details appear in Appendix~\ref{app:expanded_protocols}.}}
\label{tab:scan_main}

\begin{tabularx}{\linewidth}{@{}Lr@{}}
\toprule
Method & Acc.\\
\midrule
\multicolumn{2}{l}{\emph{\RevisionBlack{Supervised references}}}\\
\leavevmode\RevisionRed{PDNet \citep{yin2024pdnet}} & \RevisionRed{88.5}\\
\leavevmode\RevisionRed{PCM \citep{zhang2025pcm}} & \RevisionRed{88.1}\\
\leavevmode\RevisionRed{DeepLA-24 \citep{zeng2025deepla}} & \RevisionRed{90.6}\\
\midrule
\multicolumn{2}{l}{\emph{Frozen features: native protocols}}\\
Point-MAE \citep{pang2022masked} & 77.7\\
Point-M2AE \citep{zhang2022pointm2ae} & 84.1\\
CrossPoint \citep{afham2022crosspoint} & 81.7\\
I2P-MAE \citep{zhang2023i2p} & 87.1\\
MM-Point \citep{yu2024mmpoint} & 87.8\\
Point-MAE + TPM \citep{liu2025tpm} & 78.2\\
Point-M2AE + TPM \citep{liu2025tpm} & 84.0\\
MS-Point (DGCNN) & 89.6\\
\midrule
\multicolumn{2}{l}{\emph{Controlled OBJ-BG linear SVM}}\\
Point-MAE \citep{pang2022masked} & 78.1\\
Inter-MAE \citep{liu2024intermae} & 84.8\\
MM-Point (DGCNN) & 87.9\\
Point-MAE + TPM \citep{liu2025tpm} & 79.2\\
MS-Point (DGCNN) & 89.6\\
\rowcolor{gray!15} MS-Point-T (Transformer) & \textbf{90.4}\\
\bottomrule
\end{tabularx}
\end{minipage}
\end{table}

The first term regularizes each branch across input perturbations; the second coordinates semantic states across depth. Direct visual supervision and fused alignment are
\begin{equation}
\mathcal L_{\rm level}=|\mathcal A|^{-1}\sum_m\ell(\bar z_m,h_{\rho(m)}),\qquad
\mathcal L_{\rm fused}=\ell(\bar z_f,h_f).
\label{eq:v11_alignment}
\end{equation}
These terms anchor structural descriptors and their mixture to the same object semantics. To discourage redundant response directions, we add
\begin{equation}
\mathcal L_{\rm band}=|\mathcal S|^{-1}\sum_{l\in\mathcal S}\cos^2(z_{{\rm s},l},z_{{\rm d},l}),\qquad
\mathcal L=\sum_{r\in\mathcal R}\lambda_r\mathcal L_r,
\label{eq:v10_objective}
\end{equation}
where $\mathcal R=\{\mathrm{same},\mathrm{cross},\mathrm{level},\mathrm{fused},\mathrm{band}\}$ and the diversity term is averaged over point views and samples. The finite cosine penalty encourages specialization while the alignment terms preserve common semantics. All terms update the shared encoder jointly, balancing agreement across observations with differentiation across structural responses. Loss-removal and weight-transfer experiments test this balance; exact reductions and optimization settings remain in Appendix~\ref{app:detailed_formulation}.

The shared parameters $\theta$ receive gradients from both agreement and specialization:
\begin{equation}
\nabla_\theta\mathcal L=\sum_{r\in\mathcal R}\lambda_r\nabla_\theta\mathcal L_r.
\label{eq:v11_gradient}
\end{equation}
Branch-level alignment directly constrains states that the global loss reaches only through subsequent aggregation. Fused alignment instead rewards their collective prediction, while the diversity term discourages both response heads from choosing the same direction. Their weights control the balance of these training signals. \RevisionBlack{We evaluate each objective by removing it while keeping the complete architecture fixed.}

\par\endgroup
\begingroup
\section{Experiments}
\label{sec:exp}
\subsection{Training and Evaluation Protocols}
\label{sec:impl}\label{sec:tasks}
We pretrain on ShapeNet point--image pairs~\citep{chang2015shapenet} with two augmented point views and one encoded image per iteration. DGCNN and ResNet-50 form the default backbone pair; point Transformers and modern image encoders test portability~\citep{wang2019dynamic,he2016deep}. Optimization settings are deferred to Appendix~\ref{app:training_config}.

\RevisionRed{Our controlled reruns use the same pretraining collection, 1,024 input points, frozen-feature linear SVM evaluation, and no voting. On ScanObjectNN, these reruns use the OBJ-BG split.} \RevisionRed{Same-backbone comparisons test the objective; capacity- and computation-matched controls assess neighborhood information.} Published results retain their native protocols, while few-shot and dense tasks test broader transfer (Appendix~\ref{app:expanded_protocols}).

\subsection{Classification and Recent Baselines}
Tables~\ref{tab:modelnet}--\ref{tab:scan_main} compare synthetic ModelNet objects~\citep{wu20153d} and real ScanObjectNN scans~\citep{uy2019revisiting}. DGCNN-based MS-Point reaches 93.6/96.5 on ModelNet40/10 and 89.6 on ScanObjectNN, exceeding published MM-Point by 1.2/1.1/1.8 points. \RevisionRed{Controlled gains over MM-Point are 1.1/1.7 percentage points on ModelNet40/ScanObjectNN.} MS-Point-T reaches 94.3/97.2 on ModelNet40/10 and 90.4 on ScanObjectNN, improving over the evaluated masked-learning systems, including Inter-MAE and Point-MAE with TPM. \RevisionBlack{Accuracy improves on both synthetic objects and real scans.}

\subsection{Backbone Generality and Structural Controls}
\label{sec:modern_main}
With the point Transformer and ResNet-50 fixed, hierarchy, responses, and fusion progressively improve three downstream tasks (Table~\ref{tab:main_transformer}). The full objective gains 1.1 ModelNet40 points, 2.1 ScanObjectNN points, and 1.2 ShapeNetPart mIoU over global-only alignment. The benefit therefore extends to token features beyond EdgeConv.

Keeping heads, gates, losses, parameters, and computation fixed, real-neighborhood responses outperform duplicated features by 0.7/1.4 classification points and 0.6 part mIoU. Random neighborhoods and shuffled features are also weaker (Table~\ref{tab:main_controls}). \RevisionRed{With the image-target construction fixed, enabling point responses improves ScanObjectNN accuracy by 0.7--0.9 percentage points (Appendix~\ref{app:fixed_image_controls}).}

\subsection{Few-Shot Classification}
\begin{table}[!tbp]
\centering
\begin{minipage}[t]{.59\linewidth}
\vspace{0pt}

  \centering
  \caption{Few-shot classification on ModelNet40 and ScanObjectNN (mean$\pm$standard deviation).}
  
  \label{tab:few_shot}
  \renewcommand{\arraystretch}{1.0}
  \setlength{\tabcolsep}{1.4pt}
  \fontsize{7.7}{8.7}\selectfont
  \begin{tabularx}{\linewidth}{@{\hspace{1.5pt}}Lcccc@{}}
    \toprule
    \multirow{2}{*}{Method} & \multicolumn{2}{c}{5-way} & \multicolumn{2}{c}{10-way} \\
    \cmidrule(lr){2-3} \cmidrule(lr){4-5}
    & 10-shot & 20-shot & 10-shot & 20-shot \\
    \midrule
    \multicolumn{5}{c}{\textit{Results on ModelNet40}} \\
    \midrule
    3D-GAN \RevisionRed{\citep{wu2016learning}} & 55.8$\pm$3.4 & 65.8$\pm$3.1 & 40.3$\pm$2.1 & 48.4$\pm$1.8 \\
    PointCNN \RevisionRed{\citep{li2018pointcnn}} & 65.4$\pm$2.8 & 68.6$\pm$2.2 & 46.6$\pm$1.5 & 50.0$\pm$2.3 \\
    RSCNN \RevisionRed{\citep{liu2019relation}} & 65.4$\pm$8.9 & 68.6$\pm$7.0 & 46.6$\pm$4.8 & 50.0$\pm$7.2 \\
    \midrule
    Jigsaw3D \RevisionRed{\citep{sauder2019self}} & 34.3$\pm$1.3 & 42.2$\pm$3.5 & 26.0$\pm$2.4 & 29.9$\pm$2.6 \\
    OcCo \RevisionRed{\citep{wang2021unsupervised}} & 90.6$\pm$2.8 & 92.5$\pm$1.9 & 82.9$\pm$1.3 & 86.5$\pm$2.2 \\
    CrossPoint \RevisionRed{\citep{afham2022crosspoint}} & 92.5$\pm$3.0 & 94.9$\pm$2.1 & 83.6$\pm$5.3 & 87.9$\pm$4.2 \\
    MM-Point \RevisionRed{\citep{yu2024mmpoint}} & 96.5$\pm$2.8 & 97.2$\pm$1.4 & 90.3$\pm$2.1 & 94.1$\pm$1.9 \\
    \rowcolor{gray!15} \textbf{\method (Ours)} & \textbf{97.4$\pm$2.5} & \textbf{98.1$\pm$1.2} & \textbf{91.6$\pm$1.8} & \textbf{95.3$\pm$1.5} \\
    \midrule
    \multicolumn{5}{c}{\textit{Results on ScanObjectNN}} \\
    \midrule
    Jigsaw3D \RevisionRed{\citep{sauder2019self}} & 65.2$\pm$3.8 & 72.2$\pm$2.7 & 45.6$\pm$3.1 & 48.2$\pm$2.8 \\
    OcCo \RevisionRed{\citep{wang2021unsupervised}} & 72.4$\pm$1.4 & 77.2$\pm$1.4 & 57.0$\pm$1.3 & 61.6$\pm$1.2 \\
    CrossPoint \RevisionRed{\citep{afham2022crosspoint}} & 74.8$\pm$1.5 & 79.0$\pm$1.2 & 62.9$\pm$1.7 & 73.9$\pm$2.2 \\
    MM-Point \RevisionRed{\citep{yu2024mmpoint}} & 88.0$\pm$6.5 & 90.7$\pm$5.1 & 76.7$\pm$1.9 & 83.9$\pm$4.2 \\
    \rowcolor{gray!15} \textbf{\method (Ours)} & \textbf{89.8$\pm$5.8} & \textbf{92.4$\pm$4.5} & \textbf{78.5$\pm$1.7} & \textbf{85.8$\pm$3.8} \\
    \bottomrule
  \end{tabularx}

\end{minipage}\hfill
\begin{minipage}[t]{.39\linewidth}
\vspace{0pt}

\centering\footnotesize
\setlength{\tabcolsep}{3pt}
\caption{ShapeNetPart: OA and mIoU (\%).}
\label{tab:part_seg}

\begin{tabularx}{\linewidth}{@{}Lrr@{}}
\toprule
Method & OA & mIoU\\
\midrule
PointNet \citep{qi2017pointnet} & -- & 83.7\\
PointNet++ \citep{qi2017pointnetplusplus} & -- & 85.1\\
DGCNN \citep{wang2019dynamic} & -- & 85.1\\
\midrule
OcCo \citep{wang2021unsupervised} & 94.4 & 85.0\\
CrossPoint \citep{afham2022crosspoint} & 94.4 & 85.3\\
MM-Point \citep{yu2024mmpoint} & 94.5 & 85.7\\
MS-Point & \textbf{94.7} & \textbf{86.0}\\
\bottomrule
\end{tabularx}

\vspace{7pt}

\centering\footnotesize
\setlength{\tabcolsep}{3pt}
\caption{S3DIS: OA and mIoU over 13 classes~\citep{armeni20163d}.}
\label{tab:s3dis}

\begin{tabularx}{\linewidth}{@{}Lrr@{}}
\toprule
Method & OA & mIoU\\
\midrule
Jigsaw3D \citep{sauder2019self} & 84.1 & 55.6\\
OcCo \citep{wang2021unsupervised} & 84.6 & 58.0\\
CrossPoint \citep{afham2022crosspoint} & 86.7 & 58.4\\
MM-Point \citep{yu2024mmpoint} & 88.7 & 59.1\\
MS-Point & \textbf{89.4} & \textbf{59.8}\\
\bottomrule
\end{tabularx}

\end{minipage}
\end{table}

Table~\ref{tab:few_shot} evaluates 5-way and 10-way transfer. MS-Point leads the listed methods across both datasets, including generative and supervised references~\citep{wu2016learning,li2018pointcnn,liu2019relation}. On ScanObjectNN 5-way 10-shot, it reaches 89.8\%, compared with 88.0\% for MM-Point and 74.8\% for CrossPoint. The consistent ranking across label budgets supports reuse of the pretrained representation when task labels are scarce.

\subsection{Segmentation Transfer}
We fine-tune the pretrained encoder for point-wise prediction on ShapeNetPart~\citep{yi2016scalable} and S3DIS~\citep{armeni20163d}. MS-Point attains 86.0\% and 59.8\% mIoU, improving over MM-Point by 0.3 and 0.7 points (Tables~\ref{tab:part_seg}--\ref{tab:s3dis}). These gains show that features learned with object-level visual targets transfer to point-wise prediction. \RevisionBlack{The local extension adds spatial correspondences between point and image features to support dense prediction.}

\RevisionRed{Figure~\ref{fig:v8_part_segmentation_airplane_chair_earphone} shows MS-Point predictions for airplane, chair and earphone. Error and entropy maps locate part-label mistakes and uncertain predictions; sampling details are given in Appendix~\ref{sec:evaluation_design}.}

\begin{figure}[!t]
\centering
\includegraphics[width=.92\linewidth,height=.64\textheight,keepaspectratio]{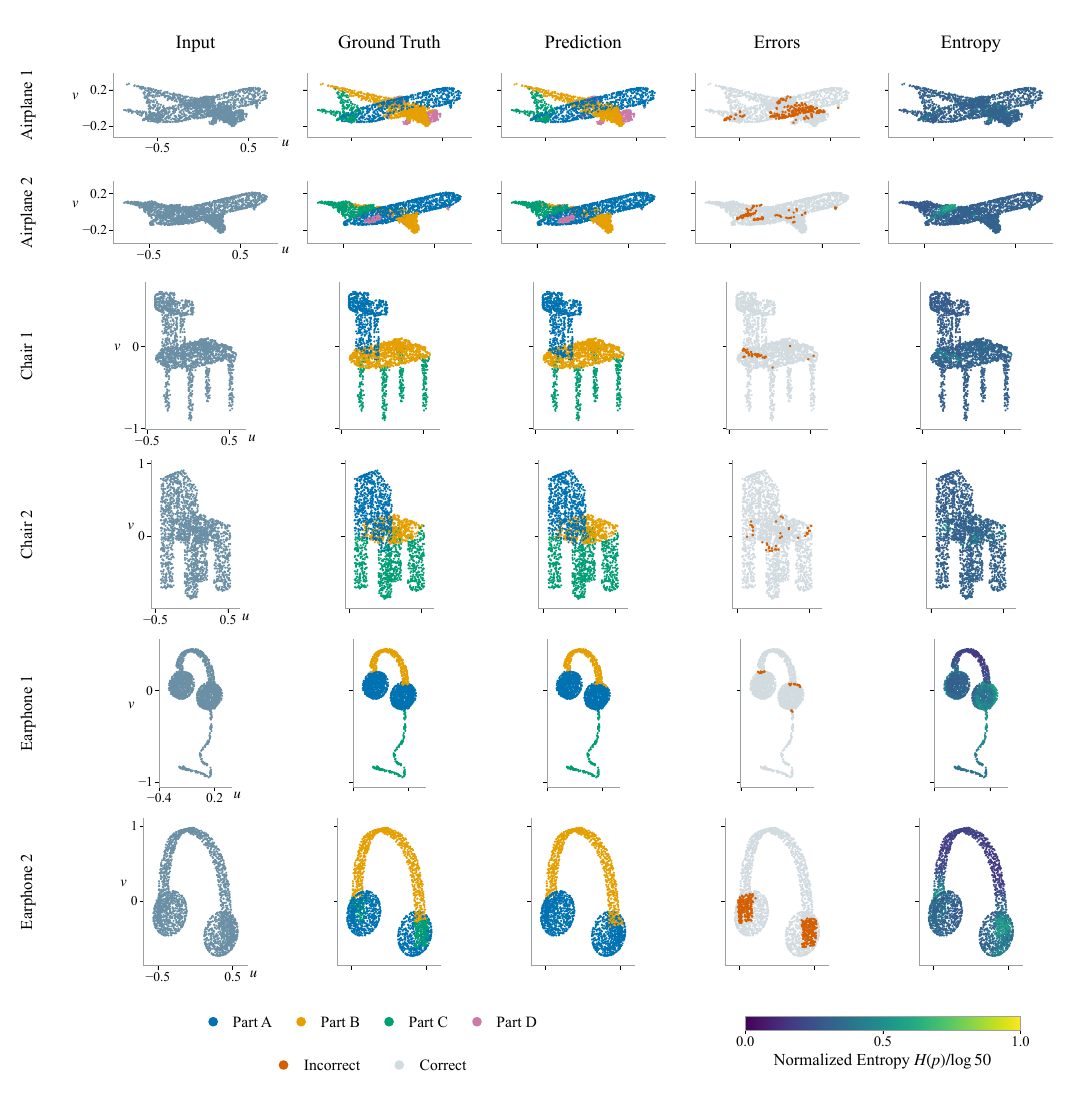}
\caption{\RevisionRed{\textbf{MS-Point part segmentation on ShapeNetPart.} Two examples per category, with 2,048 points each. Columns show input, ground truth, prediction, errors and entropy. Orange marks incorrect predictions; entropy is normalized by $\log 50$. Part colors are category-specific.}}
\label{fig:v8_part_segmentation_airplane_chair_earphone}
\end{figure}

\subsection{Ablations and Statistical Reliability}
\label{sec:ablation}

Intermediate supervision, cross-level coordination, and fusion progressively improve classification. Using both responses at $l_2,l_4,g$ reaches 89.6\%, versus 87.9\% for smoothing and 88.1\% for deviation alone (Tables~\ref{tab:ablate_hier}--\ref{tab:ablate_spec}). The fixed-capacity controls above complement this source-selection study.

\RevisionRed{For the five-seed full-training comparisons, the paired 95\% confidence intervals for accuracy gains over the global baseline and MM-Point lie above zero on both classification datasets} (Appendix~\ref{sec:expanded_results}).

\subsection{Efficiency and Dense Transfer}
\RevisionBlack{MS-Point obtains the features for all branches from a shared encoder forward pass.} Compared with the one-image global baseline, it adds 2.13\% parameters and 23.98\% GPU time; compared with four-image MM-Point, it reduces GPU time by 17.34\% and peak memory by 21.64\% (Table~\ref{tab:completed_cost}). MS-Point-T takes 30.4 H100 hours versus 72.0--76.5 for the evaluated MAE systems.

Geometry-preserving local alignment increases ShapeNetPart instance mIoU from 86.1 to 86.8 and S3DIS mIoU from 67.5 to 69.0 in its evaluation protocol (Table~\ref{tab:reported_K06}). On SemanticKITTI~\citep{behley2019semantickitti}, the local model obtains 47.3 mIoU with 1\% labels, compared with 34.2 from scratch; the full-label comparison is 65.2 versus 63.4 (Table~\ref{tab:completed_outdoor}). \RevisionBlack{The larger gain with fewer labels supports the value of pretraining for low-label semantic segmentation.}

\subsection{Representation Diagnostics}
\label{sec:visual_diagnostics}

\begin{figure}[!tbp]
\centering\begin{minipage}[t]{.48\linewidth}\vspace{0pt}

\centering
\includegraphics[width=\linewidth]{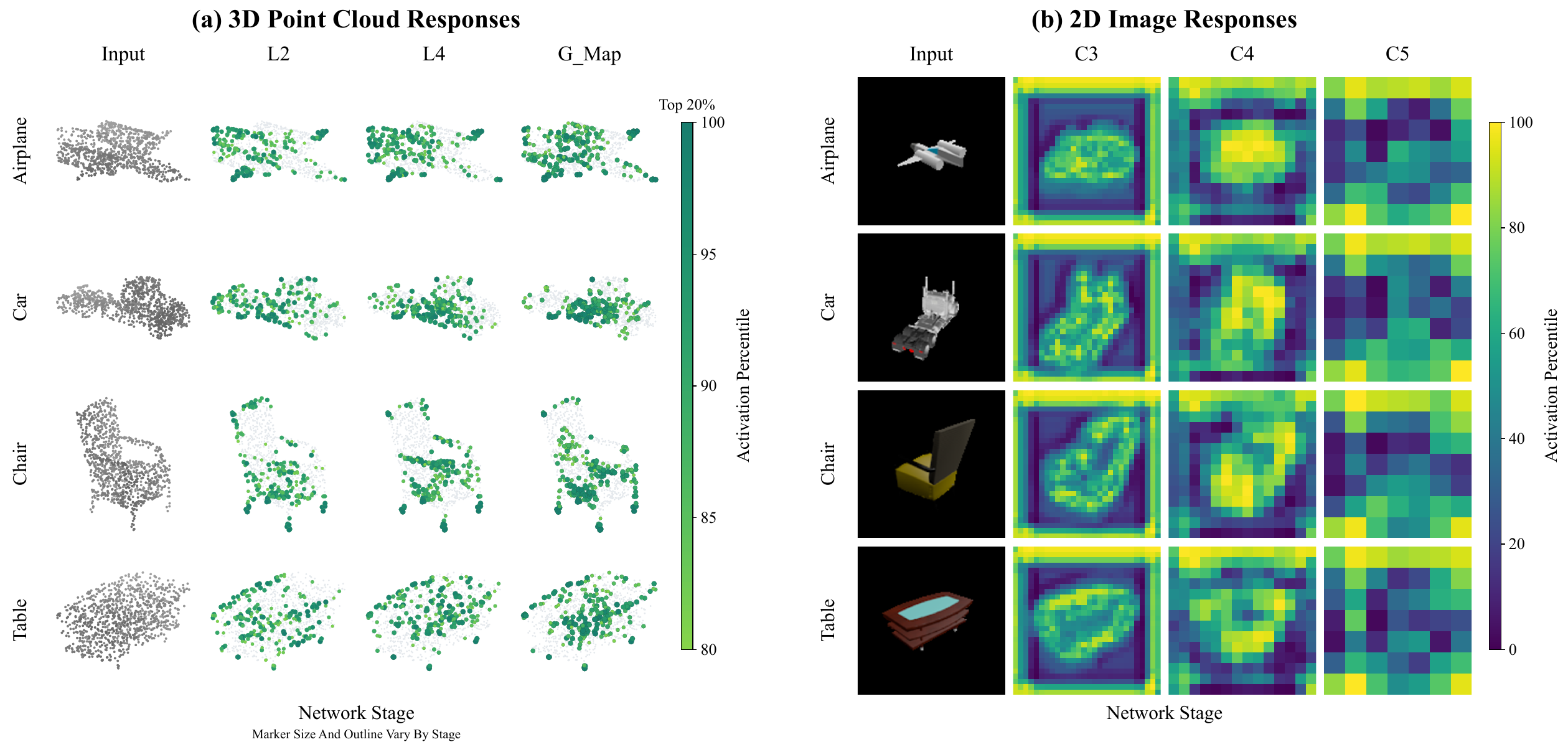}
\caption{\textbf{Depth-dependent responses.} Colors show within-stage feature-norm ranks for four objects.}
\label{fig:depth_activations}

\end{minipage}\hfill\begin{minipage}[t]{.48\linewidth}\vspace{0pt}

\centering
\includegraphics[width=\linewidth]{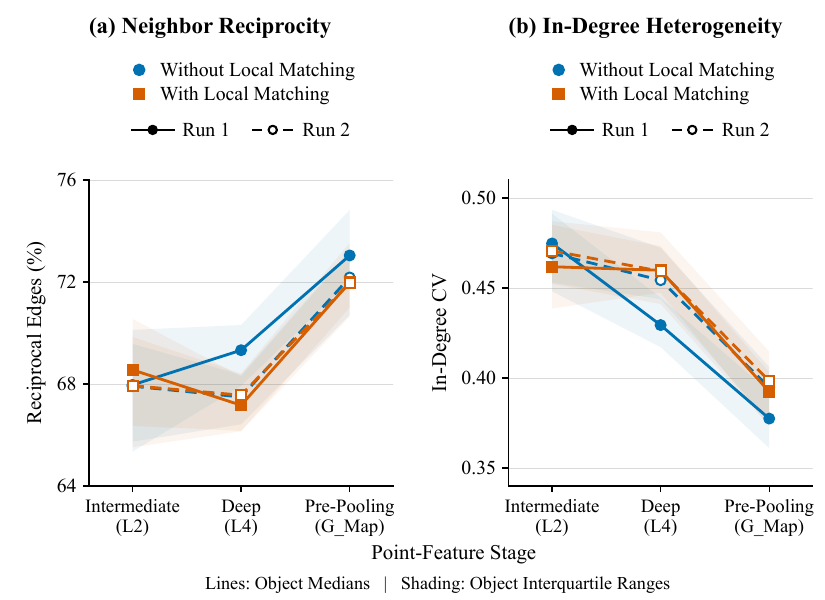}
\caption{\textbf{Directed neighborhoods.} Reciprocity and in-degree variation over 52 objects; bands show interquartile ranges.}
\label{fig:directed_neighborhoods}

\end{minipage}\end{figure}

Figure~\ref{fig:depth_activations} shows how activation patterns change across point and image stages. \RevisionRed{Features at different encoder stages emphasize different spatial regions, providing distinct inputs to the branch objectives.} Color ranks show the relative activation of locations within each stage.

We further inspect 52 ShapeNet objects across two validation-selected runs, with and without pooled local matching. These diagnostics examine the internal state underlying the transfer results, using the same object set for each comparison. \RevisionBlack{Sampling and model-selection details are provided} in Appendix~\ref{app:diagnostic_protocols}.

From $l_4$ to the pre-pooling state, neighborhood reciprocity rises and in-degree variation falls (Figure~\ref{fig:directed_neighborhoods}). Deep aggregation thus changes the organization of the feature graph. Non-reciprocal edges persist, supporting a directed graph-shift formulation for the implementation and a symmetric reference for spectral analysis.

\RevisionBlack{The neighborhood connections change across encoder stages, so response branches aggregate features over different local relations.} \RevisionBlack{Reciprocity measures the fraction of neighborhood links that are mutual.}

Point bins and image grids show how many spatial locations contribute nonzero descriptors in each run (Figure~\ref{fig:local_token_activity}). This variation motivates retaining spatial support explicitly when constructing local alignment targets.

\begin{figure}[!tbp]
\centering\begin{minipage}[t]{.48\linewidth}\vspace{0pt}

\centering
\includegraphics[width=\linewidth]{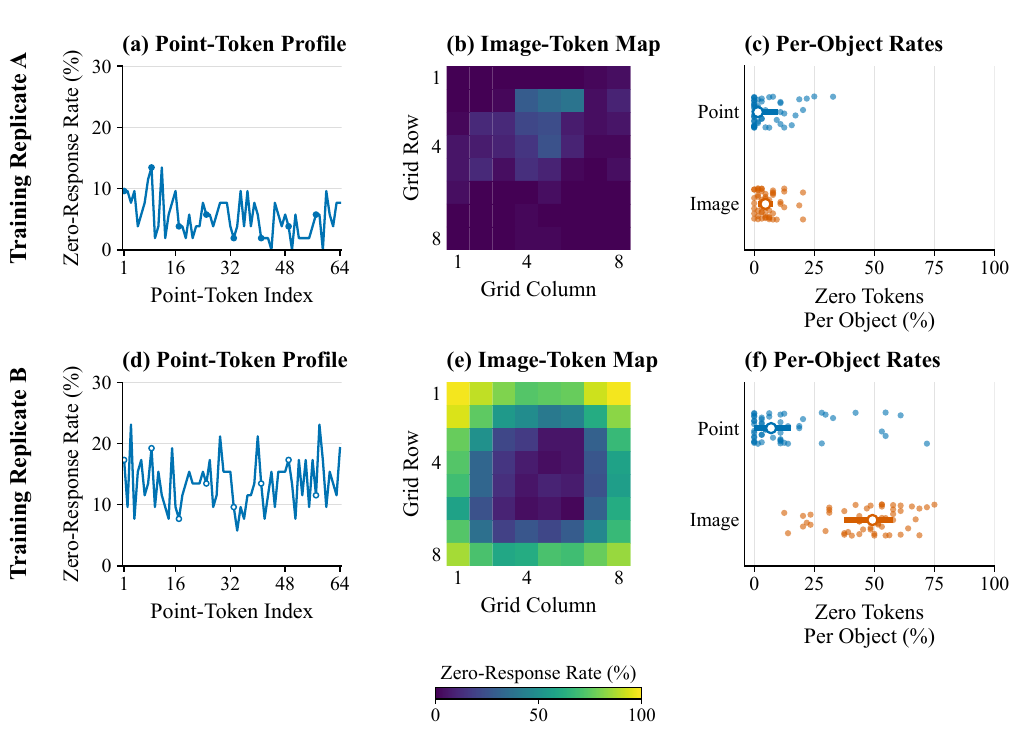}
\caption{\textbf{Local-token activity.} Zero-vector rates for point bins and image grids in two local-matching runs.}
\label{fig:local_token_activity}

\end{minipage}\hfill\begin{minipage}[t]{.48\linewidth}\vspace{0pt}

\centering
\includegraphics[width=\linewidth]{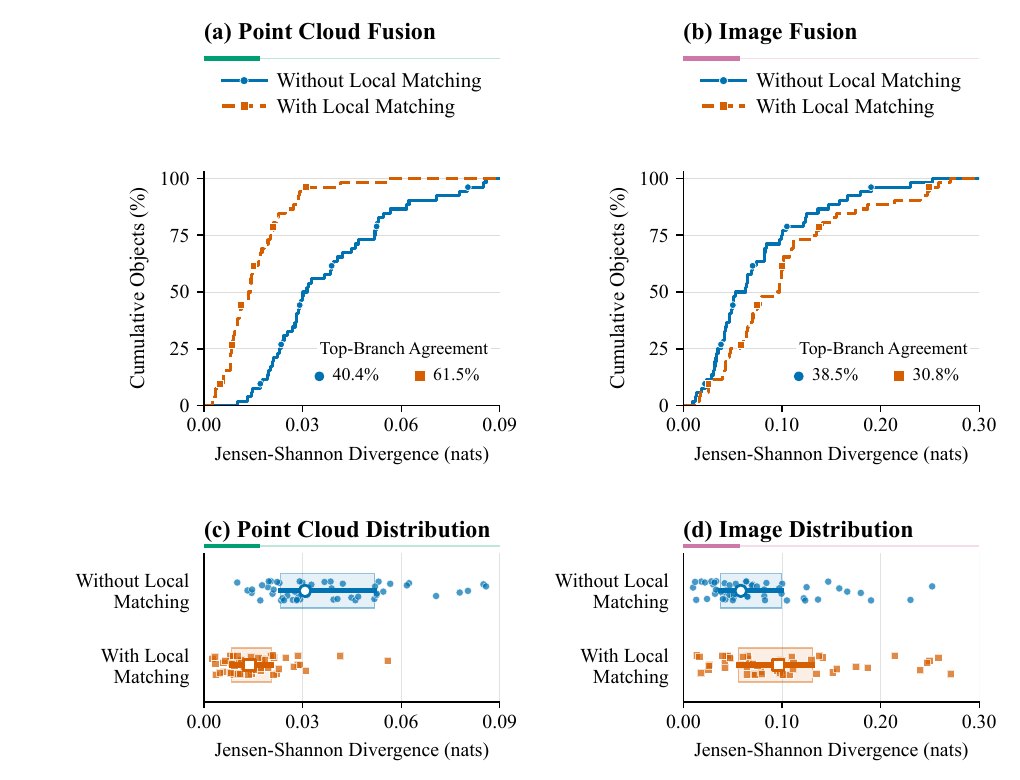}
\caption{\textbf{Fusion consistency.} Cross-run gate divergence and top-branch agreement, measured separately in each modality.}
\label{fig:gate_consistency}

\end{minipage}\end{figure}

This distinction guides geometry-preserving alignment: token construction should maintain useful spatial support, and visibility should determine which observations share evidence. Local supervision complements object identity by specifying where a cross-modal match is supported.

With local matching, point-side agreement on the highest-weight branch rises from 40.4\% to 61.5\% across runs (Figure~\ref{fig:gate_consistency}). Image-side agreement changes from 38.5\% to 30.8\%.

\RevisionBlack{These diagnostics characterize spatial activation patterns, neighborhood connectivity, and branch activity. They complement the downstream comparisons by showing how features and fusion weights vary across stages and runs.}

\par\endgroup
\FloatBarrier
\begingroup
\section{Conclusion}\label{sec:main_conclusion}

\RevisionRed{MS-Point adds cross-modal supervision at selected encoder stages. Separate branches pool and project the neighborhood averages and residuals, then align the resulting descriptors with image features.} Controlled comparisons and transfer evaluations support this design across graph and token encoders. The resulting principle extends naturally toward local objectives in which geometry and visibility identify the information shared across modalities.

\RevisionBlack{The results support supervising intermediate features and neighborhood responses alongside the final object representation. Explicit geometric correspondence and visibility provide a basis for extending these targets to dense prediction.}
\par\endgroup
\clearpage
\begingroup
\section*{Reproducibility Statement}
Section~\ref{sec:method} defines the branches, neighborhood operators, pooling, and losses. Appendices~\ref{app:implementation}--\ref{app:extra_ablation} give dataset and training settings. Appendix~\ref{sec:expanded_results} provides capacity-matched controls and multi-seed evaluations, Appendix~\ref{app:reported_observations} gives the independently tuned baselines, and Appendix~\ref{app:token_visibility} specifies the token-encoder interface.

\par\endgroup
\begingroup
\section*{Ethics Statement}
This work studies representations learned from existing point-cloud benchmarks and paired images. \RevisionBlack{Deployment requires additional evaluation across sensors, environments, and demographic contexts.} Any reuse of the data must follow the original dataset terms. Dataset-specific biases and the limitations of rendered views should be considered when transferring the models to real scenes.

\par\endgroup
\begingroup
\section*{AI Use Statement}
We used generative AI for language editing, grammar checks, and LaTeX formatting, as well as for feedback on experimental plans, run summaries, and figure and table preparation. The authors conducted the experiments, checked numerical comparisons against experiment records, and take full responsibility for the manuscript.

\clearpage
\appendix

\par\endgroup
\makeatletter
\let\caption\VnineCaption
\let\@caption\VnineAtCaption
\let\@makecaption\VnineMakeCaption
\makeatother
\UseAppendixRevisionStyle
\section{Analysis of Structured Supervision}
\label{sec:interface_analysis}

\RevisionBlack{The visual analyses examine nine-branch pretraining models and DGCNN segmentation models. The corresponding sections specify the model configurations, sampling procedures and visualization scales.}

MS-Point changes both the location and the inputs of cross-modal supervision. We analyze these choices separately from backbone expressiveness and additional head capacity, then describe how the joint losses and fusion affect their interpretation.

\begingroup\centering\small\setlength{\tabcolsep}{3pt}\renewcommand{\arraystretch}{1.04}
\noindent\begin{minipage}[t]{.485\linewidth}\vspace{0pt}\centering\footnotesize\setlength{\tabcolsep}{2pt}
\intablecaption{Hierarchy and fusion on ScanObjectNN.}\label{tab:ablate_hier}
\begin{tabularx}{\linewidth}{@{}Lr@{}}
\toprule
Cumulative change & Acc.\\
\midrule
Global only & 81.7\\
+ Intermediate targets & 84.5\\
+ Cross-level coordination & 85.8\\
+ Adaptive fusion & \textbf{86.5}\\
\bottomrule
\end{tabularx}
\par\smallskip{\footnotesize Single-image baseline protocol. The 86.5\% hierarchy reference in Table~\ref{tab:ablate_spec} includes adaptive fusion. For full-training G and H, see Table~\ref{tab:factorial_part_complete}; five-seed results are in Table~\ref{tab:completed_seeds}.}

\end{minipage}\hfill\begin{minipage}[t]{.485\linewidth}\vspace{0pt}\centering\footnotesize\setlength{\tabcolsep}{2pt}
\intablecaption{Response type and source levels.}\label{tab:ablate_spec}\label{tab:app_spectral_levels}
\begin{tabularx}{\linewidth}{@{}LLrr@{}}
\toprule
Response & Sources & Branches & Acc.\\
\midrule
Hierarchy only & None & 0 & 86.5\\
Both & \(\{l_2\}\) & 2 & 87.4\\
Both & \(\{l_4\}\) & 2 & 87.2\\
Both & \(\{g\}\) & 2 & 87.0\\
Both & \(\{l_2,l_4\}\) & 4 & 88.3\\
Smoothing & $l_2,l_4,g$ & 3 & 87.9\\
Deviation & $l_2,l_4,g$ & 3 & 88.1\\
Both & \(\{l_2,l_4,g\}\) & 6 & \textbf{89.6}\\
\bottomrule
\end{tabularx}
\par\smallskip{\footnotesize Each source contributes one branch per active response type. Table~\ref{tab:main_controls} gives the capacity-matched controls.}

\end{minipage}
\endgroup\par\medskip

\subsection{Supervision Location and Observable Statistics}
Let \(H_l(X;\theta)\) be the feature map at stage \(l\), \(\Phi\) an object-pooling operator, and \(q_l\) a projection head. Final-only alignment evaluates \(q_g\Phi(H_g)\), whereas a hierarchy objective evaluates a family \(\{q_l\Phi(H_l)\}\). The final loss already updates earlier stages through the chain rule. Direct targets add stage-specific readouts and gradient paths. Concatenating stage features before pooling gives the global loss a single descriptor, whereas branch-level targets supervise each selected stage directly.

\RevisionBlack{Matching the number of active projection heads at the final map and at intermediate stages provides a control for supervision location. Active heads contribute to optimization, whereas unused parameters leave the training path unchanged.}

\paragraph{Pooling can hide different neighborhood arrangements.}
For a fixed pooling operator $\Phi$ and deterministic projection $q$, $\Phi(H)=\Phi(H')$ implies that $q\circ\Phi$ and its loss against any image embedding assign the same value to both feature maps. The structural distinction is absent from the objective's current observations. A neighborhood-dependent readout can separate this equivalence class, giving the loss direct access to a difference that additional post-pooling heads cannot recover.

Consider a three-node path with self-loops. \RevisionRed{In this example, \(P\) is its row-normalized graph operator:}
\begin{equation}
 P=\begin{bmatrix}1/2&1/2&0\\1/3&1/3&1/3\\0&1/2&1/2\end{bmatrix},
 \quad h=(1,0,0)^\top,\quad h'=(0,1,0)^\top.
 \label{eq:pool_example}
\end{equation}
Both signals have max--mean descriptor \((1,1/3)\). However,
\begin{equation}
 Ph=(1/2,1/3,0)^\top,\qquad Ph'=(1/2,1/3,1/2)^\top,
\end{equation}
whose means are \(5/18\) and \(4/9\). The deviation maxima also differ: \(1/2\) versus \(2/3\). Responses before pooling can therefore distinguish these two arrangements. \RevisionBlack{This algebraic example illustrates how neighborhood responses distinguish features with identical pooled descriptors.}

\paragraph{Max and mean pooling of residual responses.}
For a doubly stochastic neighborhood matrix $P$, the raw residual $R=(I-P)H$ satisfies $\mathbf{1}^{\top}R=0$. Positive and negative entries thus cancel in the channel mean. In every nonzero residual channel, however, the maximum is positive. Combining the channel maximum and mean (Eq.~\ref{eq:v11_pool}) can therefore retain local deviations whose mean alone is zero.

\subsection{Interactions among the Training Objectives}
The five objectives act simultaneously on shared encoder parameters, so their effects depend on their interactions. Same-branch contrast encourages agreement across point augmentations. Cross-level contrast coordinates features across depths with a default weight of $\lambda_{\rm cross}=0.4$. Branch-level image alignment supplies direct visual targets, while fused alignment constrains the aggregate. \RevisionRed{When branch alignment is removed, the fused objective continues to provide image supervision.}

Soft directional decoupling acts in the projected space, where each response can retain object identity while expressing a different direction. Appendix~\ref{app:graph_scope} derives its tangential gradient. We evaluate this mechanism through transfer accuracy and response similarity together: \RevisionBlack{we seek complementary responses that improve transfer while preserving shared object semantics.}

\begin{figure}[!tbp]
\centering
\begin{minipage}{1.0\linewidth}
\centering
\includegraphics[width=\linewidth]{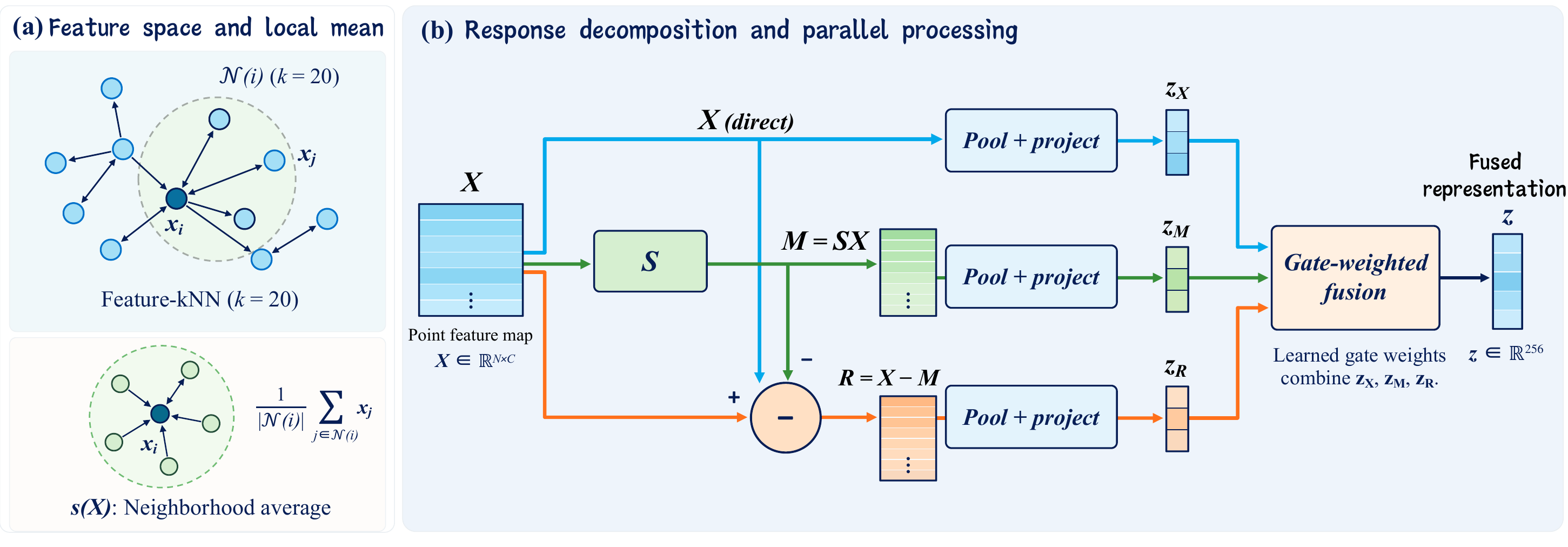}
\caption{\textbf{From a local neighborhood to independent response paths.} A feature-space neighborhood defines an average and a node-wise deviation. The direct, mean, and residual paths are pooled and projected independently before gated fusion. The illustrated neighborhood uses $k=20$.}
\label{fig:branch_paths}
\end{minipage}
\end{figure}

\begin{figure}[!htbp]
\centering
\begin{minipage}{0.88\linewidth}
\centering
\includegraphics[width=\linewidth]{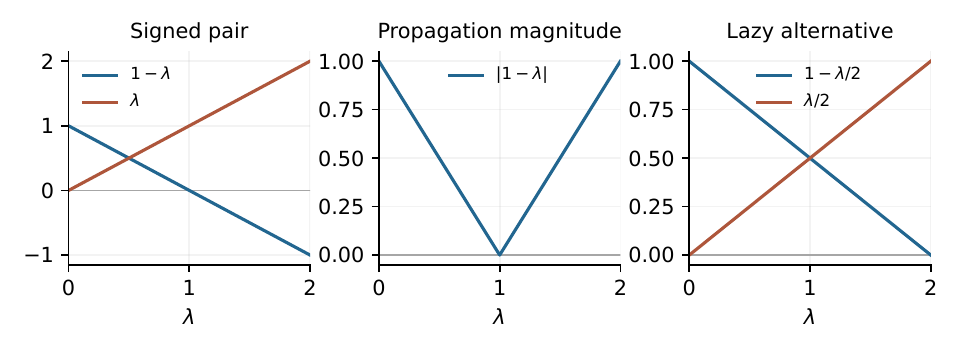}
\caption{Theoretical spectral response profiles on a symmetric reference graph. Left: the standard signed pair $1-\lambda$ and $\lambda$. Center: the magnitude $|1-\lambda|$ is not monotone on $[0,2]$. Right: a lazy reference pair with $\eta=1/2$ has nonnegative monotone responses. The horizontal axis gives the eigenvalue $\lambda$ of the symmetric normalized Laplacian.}
\label{fig:analytic_responses}
\end{minipage}
\end{figure}

\begin{figure}[!htbp]
\centering
\includegraphics[width=1.0\linewidth,height=.78\textheight,keepaspectratio]{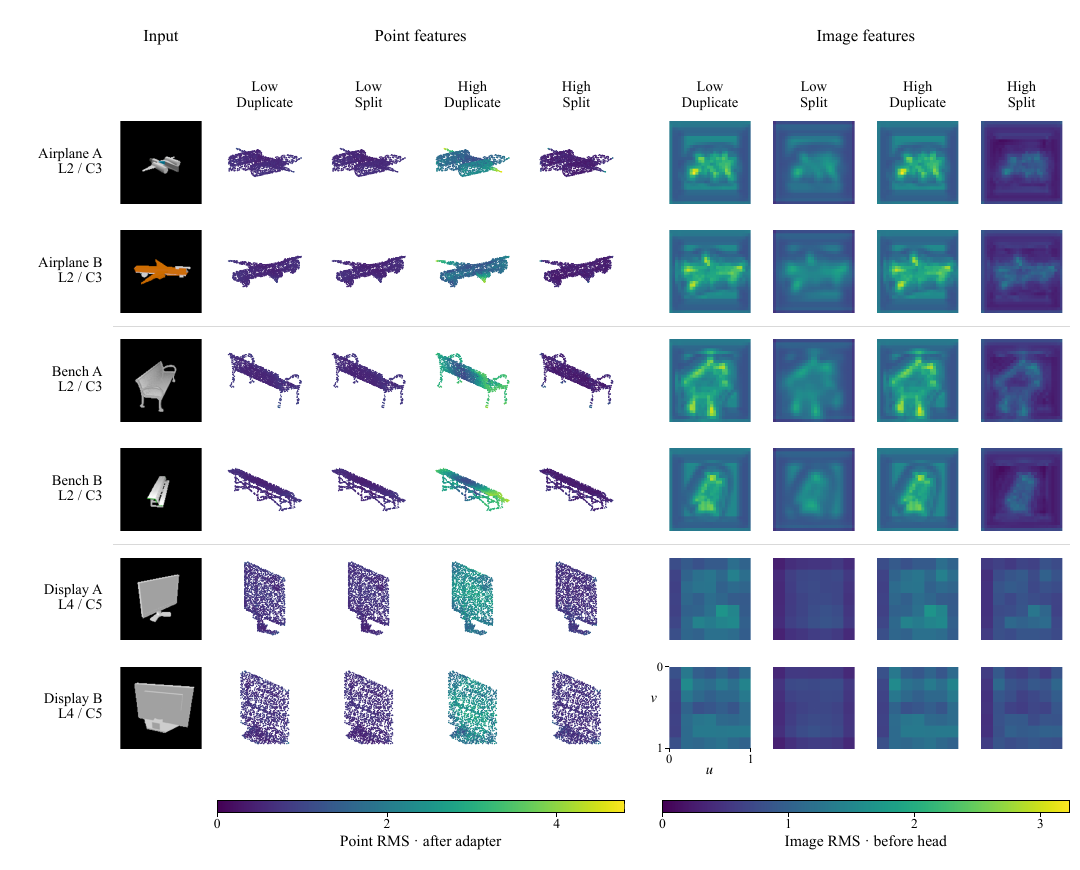}
\caption{\textbf{Response construction with fixed weights.} Six objects from the nine-branch response model \RevisionBlack{(trained for 100 epochs)}. Duplicate sends the same feature map to both paths; Split uses neighborhood consensus and deviation. Point colors show RMS after each adaptor; image colors show RMS before the head, with a separate scale. Low/high denote the consensus/deviation paths. This is a frozen-weight input intervention.}
\label{fig:v8_operator_response_comparison}
\end{figure}

\paragraph{Responses before and after decomposition.}
Figure~\ref{fig:v8_operator_response_comparison} compares duplicated features with consensus and deviation inputs in a frozen model. Switching the inputs changes the spatial activation pattern within each adaptor, showing which regions respond to the neighborhood mean and which to the residual.

\subsection{Fusion and the Deployed Representation}
For branch embeddings \(z_m\), the gates produce
\begin{equation}
 \alpha_m=\frac{\exp(a_m(z_m))}{\sum_{n\in\mathcal A}\exp(a_n(z_n))},
 \qquad z_f=q_f\!\left(\sum_{m\in\mathcal A}\alpha_m z_m\right).
\end{equation}
\RevisionBlack{The gate \(\alpha_m\) specifies a branch's weight in the fused representation.} An input branch with a small weight can still affect training through its direct loss. In addition, gate weights and embedding norms jointly determine the weighted sum. Useful diagnostics include gate entropy, branch norms, retrieval quality per branch, and controlled branch interventions.

\RevisionBlack{In backbone-only transfer (Figure~\ref{fig:training_transfer_paths}), the object head receives a pooled point descriptor and the dense head receives point features. Both paths bypass the auxiliary branches and fusion gates at inference.}

\section{Expanded Evaluation of Structured Supervision}
\label{sec:expanded_results}
We evaluate recent baselines, modern encoders, controlled response inputs, repeated runs, and label-efficient scene learning. The complete results below supplement the main comparisons. Full-training experiments and exploratory runs are analyzed separately because they use different budgets and evaluation endpoints.

\subsection{Recent Baselines under a Common Evaluator}
Our reruns use the same ShapeNet subset or point--image pairs, 1,024 points, ScanObjectNN OBJ-BG, no voting, and a frozen linear SVM. Optimizer-search scope, augmentation settings, and seed count are held common in this comparison. \RevisionBlack{TPM supplies the masking strategy for the Point-MAE backbone.} \RevisionBlack{These comparisons evaluate complete pretraining systems, including their respective point-only or multimodal inputs.}

MS-Point improves over the controlled MM-Point rerun by 1.1 and 1.7 percentage points. These differ from the 1.2/1.8-point gaps to the published MM-Point row because the reference is different. The Transformer instance reaches 94.3/90.4. Appendix~\ref{app:expanded_protocols} separates the literature compilation from these reruns.

\subsection{Changing the Backbone while Keeping the Objective Testable}

\par\noindent\begin{minipage}[t]{.49\linewidth}\vspace{0pt}

\centering\footnotesize
\setlength{\tabcolsep}{2pt}
\intablecaption{Same point Transformer and ResNet-50 image encoder. Part denotes ShapeNetPart mIoU.}
\label{tab:completed_transformer}\label{tab:main_transformer}
\begin{tabularx}{\linewidth}{@{}Lrrrr@{}}
\toprule
Objective & MN40 & Scan & Part & H100 h\\
\midrule
Global only & 93.2 & 88.3 & 85.7 & 27.4\\
+ Hierarchy & 93.6 & 89.0 & 86.1 & 28.5\\
+ Responses & 94.0 & 89.8 & 86.5 & 29.6\\
+ Fusion (MS-Point-T) & 94.3 & 90.4 & 86.9 & 30.4\\
\bottomrule
\end{tabularx}

\end{minipage}\hfill\begin{minipage}[t]{.48\linewidth}\vspace{0pt}
On the same Transformer, hierarchy, neighborhood responses, and fusion successively improve ScanObjectNN accuracy by 0.7, 0.8, and 0.6 points. Appendix~\ref{app:token_visibility} specifies the token graph and selected blocks.
\end{minipage}\par\medskip
\begin{table}[!htbp]\centering\small\setlength{\tabcolsep}{3pt}\renewcommand{\arraystretch}{1.08}
\caption{Neighborhood controls and independent point and image response comparisons.}
\label{tab:completed_compute}\label{tab:main_controls}\label{tab:reported_NS03}\label{tab:geometry_part_complete}\label{tab:control_times_complete}
\begin{tabularx}{\linewidth}{@{}LLLrrrr@{}}
\toprule
Construction & \makecell{Same graph\\cost} & \makecell{Same active\\branches} & MN40 & Scan & Part & \makecell{min/\\epoch}\\
\midrule
Duplicate & $\checkmark$ & $\checkmark$ & 92.9 & 88.2 & 85.5 & 50.1\\
Random graph & $\checkmark$ & $\checkmark$ & 93.0 & 88.3 & 85.6 & 50.3\\
Shuffled features & $\checkmark$ & $\checkmark$ & 93.1 & 88.5 & 85.7 & 50.2\\
Smooth only & $\checkmark$ & One active & 93.2 & 88.8 & 85.7 & --\\
Deviation only & $\checkmark$ & One active & 93.3 & 89.0 & 85.8 & --\\
Pair & $\checkmark$ & $\checkmark$ & 93.6 & 89.6 & 86.1 & 50.0\\
\bottomrule
\end{tabularx}
\par\smallskip{\footnotesize The duplicate, random-graph, shuffled-feature, and paired-response controls match in parameters and computation. Table~\ref{tab:reported_GS03} gives the separate single-response study.}
\par\smallskip{\footnotesize\textbf{Point and image responses varied independently.}\par
\begin{tabularx}{\linewidth}{@{}LLrrr@{}}\toprule
Point input & Image target & MN40 & Scan & Part\\\midrule
Duplicated features & Duplicated features & 92.9 & 88.2 & 85.5\\
Mean and residual & Duplicated features & 93.3 & 88.9 & 85.8\\
Duplicated features & Mean and residual & 93.2 & 88.7 & 85.7\\
Mean and residual & Mean and residual & 93.6 & 89.6 & 86.1\\\midrule
Point gain (pp) & Duplicated features & \textbf{+0.4} & \textbf{+0.7} & \textbf{+0.3}\\
Point gain (pp) & Mean and residual & \textbf{+0.4} & \textbf{+0.9} & \textbf{+0.4}\\\bottomrule
\end{tabularx}\par All four runs use 73.37M parameters, 15.2 GFLOPs, and 30.5 H100 hours per training run. Scores are means over three paired pretraining seeds. Scan denotes OBJ-BG and Part denotes ShapeNetPart instance mIoU.}
\end{table}

\begin{figure}[!tbp]
\begin{minipage}[t]{0.485\linewidth}
\centering
\includegraphics[width=\linewidth]{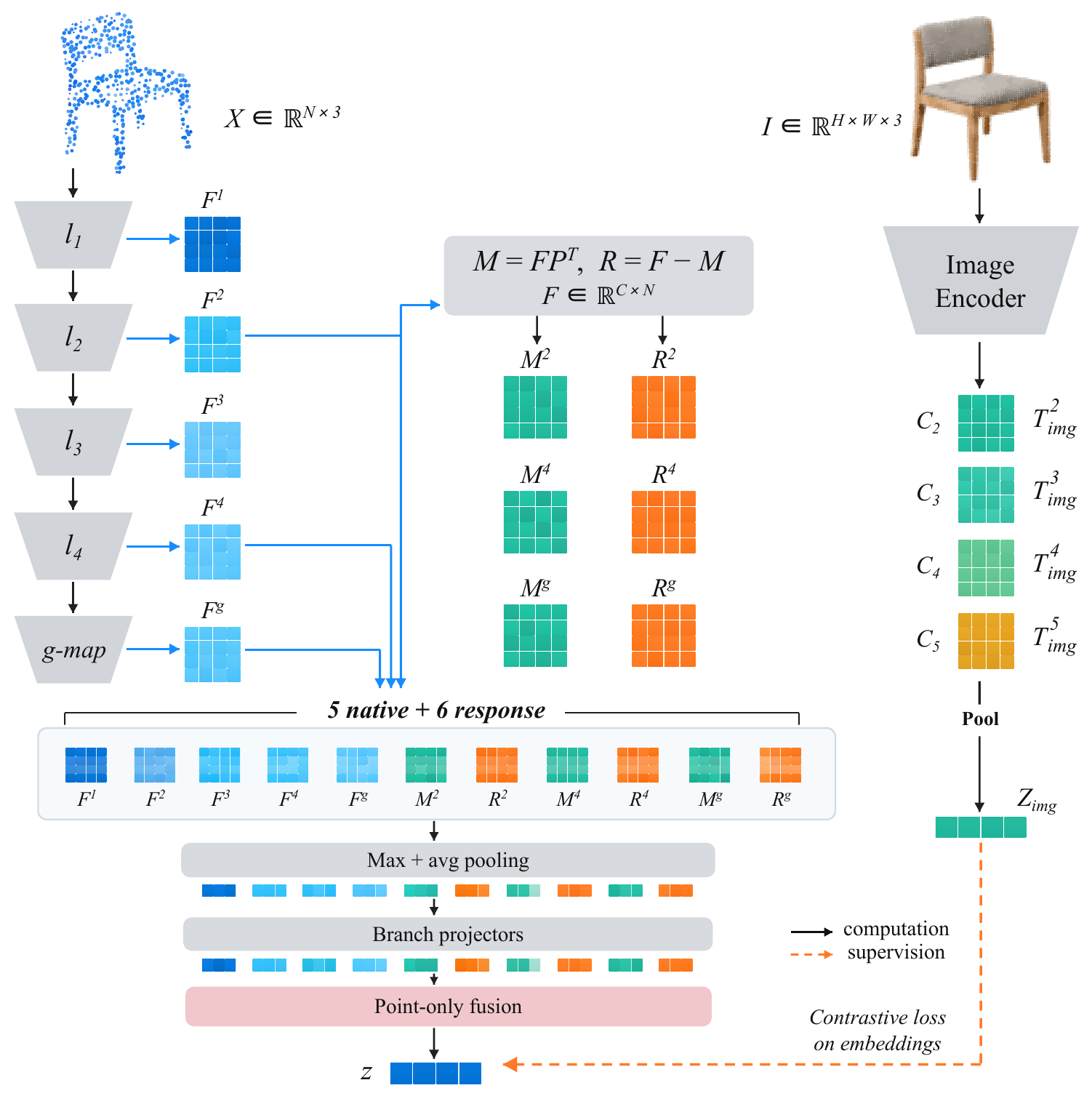}
\caption{\textbf{Eleven point branches.} Five native maps and six responses at $l_2,l_4,g$ enter pooling, independent projectors, and point-side fusion. Blue lines show computation; orange dashed lines denote embedding supervision. Image stages provide training targets.}
\label{fig:eleven_branch_flow}
\end{minipage}\hfill
\begin{minipage}[t]{0.485\linewidth}
\centering
\includegraphics[width=\linewidth]{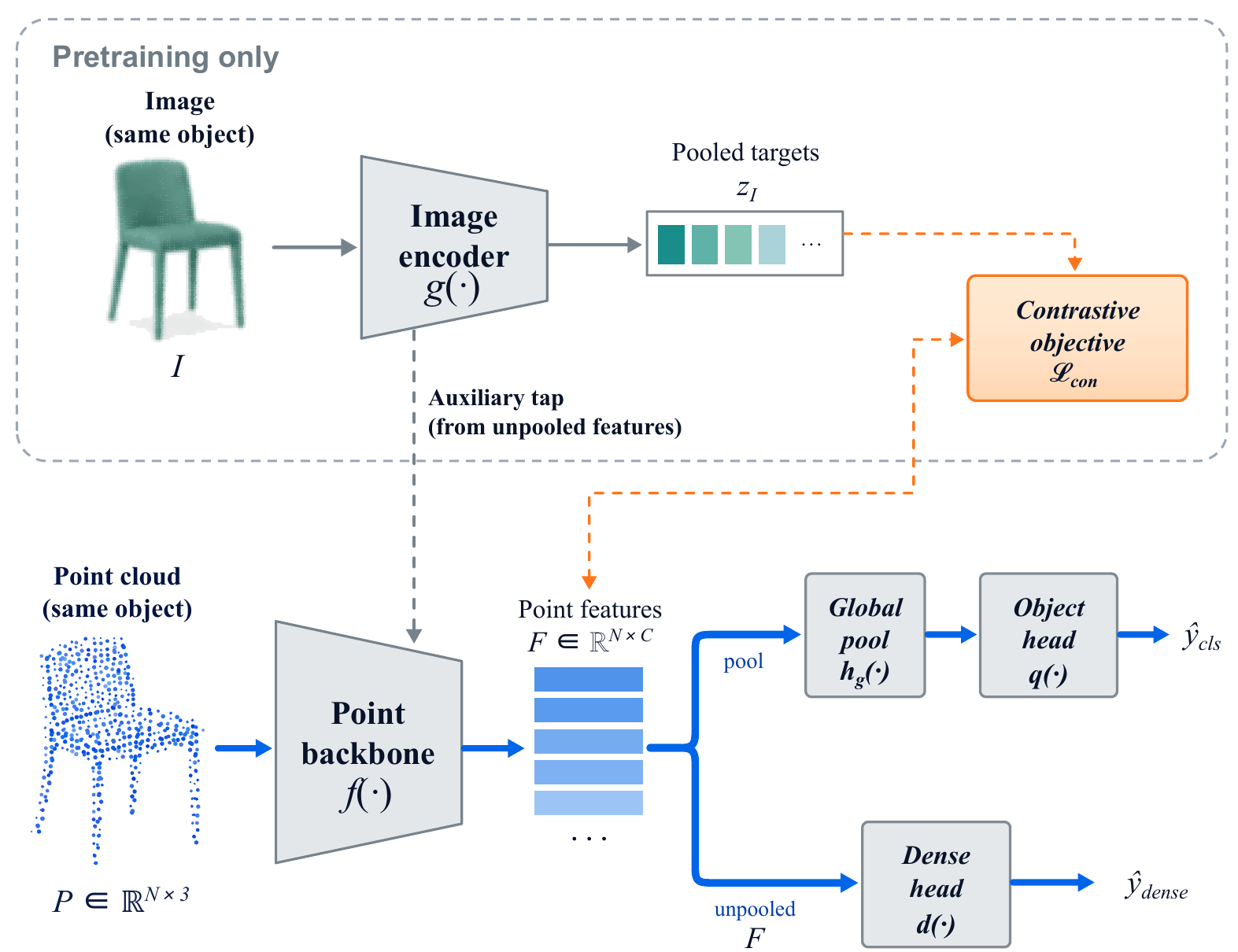}
\caption{\textbf{Training and transfer paths.} Image features provide auxiliary supervision during pretraining. The point encoder then supplies a pooled descriptor to an object head or unpooled features to a dense head. Downstream point-only inference does not require the image pathway.}
\label{fig:training_transfer_paths}
\end{minipage}
\end{figure}

\begin{figure}[!tbp]
\centering
\begin{minipage}{0.84\linewidth}
\centering
\includegraphics[width=\linewidth]{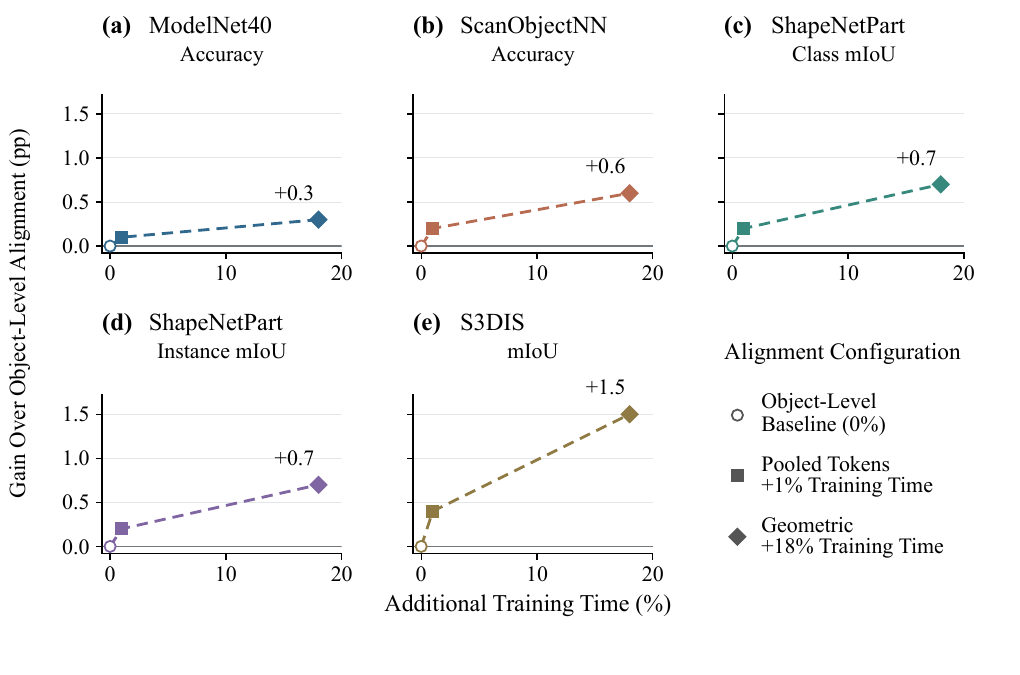}
\caption{\textbf{Local targets and training cost.} Gains are relative to object-level alignment within each task (Table~\ref{tab:reported_K06}). Pooled tokens add 1\% training time and geometry-preserving targets add 18\%. ShapeNetPart class and instance mIoU are separate metrics. The S3DIS gains use the object-level baseline of 67.5 mIoU from the same study.}
\label{fig:local_alignment_tradeoff}
\end{minipage}
\end{figure}

\subsection{Do Neighborhood Inputs Matter beyond Additional Heads?}\label{app:fixed_image_controls}
The random-graph and shuffled-feature controls change only the point neighborhoods or feature assignments. Both use $3\times3$ image mean and residual targets. In the four-way comparison, image stages ($C_3,C_5,C_5$), views, heads, gates, and loss weights are fixed. \RevisionRed{Point responses add 0.7 or 0.9 ScanObjectNN points with duplicated or filtered image targets, respectively.} Classification uses a frozen-feature linear SVM, 1,024 points, and no voting.

For each frozen nine-branch model, both point inputs use identical image embeddings. The margin is the paired cosine similarity minus the mean similarity to the 25 other images.
\par\smallskip\noindent\begin{minipage}{\linewidth}\centering\footnotesize
\textbf{Matching margin with fixed image embeddings (26 objects).}\par\smallskip
\begin{tabularx}{\linewidth}{@{}Lrrrr@{}}\toprule
& \multicolumn{2}{c}{Frozen model 1} & \multicolumn{2}{c}{Frozen model 2}\\
Image target & Duplicate $\to$ responses & $\Delta$ & Duplicate $\to$ responses & $\Delta$\\\midrule
Duplicated features & $0.1466\to0.3161$ & $+0.1695$ & $0.0110\to0.3227$ & $+0.3117$\\
Mean and residual & $0.4362\to0.8035$ & $+0.3673$ & $-0.0140\to0.7355$ & $+0.7495$\\\bottomrule
\end{tabularx}\end{minipage}\par

\subsection{Repeated Runs and Practical Efficiency}
We use five independent pretraining seeds, select checkpoints on validation data, and evaluate the test set once after selection. Table~\ref{tab:completed_seeds} gives the mean and standard deviation. Paired ModelNet40 intervals are $[+0.62,+0.98]$ against the global baseline and $[+0.85,+1.35]$ against MM-Point. The corresponding ScanObjectNN intervals are $[+1.38,+2.02]$ and $[+1.34,+2.06]$. All four intervals are above zero. Pairing is performed across runs; marginal standard deviations alone do not determine the paired uncertainty.
\begin{table}[!tbp]\centering\small\setlength{\tabcolsep}{3pt}\renewcommand{\arraystretch}{1.04}
\caption{System costs, accuracy, and parameter accounting.}\label{tab:completed_cost}\label{tab:reported_KS03}\label{tab:view_protocol_complete}\label{tab:cost_full_complete}

\textbf{A. System costs.}\par\smallskip
\begin{tabularx}{\linewidth}{@{}Lrrrrrr@{}}
\toprule
Method & 3D/2D passes & Params. M & GFLOPs & GiB & min/epoch & GPU h\\
\midrule
Global & 2/1 & 71.84 & 13.1 & 9.72 & 40.3 & 24.6\\
MM-Point & 2/4 & 73.10 & 21.8 & 12.80 & 59.8 & 36.9\\
MS-Point & 2/1 & 73.37 & 15.2 & 10.03 & 50.0 & 30.5\\
Point-MAE & 1/0 & 22.1 & 17.9 & 11.40 & 105.0 & 72.0\\
Inter-MAE & 1/0 & 29.6 & 19.6 & 12.10 & 111.0 & 76.5\\
\bottomrule
\end{tabularx}
\par\smallskip\begin{minipage}[t]{.485\linewidth}\vspace{0pt}\centering\footnotesize\setlength{\tabcolsep}{2pt}
\textbf{B. Candidate views and classification for the three shared systems.}\par\smallskip
\begin{tabularx}{\linewidth}{@{}Lrrr@{}}
\toprule
Method & \makecell{Offline views\\per object} & MN40 & Scan\\
\midrule
Global & -- & 92.8 & 87.9\\
MM-Point & 24 & 92.5 & 87.9\\
MS-Point & 2 & 93.6 & 89.6\\
\bottomrule
\end{tabularx}
\end{minipage}\hfill\begin{minipage}[t]{.485\linewidth}\vspace{0pt}\centering\footnotesize\setlength{\tabcolsep}{2pt}
\textbf{C. Parameter accounting (M).}\par\smallskip\begin{tabularx}{\linewidth}{@{}Lrr@{}}
\toprule
\makecell{Added\\component} & \makecell{Cumulative\\(M)} & \makecell{Increment\\(M)}\\
\midrule
Global & 71.84 & --\\
Adaptors & 72.38 & 0.54\\
Branch heads & 73.06 & 0.68\\
Gates + fusion & 73.37 & 0.31\\
\bottomrule
\end{tabularx}
\end{minipage}\par\smallskip{\footnotesize Panel A counts encoded point and image views per object per iteration. Panel B counts offline candidate renderings per object.}
\end{table}

\begin{table}[!htbp]\centering\footnotesize\setlength{\tabcolsep}{2pt}\renewcommand{\arraystretch}{1.08}
\caption{Five-seed full-training summaries and paired comparisons.}
\label{tab:completed_seeds}\label{tab:paired_complete}
\begin{minipage}[t]{.485\linewidth}\vspace{0pt}\textbf{A. Marginal mean and standard deviation.}\par\smallskip
\begin{tabularx}{\linewidth}{@{}Lrrr@{}}
\toprule
Metric & Global & MM-Point & MS-Point\\
\midrule
$n$ & 5 & 5 & 5\\
MN40 & $92.80\pm0.18$ & $92.50\pm0.20$ & $93.60\pm0.12$\\
Scan & $87.90\pm0.32$ & $87.90\pm0.35$ & $89.60\pm0.24$\\
Part & $85.45\pm0.16$ & $85.30\pm0.18$ & $86.10\pm0.13$\\
S3DIS & $66.95\pm0.27$ & $66.80\pm0.29$ & $68.20\pm0.22$\\
\bottomrule
\end{tabularx}
\end{minipage}\hfill\begin{minipage}[t]{.485\linewidth}\vspace{0pt}\textbf{B. Paired differences and 95\% confidence intervals (percentage points).}\par\smallskip
\begin{tabularx}{\linewidth}{@{}Lcc@{}}
\toprule
MS-Point minus & MN40 $\Delta$ [CI] & Scan $\Delta$ [CI]\\
\midrule
Global & \makecell{$+0.80$\\$[+0.62,+0.98]$} & \makecell{$+1.70$\\$[+1.38,+2.02]$}\\
MM-Point & \makecell{$+1.10$\\$[+0.85,+1.35]$} & \makecell{$+1.70$\\$[+1.34,+2.06]$}\\
\bottomrule
\end{tabularx}
\end{minipage}\par\smallskip{\footnotesize Five pretraining seeds per method. Part: instance mIoU; S3DIS: Area-5 mIoU. \RevisionBlack{Paired intervals summarize differences across matched runs.}}
\end{table}

Relative to the one-image global baseline, MS-Point adds 2.13\% parameters and 23.98\% GPU time while improving both classification metrics. Relative to four-image MM-Point, it uses 17.34\% less GPU time and 21.64\% less peak memory. Reusing intermediate maps avoids branch-specific backbone passes.

\subsection{Local Alignment and Low-Label Outdoor Transfer}
The completed local comparison reports ShapeNetPart instance mIoU of 86.1, 86.3, and 86.8 for object-level, pooled multi-token, and geometry-preserving alignment, respectively. S3DIS mIoU is 67.5, 67.9, and 69.0, with relative times 1.00, 1.01, and 1.18 (Table~\ref{tab:reported_K06}). The object-level baseline here is 67.5 mIoU. The 13-class S3DIS benchmark in Table~\ref{tab:s3dis} gives 59.8 mIoU.

For outdoor evaluation, we use a scene-level MS-Point instance with MinkUNet-34C and the SemanticKITTI training sequences 00--07 and 09--10; sequence 08 supplies validation mIoU over 19 classes~\citep{behley2019semantickitti}. \RevisionBlack{The structured objective supervises intermediate voxel features in this scene-level encoder.}
\begin{table}[!htbp]
\centering
\begin{minipage}{0.74\linewidth}
\centering\footnotesize
\setlength{\tabcolsep}{3pt}
\caption{SemanticKITTI sequence-08 mIoU by label budget. \RevisionBlack{A dash indicates an unavailable result.}}
\label{tab:completed_outdoor}
\begin{tabularx}{\linewidth}{@{}Lrrrr@{}}
\toprule
Pretraining / initialization & 1\% & 5\% & 10\% & 100\%\\
\midrule
Random initialization & 34.2 & 48.8 & 53.7 & 63.4\\
Scene-only SSL & 40.8 & -- & -- & --\\
Global MS-Point & 44.6 & 53.9 & 58.1 & 64.7\\
MS-Point + local & 47.3 & 55.9 & 59.8 & 65.2\\
\bottomrule
\end{tabularx}
\end{minipage}
\end{table}
The local instance improves over random initialization by 13.1 points at the 1\% budget and 1.8 points at full labels. This pattern supports label-efficient representation reuse in this scene setting.  

\section{Controlled Experiments and Further Analysis}
\label{sec:additional_evidence}

We use benchmark evaluations, full-training controls, and short-run pilots to test complementary aspects of the method. Appendix~\ref{sec:expanded_results} gives the full-training comparisons. Here we examine which factors each control isolates, how hierarchy and responses interact, and how training budget changes the evidence. Appendices~\ref{app:reported_observations} and~\ref{app:expanded_protocols} provide the corresponding configurations and task results.

\subsection{Parameter-Matched Paired Runs}
The ten-seed experiment in Table~\ref{tab:paired_pilot} compares two implementations with exactly 73,367,890 trainable parameters. Both branches, projection heads, adaptors, gates, and losses remain active in the duplicated-feature control. The mean terminal ModelNet40 SVM difference is \(+0.1337\) percentage points, with a 95\% paired interval of \([-0.1334,+0.4008]\).

\RevisionRed{The duplicated-feature control supplies unfiltered source features to both the point and image response paths. The paired-response configuration enables point-neighborhood averages and deviations together with image averages and residuals. The heads and losses remain active in both configurations, so this experiment evaluates the joint multimodal response design.}

\RevisionBlack{In this five-epoch study, mean runtime increases from 442.78 to 551.73 seconds (24.6\%), while peak memory is 10.026 and 10.028 GiB. The full-training controls in Table~\ref{tab:completed_compute} additionally match GFLOPs and have epoch times of 50.0--50.3 minutes.}

\subsection{Hierarchy--Response Interaction}
For binary hierarchy and response factors, let \(Y_{ab}\) be the metric when hierarchy is \(a\in\{0,1\}\) and responses are \(b\in\{0,1\}\). We define
\begin{equation}
 \Delta_{\mathrm{hier}|0}=Y_{10}-Y_{00},\quad
 \Delta_{\mathrm{resp}|0}=Y_{01}-Y_{00},\quad
 \delta_{\mathrm{int}}=Y_{11}-Y_{10}-Y_{01}+Y_{00}.
\end{equation}
\RevisionBlack{The combined hierarchy--response configuration exceeds either single component in Table~\ref{tab:reported_NS02}. The interaction contrasts are \(\delta_{\mathrm{int}}=-0.1\) on ModelNet40 and \(-0.5\) on ScanObjectNN, indicating overlapping gains. Fusion provides a further improvement.}

\RevisionRed{Table~\ref{tab:reported_NS02} lists the parameter count and runtime of each configuration. These vary across the four cells; Table~\ref{tab:completed_compute} provides the capacity- and computation-matched neighborhood comparison.}

\begin{figure}[!tbp]
\centering
\begin{minipage}{0.88\linewidth}
\centering
\includegraphics[width=\linewidth]{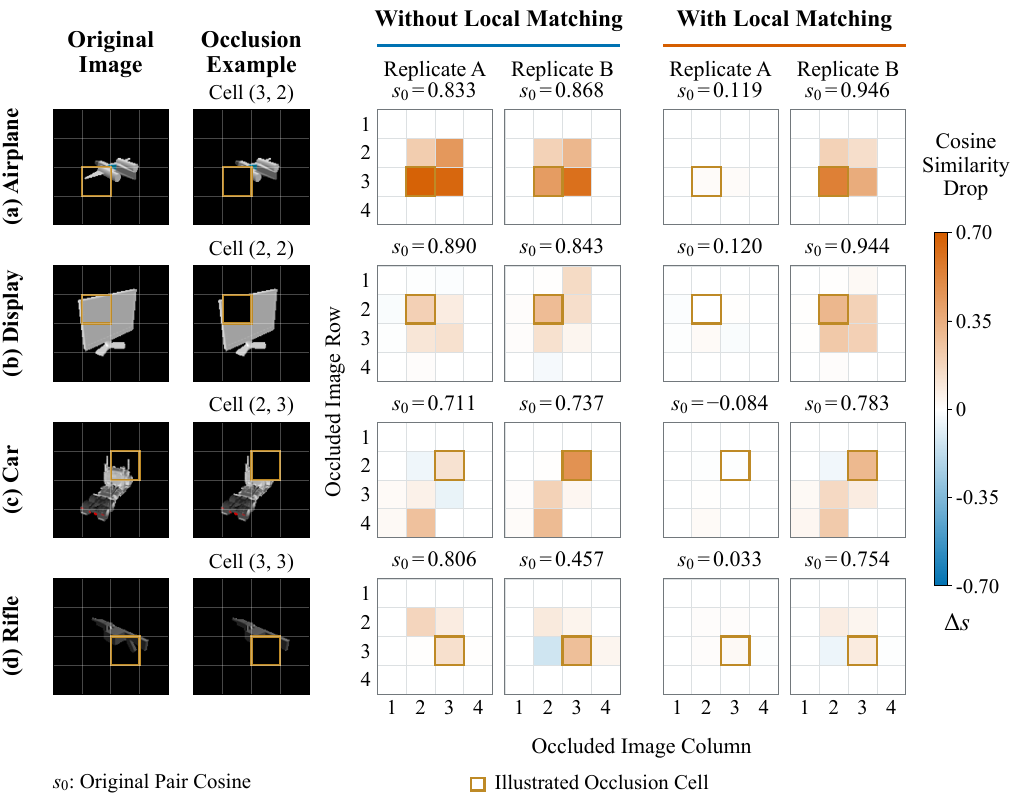}
\caption{\textbf{Spatial sensitivity to image occlusion.} One $56\times56$ cell of a $224\times224$ image is replaced by border-median RGB while the point cloud is fixed. Each heatmap shows the cosine-similarity drop $\Delta s=s_0-s_{\rm occ}$ across a $4\times4$ grid; $s_0$ is printed above it. Four objects and two training replicates compare configurations with and without local matching. Orange outlines identify the displayed intervention.}
\label{fig:occlusion_sensitivity}
\end{minipage}
\end{figure}

\begin{figure}[!tbp]
\centering
\begin{minipage}{0.80\linewidth}
\centering
\includegraphics[width=\linewidth]{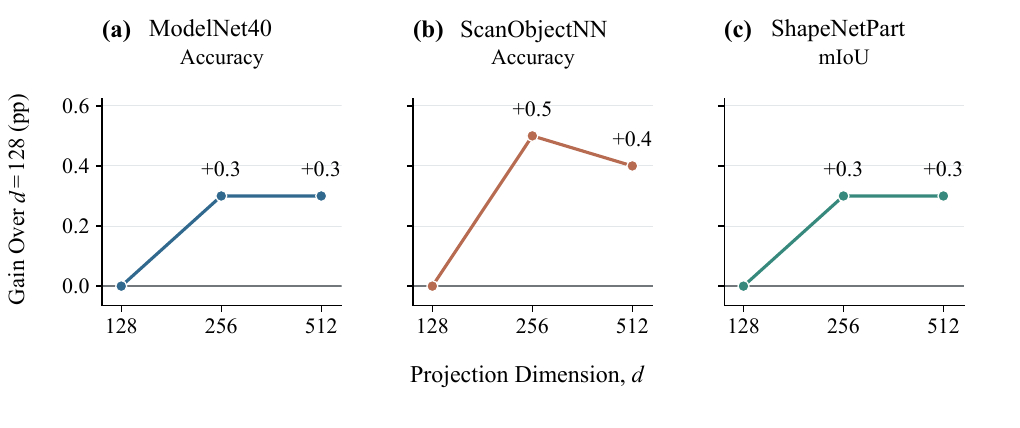}
\caption{\textbf{Projection-width sensitivity.} Gains are percentage-point differences from $d=128$ for each task (Table~\ref{tab:reported_GS05}). At the displayed precision, increasing width from 256 to 512 leaves ModelNet40 and ShapeNetPart unchanged and reduces ScanObjectNN by 0.1 point. Lines connect the three evaluated widths.}
\label{fig:projection_sensitivity}
\end{minipage}
\end{figure}

\subsection{Graph Structure, Shuffling, and Response Ablations}
The graph controls in Table~\ref{tab:reported_NS03} are designed to distinguish useful neighborhood structure from merely duplicating heads. Their interpretation depends on the exact intervention. Let \(\Pi\) be a node permutation. Jointly relabeling features and graph gives
\begin{equation}
 (\Pi P\Pi^\top)(\Pi H)=\Pi(PH).
 \label{eq:permutation_control}
\end{equation}
\RevisionBlack{Jointly permuting the graph and features preserves their association and can leave a permutation-invariant descriptor unchanged. Holding \(P\) fixed while shuffling \(H\) instead changes the graph--feature association, providing a control for neighborhood structure.}

\RevisionRed{Table~\ref{tab:reported_NS03} reports the matched neighborhood controls, Table~\ref{tab:reported_KS05} the progressive supervision study, and Table~\ref{tab:reported_GS03} the response-pathway study. Their single-response rows are interpreted against the baseline within each table. The matched-capacity conclusion uses the duplicated, random-graph, shuffled-feature, and paired-response rows of Table~\ref{tab:reported_NS03}.}

\begin{figure}[!htbp]
\centering
\includegraphics[width=1.0\linewidth,height=.78\textheight,keepaspectratio]{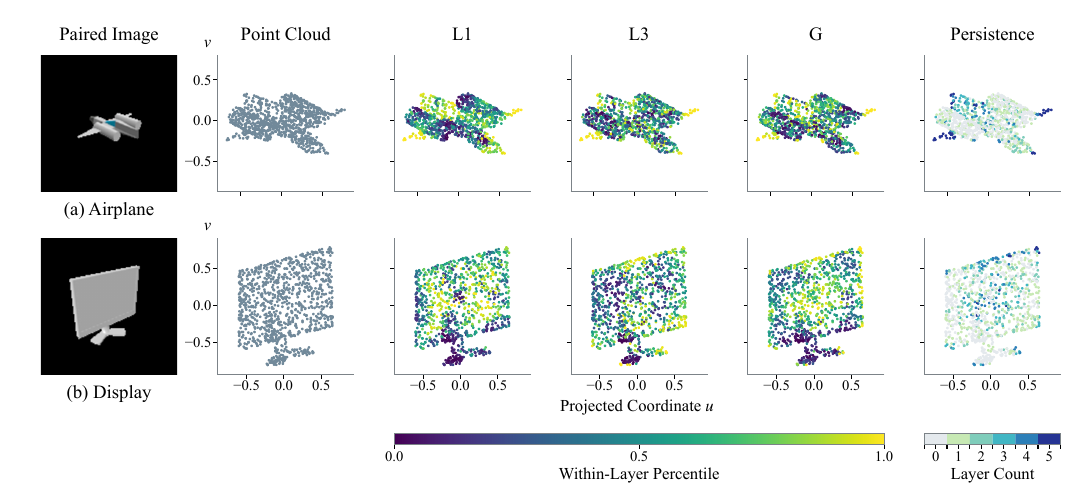}
\caption{\textbf{Point-response ranks and persistence.} Airplane and display examples from the nine-branch local-matching model. Colors in L1, L3 and G are within-layer percentiles of feature-vector norms. Persistence counts how many of L1--L4 and G reach percentile $\geq0.8$. G is the pre-pooling feature map. Each object uses a fixed projection; paired RGB views identify the object.}
\label{fig:v8_response_persistence_airplane_display}
\end{figure}

\paragraph{Response persistence across depth.}
Figure~\ref{fig:v8_response_persistence_airplane_display} follows the same points through five encoder stages. Some regions remain highly ranked across stages, while others change their relative response with depth.

\section{Extensions of the Structured Objective}
\label{sec:extensions}

The objective requires intermediate node or token features, a neighborhood relation, and image targets. This interface supports alternative encoders, learned responses, and local alignment. Appendix~\ref{sec:expanded_results} reports the completed backbone and scene studies; the following formulation explains their common interface. Appendix~\ref{app:token_visibility} gives an explicit token algorithm, distinguishing general design choices from backbone-specific configurations.

\begin{figure}[!tbp]
\begin{minipage}[t]{0.485\linewidth}
\centering
\includegraphics[width=\linewidth]{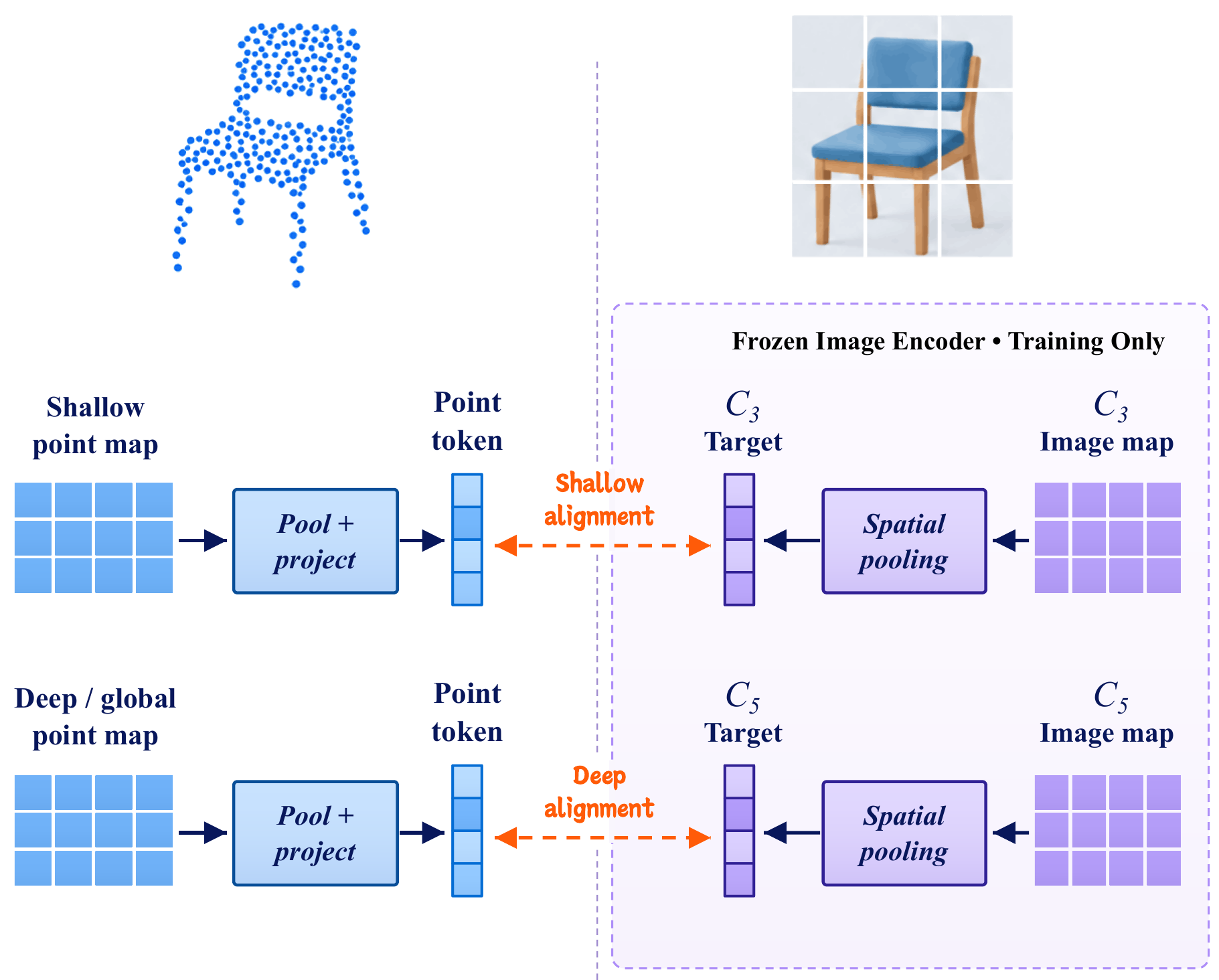}
\caption{\textbf{Frozen visual-target variant.} Shallow and deep point descriptors use C3 and C5 targets. The illustrated frozen image encoder is a training-time teacher configuration; the stage interface also supports a trainable image encoder.}
\label{fig:frozen_target_variant}
\end{minipage}\hfill
\begin{minipage}[t]{0.485\linewidth}
\centering
\includegraphics[width=\linewidth]{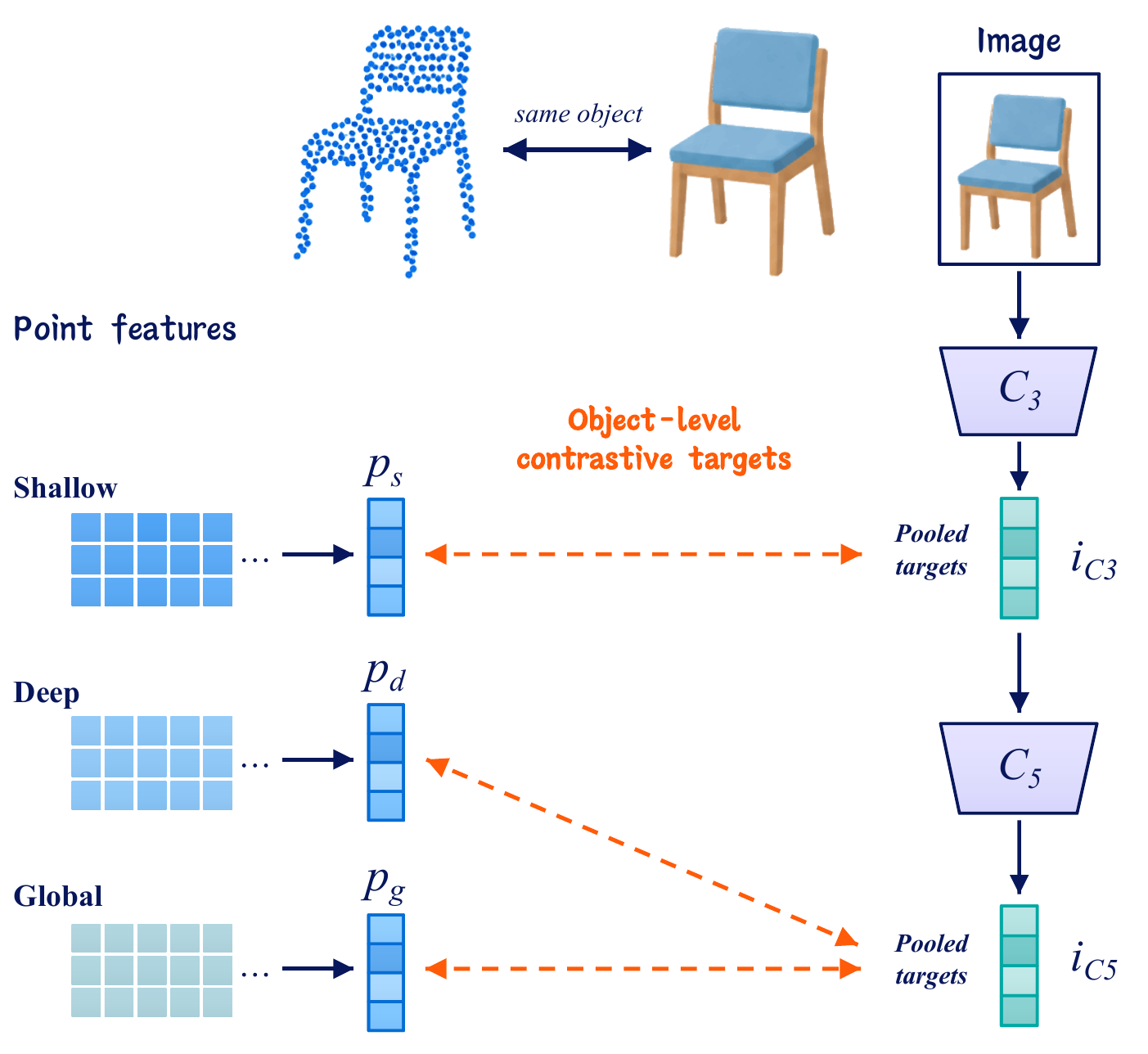}
\caption{\textbf{Compact target routing.} A shared image hierarchy supplies C3 to the shallow point branch and C5 to the deep and global branches. Each target link compares object-level descriptors from the same object.}
\label{fig:compact_target_map}
\end{minipage}
\end{figure}

Figures~\ref{fig:frozen_target_variant} and~\ref{fig:compact_target_map} illustrate two aspects of target routing: the image encoder may supply frozen teacher features, and one deep image stage may serve multiple point states through independent projections. The frozen-teacher diagram describes an interface option rather than a change to every experiment.

\subsection{Implemented Learnable and Multi-Response Filters}

\begin{figure}[!tbp]
\centering
\begin{minipage}{0.82\linewidth}
\centering
\includegraphics[width=\linewidth]{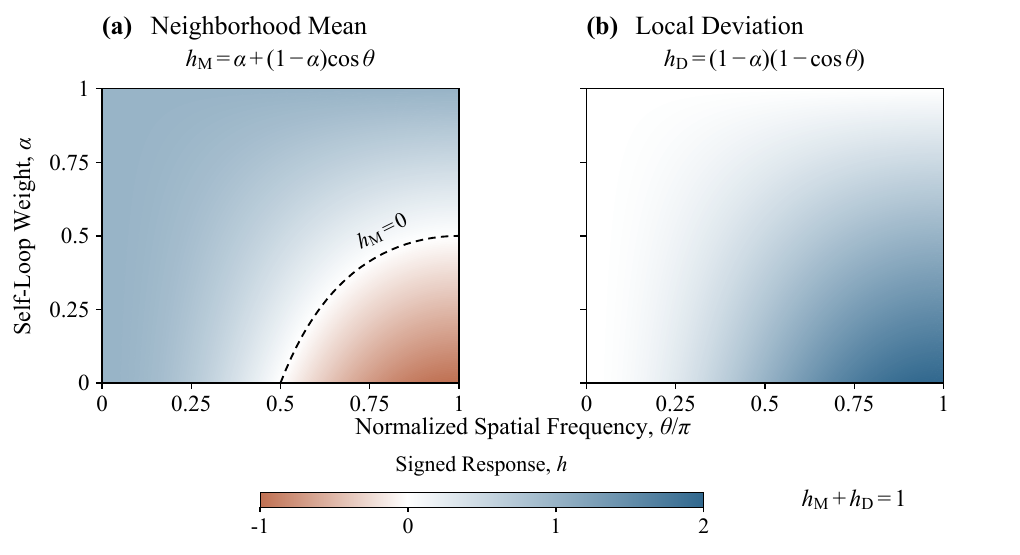}
\caption{\textbf{Self-weight controls the complementary responses.} On a one-dimensional symmetric nearest-neighbor reference, the propagation response is $h_M=\alpha+(1-\alpha)\cos\theta$ and the deviation response is $h_D=(1-\alpha)(1-\cos\theta)$. Colors encode signed gain; the dashed curve marks $h_M=0$. Their sum is one throughout.}
\label{fig:self_weight_responses}
\end{minipage}
\end{figure}

Spectral point-cloud networks can use graph filters within the backbone to compute features at inference. MS-Point uses first-order graph filters to compute point responses for auxiliary visual supervision during pretraining. Downstream tasks that use the point backbone alone require no additional filtering pass.
\RevisionBlack{The learnable first-order extension} learns one self-weight \(\alpha=\sigma(a)\) per source and modality:
\begin{equation}
 H_{\mathrm{s},\alpha}=\alpha H+(1-\alpha)PH,
 \qquad H_{\mathrm{d},\alpha}=H-H_{\mathrm{s},\alpha}.
 \label{eq:learned_first}
\end{equation}
The initial self-weight is 0.1 \RevisionBlack{in the implementation}. With three sources in each modality, the variant adds six scalar parameters. Under the symmetric construction and degree-coordinate convention, its signed responses are \(1-(1-\alpha)\lambda\) and \((1-\alpha)\lambda\). Learning \(\alpha\) changes the balance between self and neighborhood information; it does not introduce a sharp band boundary. As \(\alpha\to1\), the deviation tends to zero, \RevisionBlack{making the branch norm a useful diagnostic of this limit.}
Figure~\ref{fig:self_weight_responses} visualizes this self/neighborhood trade-off on a symmetric reference graph.

The polynomial variant uses a convex combination
\begin{equation}
 H_{\mathrm{s},\beta}=\beta_0H+\beta_1PH+\beta_2P^2H,
 \quad \beta=\operatorname{softmax}(b),
 \quad H_{\mathrm{d},\beta}=H-H_{\mathrm{s},\beta}.
 \label{eq:learned_polynomial}
\end{equation}
Three coefficients per source and modality add 18 parameters. The graph indices are fixed within one forward application, so the second propagation is genuinely \(P^2H\). Rebuilding a feature graph after the first propagation would instead give \(P_1(P_0H)\), which is not a polynomial of one fixed shift. \RevisionRed{The coefficient logits are initialized to \((-2,0,2)\), which defines a nonuniform initial mixture.}

The three-response extension is a telescoping bank:
\begin{equation}
 H_0=P^2H,\qquad H_1=PH-P^2H,\qquad H_2=H-PH,
 \qquad H_0+H_1+H_2=H.
 \label{eq:three_response}
\end{equation}
In symmetric spectral coordinates, its responses are \((1-\lambda)^2\), \(\lambda(1-\lambda)\), and \(\lambda\). The three filters overlap in frequency. With full training, the first-order pair uses 14.2M fewer parameters and scores 89.6\% versus 89.4\% on ScanObjectNN (Table~\ref{tab:reported_K05}B).

\begin{figure}[!tbp]
\centering
\begin{minipage}{0.82\linewidth}
\centering
\includegraphics[width=\linewidth]{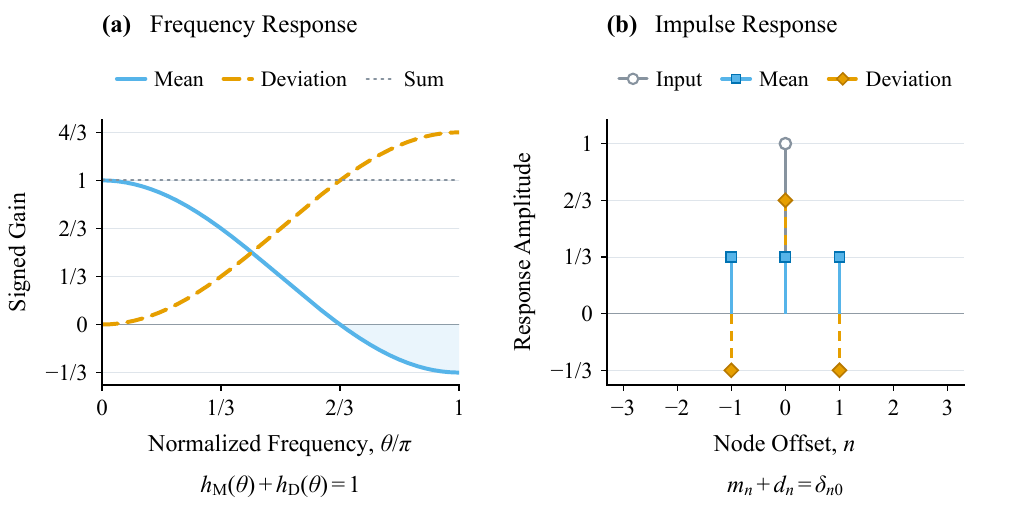}
\caption{\textbf{Frequency and impulse responses of a three-tap reference.} Uniform averaging over offsets $-1,0,1$ gives $h_M(\theta)=(1+2\cos\theta)/3$ and $h_D=1-h_M$. Their impulse responses sum to the input impulse. The negative high-frequency gain of the mean response illustrates the distinction between signed complementarity and disjoint frequency bands.}
\label{fig:reference_impulse}
\end{minipage}
\end{figure}

\begin{figure}[!htbp]
\centering
\includegraphics[width=1.0\linewidth,height=.78\textheight,keepaspectratio]{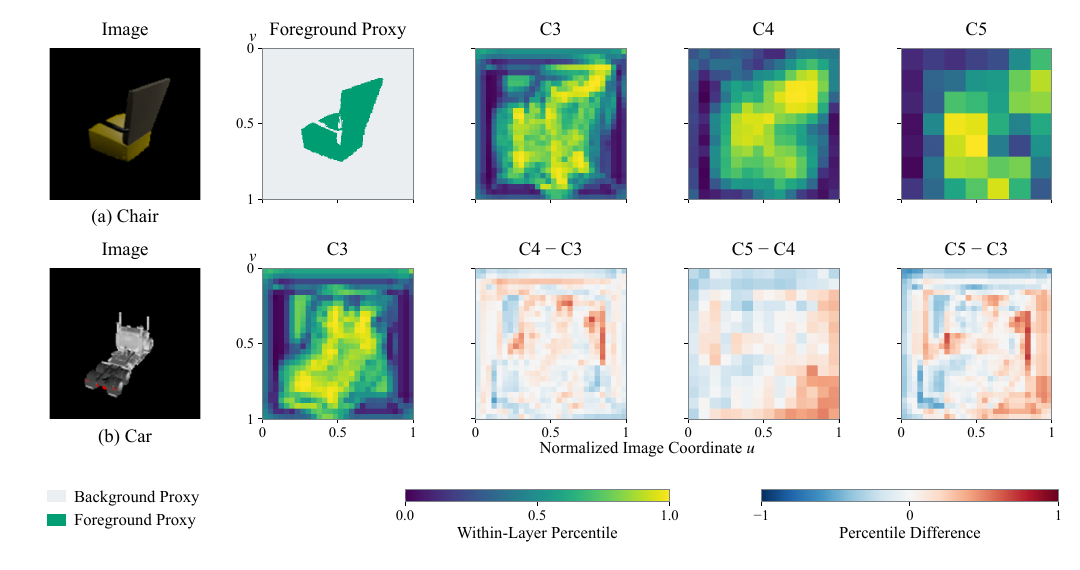}
\caption{\textbf{Image responses across stages.} Chair and car from the nine-branch local-matching model. Ranks are computed separately on each native feature grid and expanded by nearest neighbors for display. The chair foreground proxy uses RGB distance from the border-median color (threshold 0.12). Car differences subtract the displayed stage ranks. Percentiles and differences use separate fixed scales.}
\label{fig:v8_image_response_chair_car}
\end{figure}

\paragraph{Spatial responses of the image hierarchy.}
Figure~\ref{fig:v8_image_response_chair_car} shows how image-feature norm ranks vary across stages. The chair panels compare responses with an RGB-based foreground proxy; the car panels show where the ranks increase or decrease between stages.

\subsection{Scene-Level and Low-Label Transfer}
A sparse-voxel backbone can expose intermediate voxel features and spatial centers for the same structured objective. \RevisionBlack{This scene-level instantiation uses a scene-specific encoder. Its training setup specifies the initialization and transfer path: structured SSL from scratch, distillation from a point encoder, or transfer of compatible weights.}

For SemanticKITTI, the standard single-scan setting separates training sequences 00--07 and 09--10 from validation sequence 08 and evaluates 19 semantic classes~\citep{behley2019semantickitti}. \RevisionBlack{At each label budget (1\%, 5\%, and 10\%), all initializations use the same sampled LiDAR frames. We report single-scan validation mIoU on sequence 08.}

Pretraining camera use, downstream camera use, and deployed camera use are distinct fields. A point-only deployed encoder may still benefit from privileged visual pretraining information. Table~\ref{tab:completed_outdoor} reports the outdoor segmentation results.

\section{Evaluation, Efficiency, and Reproducibility}
\label{sec:evaluation_design}

\subsection{Comparable Evaluation Protocols}
A comparable result requires the data, model, and evaluator to be specified together. We represent these choices by a protocol tuple
\begin{equation}
 \mathcal P=(D_{\mathrm{pre}},T_{\mathrm{prior}},E,N,D_{\mathrm{down}},s,r,v,c),
\end{equation}
where the entries denote pretraining data, prior teacher information, encoder and initialization, input point count, downstream dataset, split, feature readout, voting policy, and checkpoint/hyperparameter selection rule. Reporting \(\mathcal P\) makes clear which contrasts isolate an objective and which compare complete systems.

Frozen linear evaluation and full fine-tuning answer different questions. The former tests the selected feature readout with the encoder fixed. The latter also measures how initialization interacts with downstream optimization. \RevisionBlack{ScanObjectNN comparisons require a common split because variants differ in background and perturbations. ShapeNetPart results specify class or instance mIoU, and S3DIS results specify Area-5 or cross-validation.}

We separate the DGCNN benchmark and local-alignment configurations. The DGCNN benchmark yields ShapeNetPart mIoU 86.0 and S3DIS mIoU 59.8. The local-alignment observations use a baseline of 86.1 instance mIoU and 67.5 S3DIS mIoU. \RevisionBlack{Gains are computed against the baseline from the same split, model, and fine-tuning setting.}

\begin{figure}[!tbp]
\centering
\begin{minipage}{0.86\linewidth}
\centering
\includegraphics[width=\linewidth]{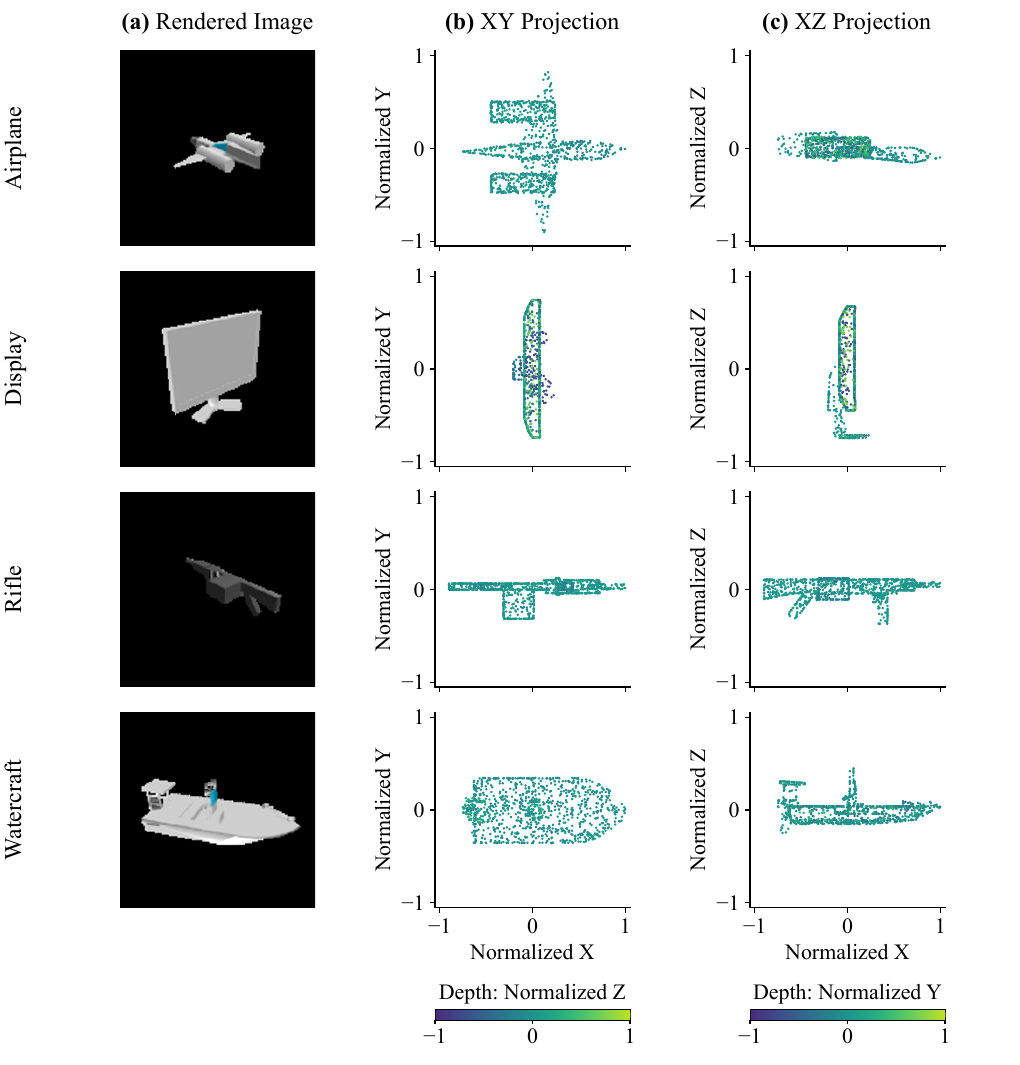}
\caption{\textbf{Paired inputs and geometric projections.} Each row shows a fixed ShapeNet rendering and XY/XZ projections of the same object's 1,024 points. Coordinates are centered and normalized to the unit sphere. Color encodes omitted Z or Y depth on $[-1,1]$. Pairing is by object identity; the orthographic views are not calibrated to the rendering camera.}
\label{fig:paired_input_geometry}
\end{minipage}
\end{figure}

\begin{figure}[!tbp]
\centering
\begin{minipage}{0.86\linewidth}
\centering
\includegraphics[width=\linewidth]{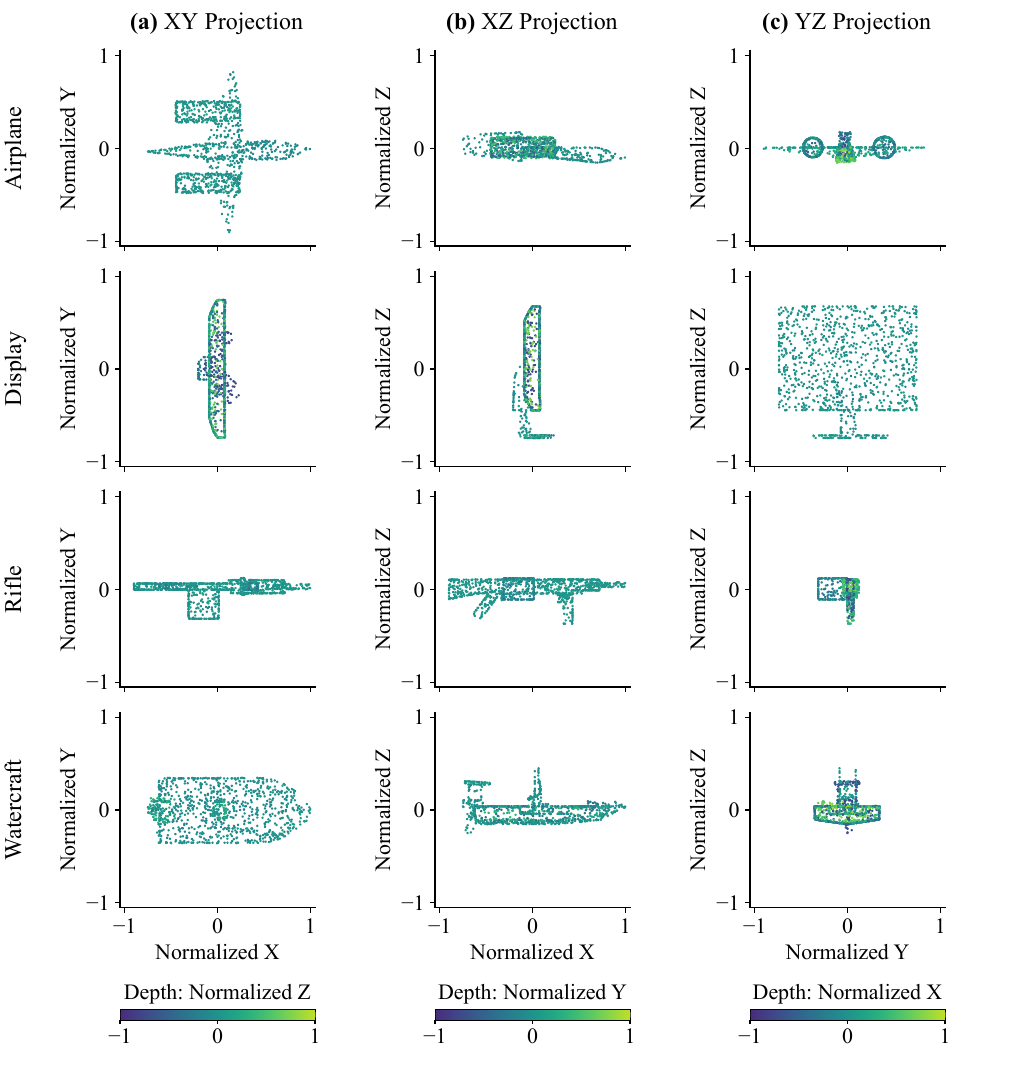}
\caption{\textbf{View-dependent geometric overlap.} XY, XZ, and YZ views project the same 1,024 points; color encodes the omitted Z, Y, and X coordinate, respectively. Equal axis scales expose changes in apparent extent and overlap across projections. These geometric views motivate explicit visibility and camera information for local matching.}
\label{fig:orthographic_geometry}
\end{minipage}
\end{figure}

\paragraph{Reading part-segmentation outputs.}
\RevisionRed{Figure~\ref{fig:v8_part_segmentation_airplane_chair_earphone} uses the DGCNN part-segmentation model fine-tuned from MS-Point pretraining. For airplane, chair and earphone, we select the first two examples in file and row order, using 2,048 points per object. Part colors are category-specific. Entropy uses the original 50-class softmax, normalized by $\log 50$; all columns share the same points and view.}

\subsection{Cost Accounting at Two Reference Points}
Structured supervision adds graph operations, projections, fusion, and batch contrastive similarities. With \(V_p\) point views and \(V_i\) encoded images, a useful accounting decomposition is
\begin{align}
 C_{\mathrm{step}}={}&V_pC_{\mathrm{point}}+V_iC_{\mathrm{image}}+C_{\mathrm{graph}}
 +C_{\mathrm{response}}+C_{\mathrm{heads}}+C_{\mathrm{contrast}},\\
 C_{\mathrm{response}}={}&O\!\left(\sum_{s\in\mathcal S}N_sk_sC_s\right),
 \quad C_{\mathrm{contrast}}=O(TB^2d).
\end{align}
Here \(T\) counts evaluated contrastive terms, \(B\) is batch size, and \(d\) is embedding width. Given the neighbor indices, gathering and aggregating features scale linearly with $k$. All branches share one backbone pass per input view. Computing all pairwise distances for exact graph construction costs $O(N^2)$ for $N$ nodes at a fixed feature width.

Two comparisons answer different cost questions. Against a one-image global baseline, the structured objective adds work. Against a four-image multimodal configuration, reduced image encoding can offset that work. Table~\ref{tab:completed_cost} reports 2.13\% more parameters, 16.03\% more GFLOPs, and 23.98\% more GPU hours than the global row; relative to four-image MM-Point, it reports 17.34\% fewer GPU hours and 21.64\% less peak memory. We compute these percentages from the matched-hardware resource study.

\RevisionBlack{Runtime comparisons include both point views and image encoding. Table~\ref{tab:completed_cost} reports total GPU hours and per-epoch timing for their respective measurement scopes; Appendix~\ref{app:expanded_protocols} summarizes these scopes.}

\section{Additional Implementation Details}
\label{app:implementation}
\label{sec:appendix}

\subsection{Dataset Usage and Evaluation Protocols}
\label{app:datasets}

We evaluate \method on widely used point cloud representation learning benchmarks. ShapeNet is used as the primary pretraining corpus, following prior cross-modal self-supervised protocols. ModelNet40 and ModelNet10 are used for clean object-level linear classification. ScanObjectNN is used to evaluate robustness under real-world scanning noise, partiality, clutter, and background contamination. ShapeNetPart and S3DIS are used for dense prediction transfer.

Figure~\ref{fig:paired_input_geometry} illustrates the object-paired inputs. Geometric depth colors describe the input coordinates, whereas Figure~\ref{fig:depth_activations} visualizes learned responses.

\begin{table}[!tbp]\centering\small\setlength{\tabcolsep}{3pt}\renewcommand{\arraystretch}{1.04}
\caption{Data protocols and default training configuration.}\label{tab:app_dataset_protocol}\label{tab:app_training_config}

\begin{minipage}[t]{.485\linewidth}\vspace{0pt}\centering\footnotesize\setlength{\tabcolsep}{2pt}
\textbf{A. Data and evaluation.}\par\smallskip\begin{tabularx}{\linewidth}{@{}LL@{}}
\toprule
Dataset & Use / evaluation\\\midrule
ShapeNet~\citep{chang2015shapenet} & SSL 2D--3D pretraining; two point views + rendered image; pretraining loss\\
ModelNet40 / ModelNet10~\citep{wu20153d} & Downstream object classification; frozen backbone + linear SVM; accuracy\\
ScanObjectNN~\citep{uy2019revisiting} & Real-world object classification / ablation; frozen backbone + linear SVM; accuracy\\
ShapeNetPart~\citep{yi2016scalable} & Downstream part segmentation; fine-tuning; OA / mIoU\\
S3DIS~\citep{armeni20163d} & Downstream indoor semantic segmentation; fine-tuning; OA / mIoU\\
\bottomrule\end{tabularx}
\end{minipage}\hfill\begin{minipage}[t]{.485\linewidth}\vspace{0pt}\centering\footnotesize\setlength{\tabcolsep}{2pt}
\textbf{B. Training defaults.}\par\smallskip\begin{tabularx}{\linewidth}{@{}LL@{}}
\toprule
Setting & Default\\\midrule
Points & 1024\\
Encoded images/iteration & 1\\
Point augmentation & Scaling, rotation, translation, jitter, dropout enabled\\
Image augmentation & Random crop, normalization enabled\\
Optimizer / schedule & Adam / cosine decay\\
Temperature $\tau$ & 0.10\\
Response sources $\setS_{\mathrm{spec}}$ & $\{l_2,l_4,g\}$\\
Cross-branch pairing & Adjacent\\
$\lambda_{\mathrm{same}},\lambda_{\mathrm{level}},\lambda_{\mathrm{fused}}$ & 1.0, 1.0, 1.0\\
$\lambda_{\mathrm{cross}},\lambda_{\mathrm{band}}$ & 0.4, 0.05\\
\bottomrule\end{tabularx}
\end{minipage}
\end{table}

For linear evaluation, the pretrained point encoder is frozen and only a linear SVM classifier is trained on the extracted object-level representation. This protocol isolates the representation quality from task-specific nonlinear adaptation. For few-shot classification, we follow the common \(N\)-way \(K\)-shot evaluation protocol and report the mean and standard deviation over repeated trials. For segmentation fine-tuning, the pretrained backbone is used as initialization and then optimized with the downstream segmentation head.

\subsection{Training Configuration}
\label{app:training_config}

We optimize with Adam~\citep{kingma2014adam} and cosine learning-rate decay, using the loss weights in Table~\ref{tab:app_training_config}.

Our configuration follows the data augmentation pipeline and contrastive temperature of CrossPoint~\citep{afham2022crosspoint}. Additional settings include the source levels, branch map, and weights for cross-level coordination, fused alignment, and response diversity. \RevisionBlack{Section~\ref{sec:evaluation_design} specifies the additional protocol factors used to assess comparability.}

The nominal settings are \(\lambda_{\mathrm{cross}}=0.4\) and \(\lambda_{\mathrm{band}}=0.05\). The former weights agreement across selected levels; the latter penalizes squared cosine between projected responses. Appendix~\ref{app:hyperparameter} reports the local sweeps around these settings.

\section{Additional Ablation Studies}
\label{app:extra_ablation}

\subsection{Effect of Cross-Level Pairing Strategy}
\label{app:pairing}

The cross-level loss depends on the selected branch pairs. Adjacent pairing connects neighboring levels and related responses; all-pairs pairing adds constraints across the full family. Adjacent pairing gives the highest accuracy in Table~\ref{tab:app_pairing}.

\subsection{Effect of Structured Image-Side Supervision}
\label{app:image_ablation}

We compare a global image target, multi-stage targets, and multi-stage targets with averaging and residual responses. This tests how image-side target construction affects the full system.

\begin{table}[!tbp]\centering\small\setlength{\tabcolsep}{3pt}\renewcommand{\arraystretch}{1.04}
\caption{Branch ablations. Panels A and C use the single-image protocol in Table~\ref{tab:ablate_hier}.}\label{tab:app_pairing}\label{tab:app_image_ablation}\label{tab:app_fusion}

\begin{minipage}[t]{.32\linewidth}\vspace{0pt}\centering\footnotesize\setlength{\tabcolsep}{2pt}
\textbf{A. Pairing.}\par\smallskip\begin{tabularx}{\linewidth}{@{}Lrr@{}}
\toprule
Pairing & $\lambda_{\mathrm{cross}}$ & Acc.\\
\midrule
None & 0.0 & 84.5\\
Adjacent & 0.4 & \textbf{85.8}\\
All pairs & 0.4 & 85.2\\
All pairs & 0.8 & 84.6\\
\bottomrule
\end{tabularx}
\end{minipage}\hfill\begin{minipage}[t]{.32\linewidth}\vspace{0pt}\centering\footnotesize\setlength{\tabcolsep}{2pt}
\textbf{B. Image targets.}\par\smallskip\begin{tabularx}{\linewidth}{@{}Lccc@{}}
\toprule
Design & Stages & Pair & Acc. \\
\midrule
Global & \(\times\) & \(\times\) & 87.1 \\
Stages & \checkmark & \(\times\) & 88.4 \\
\rowcolor{gray!15}
Stages + responses & \checkmark & \checkmark & \textbf{89.6} \\
\bottomrule
\end{tabularx}
\end{minipage}\hfill\begin{minipage}[t]{.32\linewidth}\vspace{0pt}\centering\footnotesize\setlength{\tabcolsep}{2pt}
\textbf{C. Fusion.}\par\smallskip\begin{tabularx}{\linewidth}{@{}Lcc@{}}
\toprule
Fusion & Learned & Acc. \\
\midrule
No fusion branch & \(\times\) & 85.8 \\
Uniform average & \(\times\) & 86.0 \\
Concatenation + MLP & \checkmark & 86.2 \\
\rowcolor{gray!15}
Adaptive fusion & \checkmark & \textbf{86.5} \\
\bottomrule
\end{tabularx}
\end{minipage}\par\smallskip{\footnotesize \RevisionBlack{Each panel reports a separate experiment with its own settings and accuracy.}}
\end{table}

Accuracy rises from 87.1\% with a global target to 88.4\% with multi-stage targets and 89.6\% with image responses. Table~\ref{tab:main_controls} tests the image mean and residual targets at fixed branch count and capacity. ScanObjectNN accuracy increases by 0.5 points with duplicated point features and by 0.7 with point responses.

\subsection{Effect of Fusion Strategy}
\label{app:fusion_ablation}

We compare averaging, concatenation, and adaptive branch fusion. Adaptive fusion gives the highest accuracy in Table~\ref{tab:app_fusion}.

\subsection{Effect of Response Source Levels}
\label{app:spectral_levels}

Table~\ref{tab:app_spectral_levels} compares where responses are constructed. The source set \(\{l_2,l_4,g\}\) gives the highest accuracy among these choices. These sources span intermediate and deep features. Table~\ref{tab:main_controls} tests neighborhood information separately by varying branch inputs at fixed capacity.

\section{Hyperparameter Sensitivity}
\label{app:hyperparameter}

\subsection{Sensitivity to Cross-Level Coordination Weight}
\label{app:lambda_cross}

The coefficient \(\lambda_{\mathrm{cross}}\) weights agreement across selected branches relative to the other objectives. The following sweep varies this coefficient within the configuration.

\begin{table}[!tbp]\centering\small\setlength{\tabcolsep}{3pt}\renewcommand{\arraystretch}{1.04}
\caption{Local hyperparameter sensitivity.}\label{tab:app_lambda_cross}\label{tab:app_lambda_band}\label{tab:app_temperature}

\begin{minipage}[t]{.32\linewidth}\vspace{0pt}\centering\footnotesize\setlength{\tabcolsep}{2pt}
\textbf{A. $\lambda_{\rm cross}$.}\par\smallskip\begin{tabular}{lccccc}
\toprule
\(\lambda_{\mathrm{cross}}\) & 0.0 & 0.2 & 0.4 & 0.6 & 0.8 \\
\midrule
Acc. & 88.2 & 88.9 & \textbf{89.6} & 89.1 & 88.4 \\
\bottomrule
\end{tabular}
\end{minipage}\hfill\begin{minipage}[t]{.32\linewidth}\vspace{0pt}\centering\footnotesize\setlength{\tabcolsep}{2pt}
\textbf{B. $\lambda_{\rm band}$.}\par\smallskip\begin{tabular}{lccccc}
\toprule
\(\lambda_{\mathrm{band}}\) & 0.00 & 0.01 & 0.05 & 0.10 & 0.50 \\
\midrule
Acc. & 88.2 & 89.0 & \textbf{89.6} & 89.2 & 88.5 \\
\bottomrule
\end{tabular}
\end{minipage}\hfill\begin{minipage}[t]{.32\linewidth}\vspace{0pt}\centering\footnotesize\setlength{\tabcolsep}{2pt}
\textbf{C. $\tau$.}\par\smallskip\begin{tabular}{lccccc}
\toprule
\(\tau\) & 0.03 & 0.05 & 0.10 & 0.20 & 0.30 \\
\midrule
Acc. & 88.1 & 88.8 & \textbf{89.6} & 89.0 & 88.3 \\
\bottomrule
\end{tabular}
\end{minipage}\par\smallskip{\footnotesize \RevisionBlack{Each panel reports a separate local sweep.} Table~\ref{tab:coordination_weights_complete} reports a different coordination study.}
\end{table}

The highest accuracy occurs at \(\lambda_{\mathrm{cross}}=0.4\). \RevisionBlack{Representation similarity provides a separate diagnostic of how the weights affect branch specialization.}

\subsection{Sensitivity to Response-Diversity Weight}
\label{app:lambda_band}

The coefficient \(\lambda_{\mathrm{band}}\) controls the squared-cosine penalty on projected response pairs. Table~\ref{tab:app_lambda_band} reports the highest accuracy at 0.05. \RevisionBlack{The lower accuracy at larger weights favors an intermediate degree of directional separation in this experiment.}

\subsection{Sensitivity to Contrastive Temperature}
\label{app:temperature}

Lower temperatures give more weight to high-similarity negatives in NT-Xent. As temperature increases, the softmax weights become more uniform.

\RevisionBlack{Accuracy is 88.8\%, 89.6\%, and 89.0\% at $\tau=0.05$, $0.10$, and $0.20$, respectively. The default $\tau=0.10$ gives the highest accuracy in the sweep (Table~\ref{tab:app_temperature}C).}

\section{Qualitative and Diagnostic Analyses}
\label{app:qualitative}

\subsection{Branch Utilization}
\label{app:branch_utilization}

The fusion module predicts a weight for each branch and object. Table~\ref{tab:app_branch_weight} summarizes these weights over the validation set.

The average weights are distributed across the listed branch groups. \RevisionBlack{Gate weights and adaptor-output norms jointly determine each branch's contribution to the fused vector.}

\subsection{Projected Response Similarity}
\label{app:low_high_diag}

We measure average cosine similarity between projected smoothing and deviation embeddings. The table reports lower similarity as \(\loss_{\mathrm{band}}\) receives more weight, with the highest accuracy at an intermediate setting. \RevisionBlack{This diagnostic measures the directional similarity of the projected embeddings.}

\begin{table}[!tbp]\centering\small\setlength{\tabcolsep}{3pt}\renewcommand{\arraystretch}{1.04}
\caption{Fusion and response diagnostics.}\label{tab:app_branch_weight}\label{tab:app_low_high_diag}

\begin{minipage}[t]{.485\linewidth}\vspace{0pt}\centering\footnotesize\setlength{\tabcolsep}{2pt}
\textbf{A. Mean validation fusion weights.}\par\smallskip\begin{tabularx}{\linewidth}{@{}Lr@{}}
\toprule
Group & Weight\\
\midrule
Shallow & 0.18\\
Middle & 0.21\\
Deep/global & 0.24\\
Smoothing & 0.20\\
Deviation & 0.17\\
\bottomrule
\end{tabularx}
\end{minipage}\hfill\begin{minipage}[t]{.485\linewidth}\vspace{0pt}\centering\footnotesize\setlength{\tabcolsep}{2pt}
\textbf{B. Directional regularization and Scan accuracy.}\par\smallskip\begin{tabularx}{\linewidth}{@{}Lrrr@{}}
\toprule
Penalty & $\cos$ & $\cos^2$ & Acc.\\
\midrule
None & 0.61 & 0.37 & 88.2\\
Weak & 0.42 & 0.18 & 89.0\\
Default & 0.31 & 0.10 & \textbf{89.6}\\
Strong & 0.08 & 0.01 & 88.5\\
\bottomrule
\end{tabularx}
\end{minipage}
\end{table}

Default regularization attains 89.6\% accuracy at a mean cosine of 0.31; stronger regularization reduces cosine to 0.08 but lowers accuracy to 88.5\%. The result identifies a useful balance between specialization and shared semantic alignment. The active-response controls further test whether the structured inputs improve learning beyond the projection heads themselves.

\subsection{Image-Occlusion Sensitivity}
With the point cloud fixed, Figure~\ref{fig:occlusion_sensitivity} shows how masking each image region changes its similarity to the paired point cloud. Each map gives the drop $\Delta s=s_0-s_{\rm occ}$ alongside the unoccluded similarity $s_0$, so the spatial changes can be compared across runs.

\begin{figure}[!htbp]
\centering
\includegraphics[width=1.0\linewidth,height=.78\textheight,keepaspectratio]{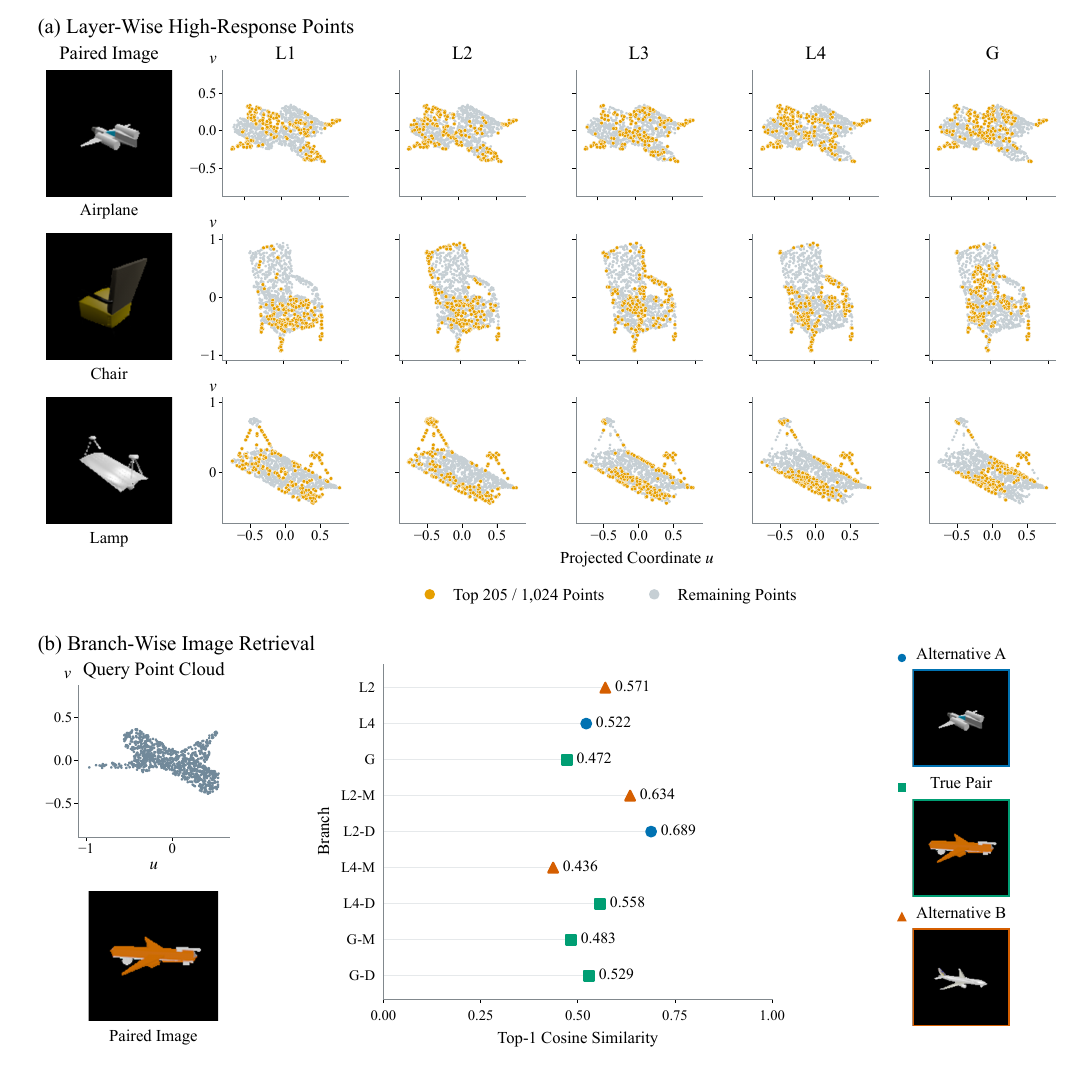}
\caption{\textbf{Layer support and branch-wise retrieval in the nine-branch local-matching model.} Top: the 205 highest-norm points among 1,024 at each stage. Bottom: each branch retrieves its highest-cosine image among 52 objects. For the query with the most distinct top matches, G, L4-D, G-M, and G-D retrieve the paired image. The other five branches retrieve two other airplanes. M and D denote consensus and deviation. The airplane in the upper row and the lower query are different instances.}
\label{fig:v8_layer_response_branch_retrieval}
\end{figure}

\paragraph{Layer support and branch retrieval.}
Figure~\ref{fig:v8_layer_response_branch_retrieval} relates layer-wise high-response regions to branch-wise image retrieval. The selected regions change across layers, and the nine branches choose three different image identities for the illustrated query.

\subsection{Representation Diagnostics: Protocols and Interpretation}
\label{app:diagnostic_protocols}

We examine where features are most active, how neighborhoods change across encoder stages, and which branches receive the greatest fusion weights. Figure~\ref{fig:depth_activations} shows feature-norm ranks for four ShapeNet objects from the \RevisionBlack{local-matching model}. The spatial pattern changes across stages in both modalities, supporting stage-specific inspection rather than reading all structure from the final descriptor. \RevisionBlack{Percentiles are computed within each stage, so colors show relative activation ranks within a feature map.}

For the remaining diagnostics, we use the same 52 in-corpus objects, with four objects from each of 13 classes and 1,024 points per object. \RevisionBlack{Two training replicates provide validation-selected models} of configurations with and without pooled local matching. These diagnostic configurations use nine branches and learned fusion in both modalities; their original pretraining uses 2,048 points. They complement the downstream evaluations with measurements of internal organization.

Figure~\ref{fig:directed_neighborhoods} examines non-self edges. From $l_4$ to the pre-pooling map, reciprocity increases and in-degree variation decreases in both configurations and runs. Nevertheless, reciprocal edges remain well below 100\%. The learned neighborhoods therefore retain directionality, which motivates the graph-shift formulation and the separate symmetric reference analysis in Section~\ref{sec:spectral}.

Local supervision also warrants an activity check. Figure~\ref{fig:local_token_activity} measures entire zero-valued token vectors in the pooled-set extension. Mean zero-token rates are 5.41\% and 13.55\% on the point side, and 5.47\% and 46.84\% on the image side, for the two runs. A zero token supplies no discriminative dot-product score. The result identifies a concrete diagnostic for local objectives: adding more targets is useful only when their descriptors remain informative. Point-bin indices are not geometric patch coordinates, and image-grid activity alone does not identify semantic foreground.

Finally, Figure~\ref{fig:gate_consistency} compares branch weights across the two runs. With local matching, point-side median Jensen--Shannon divergence falls from 0.03074 to 0.01380 nats and top-branch agreement rises from 40.4\% to 61.5\%. Image-side divergence instead increases from 0.05822 to 0.09616, with agreement declining from 38.5\% to 30.8\%. The modality-specific pattern argues for inspecting both pathways. Gate consistency describes the learned mixture, while transfer quality is established by the downstream evaluations. Appendix~\ref{app:qualitative} adds occlusion diagnostics, and Appendix~\ref{app:local_runs} details the pooled-set experiments.

\begin{figure}[!htbp]
\centering
\includegraphics[width=1.0\linewidth,height=.78\textheight,keepaspectratio]{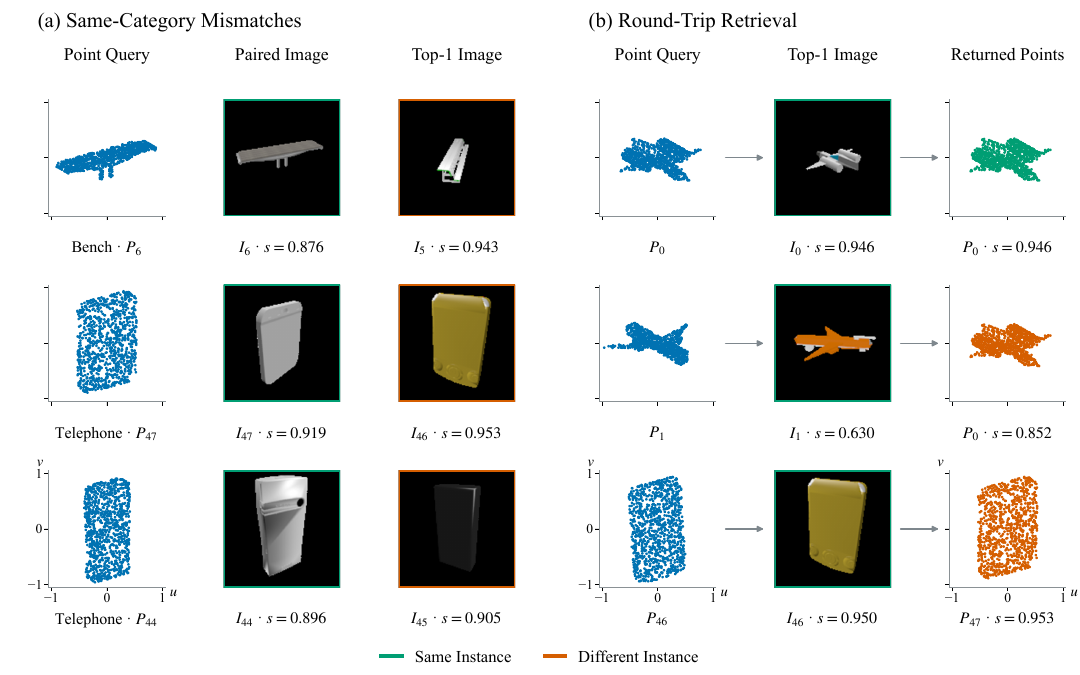}
\caption{\textbf{Instance identity in forward and round-trip retrieval.} The nine-branch local-matching model ranks 52 candidates by cosine similarity. Left: in three same-category examples, a visually similar instance outranks the paired image. The examples have the largest score gaps among such matches. Right: reversing the search returns the original point cloud in the first example and a different instance in the next two. Green marks the query instance and orange an alternative.}
\label{fig:v8_instance_identity_roundtrip}
\end{figure}

\paragraph{Instance and category identity in retrieval.}
In the three same-category examples in Figure~\ref{fig:v8_instance_identity_roundtrip}, the top-ranked image depicts another bench or telephone of the same class (cosine similarity 0.905--0.953). The round trips in the right panel test whether retrieving a paired image also returns the original point cloud when the search direction is reversed.

\section{Complementary Graph Filtering and Soft Response Decoupling}
\label{app:graph_scope}
For an undirected nonnegative graph, let $\widetilde W=W+I$, $D_{ii}=\sum_j\widetilde W_{ij}$, $S=D^{-1/2}\widetilde W D^{-1/2}$ and $L=I-S=U\Lambda U^\top$. For node-major features $H$,
\begin{equation}
 H_s=SH=U(I-\Lambda)U^\top H,\qquad H_d=(I-S)H=U\Lambda U^\top H.
\end{equation}
These first-order polynomial filters require no eigendecomposition during training~\citep{shuman2013emerging,defferrard2016convolutional}. The raw responses sum to $H$. On $[0,1]$, the consensus gain $1-\lambda$ decreases from one to zero as the deviation gain $\lambda$ increases from zero to one. Figure~\ref{fig:analytic_responses} shows both responses over the full spectrum $[0,2]$.

For nonzero projected embeddings $a,b$, the squared-cosine penalty has gradient
\begin{equation}
 u=\frac{a}{\|a\|_2},\quad v=\frac{b}{\|b\|_2},\quad c=u^\top v,
 \qquad \nabla_a c^2=\frac{2c}{\|a\|_2}(v-cu),\quad u^\top\nabla_a c^2=0.
 \label{eq:angular_gradient}
\end{equation}
It changes the relative direction to first order. The finite penalty allows both branches to retain shared semantics under visual alignment. Appendix~\ref{app:low_high_diag} reports the weight and similarity diagnostics.

For the undirected random walk $P=D^{-1}\widetilde W$, $D^{1/2}PH=S(D^{1/2}H)$. Directed feature-space neighborhoods instead use $PH$ and $(I-P)H$ as complementary graph-shift responses. A real orthogonal Fourier interpretation requires the symmetric assumptions above. Further operator properties are summarized in Appendix~\ref{app:further_theory}.

\section{Exploratory Controls and Extensions}
\label{app:pilots}

\subsection{Parameter-Matched Short Runs}
\RevisionBlack{The following pilot evaluates the implementation variant under a short warm-start training budget.} Both variants have 73,367,890 trainable parameters and use the same warm-start setting. \RevisionBlack{Ten paired training replicates} were run for five epochs with at most 2,048 training objects and batch size four. The endpoint is the terminal ModelNet40 SVM accuracy. The duplicated-feature control replaces the response inputs with two copies of the unfiltered features while retaining the heads and losses.

\begin{table}[!htbp]
\centering
\caption{\RevisionBlack{Parameter-matched five-epoch warm-start runs. Runtime is the mean duration across ten paired replicates.}}
\label{tab:paired_pilot}
\small
\begin{tabular}{lrrr}
\toprule
Variant & Accuracy (\%) & Runtime (s) & Peak memory (GiB)\\
\midrule
Duplicated-feature heads & 91.4222 & 442.78 & 10.026\\
Smoothing/deviation & 91.5559 & 551.73 & 10.028\\
\bottomrule
\end{tabular}
\end{table}

For paired differences \(d_i=a_i^{\mathrm{response}}-a_i^{\mathrm{control}}\), the mean difference is 0.1337 percentage points. The paired Student-\(t\) interval,
\begin{equation}
 \overline d\ \pm\ t_{0.975,9}\frac{s_d}{\sqrt{10}},
\end{equation}
is \([-0.1334,0.4008]\) percentage points. The two-sided paired test gives \(p=0.2867\), and an exact sign-flip test gives \(p=0.3047\). Six of ten paired differences are positive. \RevisionBlack{The response configuration increases mean runtime by 24.6\% in this five-epoch experiment.} In full pretraining at 15.2 GFLOPs (Table~\ref{tab:main_controls}), point responses add 0.7--0.9 percentage points on ScanObjectNN at fixed image-target construction. Switching on both modalities adds 1.4 points over duplicated inputs, at 50.0 versus 50.1 minutes per epoch.

\subsection{Response and View Pilots}
The response pilot uses \RevisionBlack{two training replicates}, five warm-start epochs, at most 2,048 objects, and batch size 32. Table~\ref{tab:filter_pilot} reports the best validation accuracy from each run, averaged across seeds.

The view pilot uses \RevisionBlack{three training replicates}, five warm-start epochs, at most 1,024 objects, and batch size two. Its validation results show increased runtime without a mean accuracy gain from additional encoded views (Table~\ref{tab:view_pilot}). With full pretraining, encoding a second image increases ModelNet40 accuracy to 93.7\% and ScanObjectNN accuracy to 89.8\% (Table~\ref{tab:reported_K04}).

\begin{table}[tbp]\caption{Short-run response and encoded-view pilots (separate budgets and endpoints).}\label{tab:filter_pilot}\label{tab:view_pilot}
\begin{minipage}[t]{0.485\linewidth}
\setlength{\tabcolsep}{2pt}

\centering
\textbf{A. Two-seed response pilot. Parameter counts refer to the implementation.}\par\smallskip

\footnotesize
\begin{tabularx}{\linewidth}{@{}Lrrr@{}}
\toprule
Response & Val. (\%) & Parameters & Min.\\
\midrule
Fixed pair & 90.930 & 73,367,890 & 7.36\\
Learnable & 91.082 & 73,367,896 & 7.63\\
Polynomial \(I,P,P^2\) & 91.184 & 73,367,908 & 8.58\\
Three-way & 90.955 & 87,574,488 & 8.72\\
\bottomrule
\end{tabularx}

\end{minipage}\hfill
\begin{minipage}[t]{0.485\linewidth}
\setlength{\tabcolsep}{2pt}

\centering
\textbf{B. Three-seed encoded-view pilot. The endpoint and training budget are specific to this experiment.}\par\smallskip

\footnotesize
\begin{tabular}{crr}
\toprule
Views & Val. (\%) & Sec.\\
\midrule
1 & 91.277 & 425.12\\
2 & 91.277 & 469.06\\
3 & 90.938 & 525.29\\
\bottomrule
\end{tabular}

\end{minipage}
\end{table}

The response parameterizations are specified in Section~\ref{sec:extensions}. Appendix~\ref{app:implementation_variants} distinguishes \RevisionBlack{the nine-branch implementation} from the main formulation, and Appendix~\ref{app:seed_results} gives the individual paired outcomes.

\section{Implementation Variants and Module Specifications}
\label{app:implementation_variants}
\label{app:detailed_formulation}

\par\noindent\textbf{Notation.} $X^{(s)}\in\mathbb R^{C_s\times N_s}$ is channel-first and $H^{(s)}=(X^{(s)})^\top$ is node-major. $\setL$ and $\setS_{\rm spec}$ denote base and response-source stages; $\mathcal B=\{\mathrm{smooth},\mathrm{dev}\}$ labels the two responses. $\mathcal A$ is the active branch set and $\rho$ its image-target map. $z_m^{(v)},h_m^{(u)}$ are point/image branch embeddings under views $v/u$; $z_f,h_f$ are their fused embeddings.\par

Figures~\ref{fig:eleven_branch_flow} and~\ref{fig:training_transfer_paths} separate branch computation from downstream use. The former expands the five-base/six-response construction; the latter shows how pooled and unpooled point features serve different prediction heads.

\subsection{\RevisionBlack{Nine-Branch Configuration: Correspondence and Dimensions}}
The map is \(r(l_2)=C3\), \(r(l_4)=C5\), and \(r(g)=C5\). Each base branch uses its base image target, while its smoothing/deviation branches use the corresponding image response. The two C5 consumers have independent projection heads. \RevisionBlack{The image-fusion module} fuses the projected branch family; \RevisionBlack{the active image targets in this variant are C3 and C5.}

\par\noindent\begin{minipage}[t]{.50\linewidth}\vspace{0pt}

\centering\footnotesize
\intablecaption{\RevisionBlack{Nine-branch source and projection dimensions.} Both modalities concatenate max and mean pooling.}
\label{tab:r1_dimensions}
\begin{tabularx}{\linewidth}{@{}Lrrr@{}}
\toprule
Field & $l_2$ & $l_4$ & $g_{\mathrm{map}}$\\
\midrule
Point channels & 64 & 256 & 1024\\
Point pooled dim. & 128 & 512 & 2048\\
Image stage & C3 & C5 & C5\\
Image channels & 512 & 2048 & 2048\\
Image pooled dim. & 1024 & 4096 & 4096\\
\bottomrule
\end{tabularx}

\end{minipage}\hfill\begin{minipage}[t]{.47\linewidth}\vspace{0pt}
Each projection uses \(h=\max(256,\min(d_{\mathrm{in}},1024))\) as hidden width and outputs 256 dimensions. \RevisionRed{For C5, concatenating max and mean pooling converts 2048 channels into the 4096-dimensional input to the image projection head.}
\end{minipage}\par\medskip

\par\noindent\textbf{\RevisionBlack{Nine-branch module details.}} Each point source/response has an independent bias-free Conv1d$(C,C,1)$--BN--LeakyReLU(0.2) adaptor. Independent projection heads use bias-free Linear$(d,h)$--BN--ReLU--Linear$(h,256)$. Gates use biased Linear$(256,1)$ followed by branch softmax. Separate point/image fusion heads use bias-free Linear$(256,256)$--BN--ReLU--Linear$(256,256)$. For each image feature map, we compute a $3\times3$ local mean (stride 1, one-pixel zero padding included in the average) and subtract it from the input to obtain the residual. The contrastive criterion applies $\ell_2$ normalization internally.\par

\begin{figure}[!htbp]
\centering
\includegraphics[width=1.0\linewidth,height=.78\textheight,keepaspectratio]{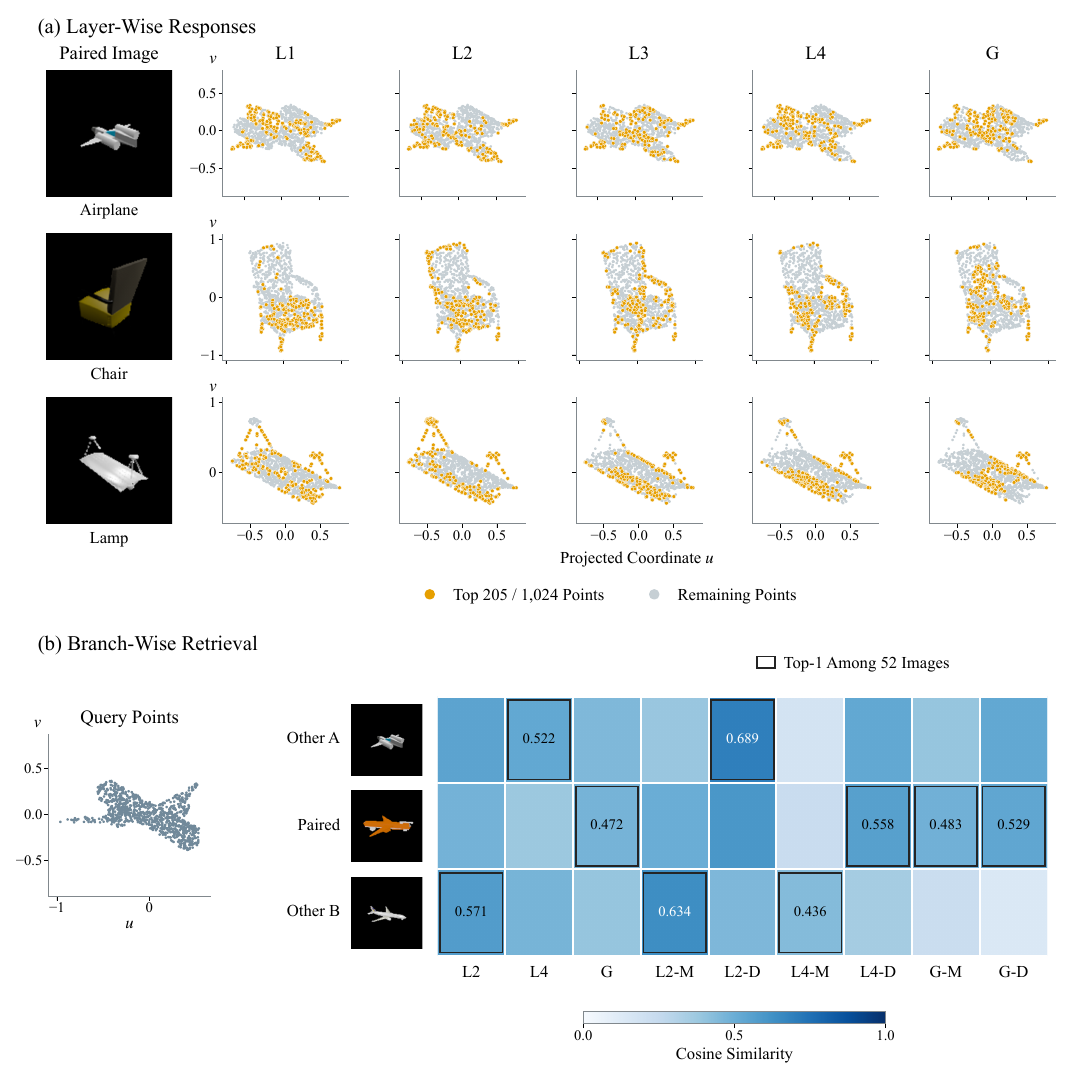}
\caption{\textbf{Matrix view of the layer and branch examples in Figure~\ref{fig:v8_layer_response_branch_retrieval}.} The upper panels repeat the same top-205 point sets. The lower matrix displays cosine scores for three image identities across nine branches of the local-matching model. Black boxes mark the winners found in the full 52-image search. The two visualizations show the same experiment.}
\label{fig:v8_layer_response_branch_retrieval_v2}
\end{figure}

\paragraph{Which image does each branch select?}
Figure~\ref{fig:v8_layer_response_branch_retrieval_v2} presents the same cases as Figure~\ref{fig:v8_layer_response_branch_retrieval} using a similarity matrix. The matrix shows the paired image alongside the two alternative identities selected by other branches.

\subsection{Exact Contrastive and Reduction Conventions}
\RevisionBlack{The symmetric criterion} uses the normalized $2B$-embedding NT-Xent form in Eq.~\eqref{eq:v11_ntxent}, excluding self logits and retaining each positive in the denominator.

\RevisionBlack{In the nine-branch variant,} same-branch terms are averaged over branches and comparisons between the first point view and each other view. Cross-level terms use the two base pairs \((l_2,l_4)\) and \((l_4,g)\), averaged over pairs and point views. Point and image prototypes are arithmetic means across their views before normalization inside the contrastive criterion. Averaging individually normalized embeddings would produce a different objective.

\RevisionRed{The diversity term averages squared cosine over source levels, point views, and batch elements, and applies to the point responses.}

\section{Supplementary Experiments and Analysis}
\label{app:reported_observations}

These studies cover hierarchy, responses, local alignment and backbone adaptation. Full-training results and pilots retain their separate budgets and endpoints.

MN40 and Scan denote classification accuracy. Part and S3DIS denote mIoU, with averaging specified where available.

\paragraph{Hierarchy, responses, and fusion}\mbox{}\par\nopagebreak[4]

\par\noindent\begin{minipage}[t]{.66\linewidth}\vspace{0pt}
\centering\footnotesize\setlength{\tabcolsep}{2pt}\renewcommand{\arraystretch}{1.08}
\intablecaption{Hierarchy, responses, fusion, and task transfer.}
\label{tab:reported_NS02}\label{tab:factorial_part_complete}
\begin{tabularx}{\linewidth}{@{}Lccrrrrr@{}}
\toprule
Config. & Hier. & Resp. & Params. (M) & Hours & MN40 & Scan & Part\\
\midrule
G & -- & -- & 71.84 & 24.6 & 92.8 & 87.9 & 85.4\\
H & $\checkmark$ & -- & 72.45 & 26.5 & 93.1 & 88.7 & 85.7\\
R & -- & $\checkmark$ & 72.76 & 27.8 & 93.2 & 88.9 & 85.8\\
H+R & $\checkmark$ & $\checkmark$ & 73.12 & 29.3 & 93.4 & 89.2 & 85.9\\
Full & $\checkmark$ & $\checkmark$ & 73.37 & 30.5 & 93.6 & 89.6 & 86.1\\
\bottomrule
\end{tabularx}
\par\smallskip{\footnotesize G: global; H: hierarchy; R: neighborhood responses. Full adds adaptive fusion to H+R. The four unfused configurations define the factorial comparison in Appendix~\ref{sec:additional_evidence}.}

\end{minipage}\hfill\begin{minipage}[t]{.31\linewidth}\vspace{0pt}
Adding hierarchy, neighborhood responses, and adaptive fusion successively raises ScanObjectNN accuracy from 87.9\% to 88.7\%, 89.2\%, and 89.6\% (+1.7 points overall). ShapeNetPart mIoU rises from 85.4\% to 86.1\%.
\end{minipage}\par\medskip

\paragraph{Neighborhood and signal controls}
The graph controls preserve the branch architecture and computation. Feature shuffling changes the graph--signal association; one-response rows use fewer active paths.

\paragraph{Progressive structural supervision}
\RevisionBlack{Table~\ref{tab:reported_KS05} evaluates progressive supervision, with each addition compared against the preceding configuration.}

\begin{table}[!tbp]\centering\small\setlength{\tabcolsep}{3pt}\renewcommand{\arraystretch}{1.04}
\begin{minipage}[t]{.485\linewidth}\vspace{0pt}\centering\footnotesize\setlength{\tabcolsep}{2pt}
\caption{Progressive structural supervision.}\label{tab:reported_KS05}
\begin{tabularx}{\linewidth}{@{}Lrr@{}}
\toprule
Supervision & MN40 & Scan\\
\midrule
Global & 92.8 & 87.9\\
Hierarchy & 93.2 & 88.7\\
Hierarchy + smooth & 93.3 & 88.9\\
Hierarchy + deviation & 93.2 & 88.8\\
Both & 93.4 & 89.2\\
Fusion full & 93.6 & 89.6\\
\bottomrule
\end{tabularx}

\end{minipage}\hfill\begin{minipage}[t]{.485\linewidth}\vspace{0pt}\centering\footnotesize\setlength{\tabcolsep}{2pt}
\caption{Loss deletion across tasks.}\label{tab:reported_NS04}\label{tab:losses_part_complete}
\begin{tabularx}{\linewidth}{@{}Lrrr@{}}
\toprule
Objective & MN40 & Scan & Part\\
\midrule
Global only & 92.8 & 87.9 & 85.4\\
Without $\mathcal L_{\rm same}$ & 93.2 & 88.9 & 85.7\\
Without $\mathcal L_{\rm cross}$ & 93.1 & 88.7 & 85.6\\
Without $\mathcal L_{\rm level}$ & 92.9 & 88.3 & 85.4\\
Without $\mathcal L_{\rm fused}$ & 93.2 & 88.9 & 85.7\\
Without $\mathcal L_{\rm band}$ & 93.3 & 89.1 & 85.8\\
Full & 93.6 & 89.6 & 86.1\\
\bottomrule
\end{tabularx}
\par\smallskip{\footnotesize Part: instance mIoU. Branch-level deletion retains fused image supervision.}

\end{minipage}
\end{table}

\paragraph{Separate response pathways}
\RevisionBlack{Table~\ref{tab:reported_GS03}A compares separate response pathways against global-only training (87.9\% on ScanObjectNN). Table~\ref{tab:main_controls} instead compares response inputs in the neighborhood-control study, including a duplicated-feature reference (88.2\%).}

\paragraph{Loss deletion}
Removing branch-level image alignment reduces MN40/Scan accuracy by 0.7/1.3 points, the largest deletion effect in this study. The fused image target remains active.

\paragraph{Full-training response variants}
\RevisionBlack{Table~\ref{tab:reported_K05} compares fixed, learnable, polynomial, and three-response variants under full training. Appendix~\ref{app:pilots} reports the short-budget study.}

\begin{table}[!tbp]\caption{Response ablations under the respective configurations.}\label{tab:reported_GS03}\label{tab:reported_K05}
\begin{minipage}[t]{0.485\linewidth}
\centering\footnotesize
\setlength{\tabcolsep}{1.7pt}
\textbf{A. Response pathways vs. global-only.}\par\smallskip

\begin{tabularx}{\linewidth}{@{}Lccrrr@{}}
\toprule
\makecell{Configuration} & Smooth & Deviation & MN40 & Scan & Part\\
\midrule
Global & -- & -- & 92.8 & 87.9 & 85.4\\
Smooth only & $\checkmark$ & -- & 93.1 & 88.6 & 85.7\\
Dev only & -- & $\checkmark$ & 93.2 & 88.9 & 85.8\\
Pair & $\checkmark$ & $\checkmark$ & 93.6 & 89.6 & 86.1\\
\bottomrule
\end{tabularx}
\end{minipage}\hfill
\begin{minipage}[t]{0.485\linewidth}
\centering\footnotesize
\setlength{\tabcolsep}{1.7pt}
\textbf{B. Full-training response variants.}\par\smallskip

\begin{tabularx}{\linewidth}{@{}Lrrrrr@{}}
\toprule
Response & MN40 & Scan & Part & Params. & Hours\\
\midrule
Fixed pair & 93.6 & 89.6 & 86.1 & 73.37M & 30.5\\
Learnable first order & 93.7 & 89.8 & 86.2 & 73.38M & 31.1\\
Second-order polynomial & 93.7 & 89.7 & 86.2 & 73.39M & 33.4\\
Three-response & 93.5 & 89.4 & 86.0 & 87.57M & 38.8\\
\bottomrule
\end{tabularx}
\end{minipage}
\end{table}

\paragraph{Encoded image-view budget}
Two encoded views add 0.1/0.2 points on MN40/Scan and 11\% time. A third adds time without increasing the mean; the short-run pilot uses a different budget.

\begin{table}[tbp]
\begin{minipage}[t]{0.485\linewidth}
\centering\footnotesize
\setlength{\tabcolsep}{3pt}
\caption{Encoded image-view budget.}
\label{tab:reported_K04}
\begin{tabularx}{\linewidth}{@{}Lrrr@{}}
\toprule
\makecell{Images/update} & MN40 & Scan & \makecell{Relative\\time}\\
\midrule
1 & 93.6 & 89.6 & 1.00\\
2 & 93.7 & 89.8 & 1.11\\
3 & 93.7 & 89.8 & 1.24\\
\bottomrule
\end{tabularx}
\end{minipage}

\end{table}

\paragraph{Global and local cross-modal targets}\mbox{}\par\nopagebreak[4]

\par\noindent\begin{minipage}[t]{.50\linewidth}\vspace{0pt}

\centering\footnotesize
\setlength{\tabcolsep}{2pt}
\intablecaption{Global and local cross-modal targets.}
\label{tab:reported_K06}
\begin{tabularx}{\linewidth}{@{}Lrrr@{}}
\toprule
Metric & Object & Pooled & Geometry\\
\midrule
MN40 & 93.6 & 93.7 & 93.9\\
Scan & 89.6 & 89.8 & 90.2\\
Part class & 84.7 & 84.9 & 85.4\\
Part instance & 86.1 & 86.3 & 86.8\\
S3DIS & 67.5 & 67.9 & 69.0\\
Relative time & 1.00 & 1.01 & 1.18\\
\bottomrule
\end{tabularx}
{\footnotesize Object: object-level; Pooled: pooled multi-token; Geometry: geometry-preserving.}

\end{minipage}\hfill\begin{minipage}[t]{.47\linewidth}\vspace{0pt}
Our geometry-preserving configuration exceeds object-level alignment by 0.7 instance mIoU points on ShapeNetPart and 1.5 mIoU points on S3DIS, with an 18\% time increase. The local-alignment study uses an object-level S3DIS baseline of 67.5 mIoU. The DGCNN benchmark in Table~\ref{tab:s3dis} gives 59.8 mIoU under its 13-class protocol.
\end{minipage}\par\medskip

Figure~\ref{fig:local_alignment_tradeoff} plots these within-study differences against additional training time.

\begin{figure}[!htbp]
\centering
\includegraphics[width=1.0\linewidth,height=.78\textheight,keepaspectratio]{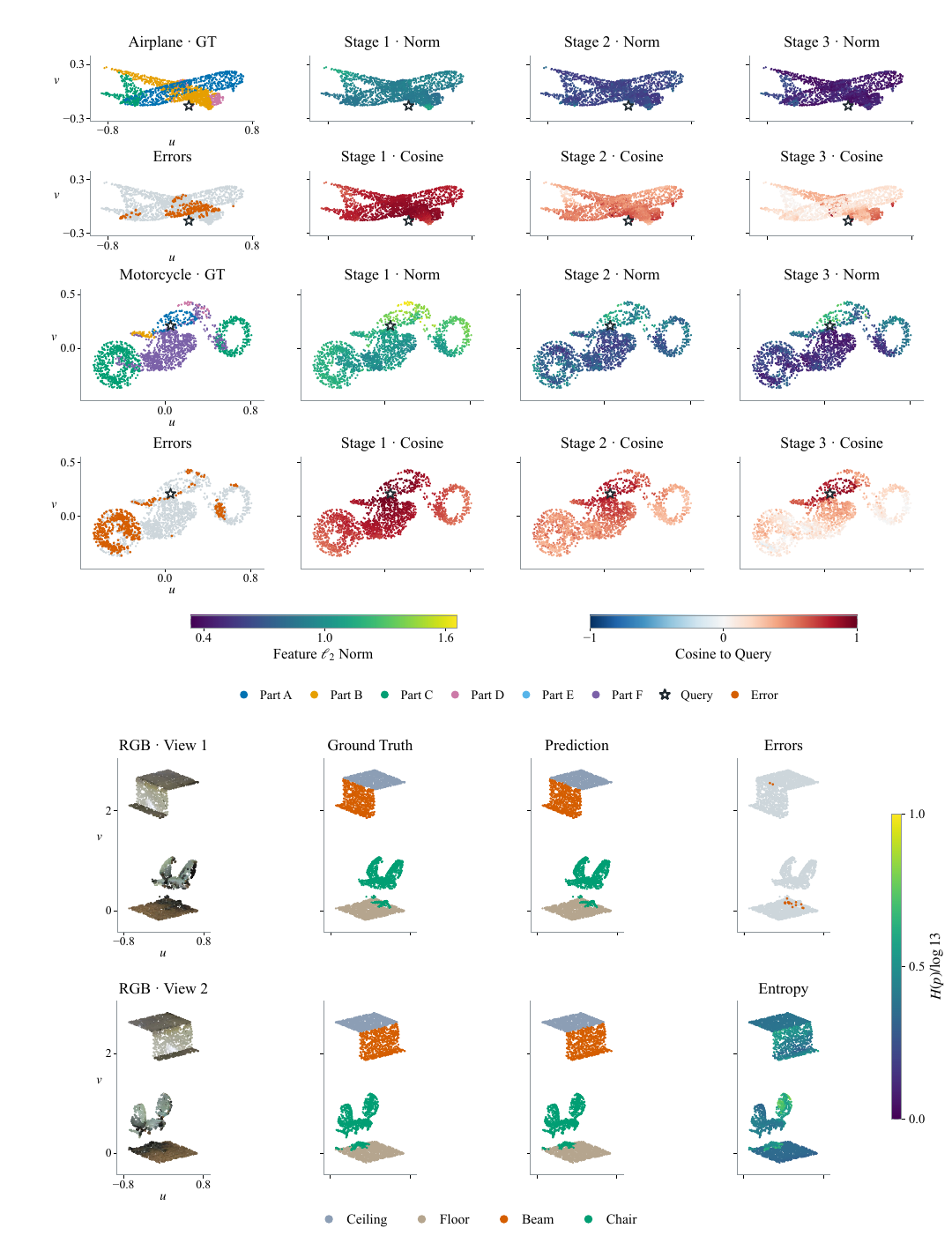}
\caption{\textbf{\RevisionBlack{DGCNN features and segmentation predictions.}} \RevisionBlack{Top: part-segmentation features from conv2/conv4/conv5, with shared norm scales and cosine similarity to the starred query.} Queries maximize the fraction of differently labeled neighbors among 16 neighbors. \RevisionBlack{Bottom: RGB, labels, predictions, errors and entropy for one Area-6 block.} Entropy is normalized by $\log 13$; the two scene rows are different views of the same block.}
\label{fig:v8_local_features_indoor_segmentation}
\end{figure}

\paragraph{Local features and dense predictions.}
Figure~\ref{fig:v8_local_features_indoor_segmentation} combines intermediate feature maps with downstream prediction errors. It shows how feature norms and similarity to a fixed query vary with depth, followed by two views of the same indoor block.

\paragraph{Transformer token relations}
Center neighborhoods give the highest accuracy and lowest time in this study. Feature neighborhoods depend on learned Transformer features, even without directly using attention matrices.

\begin{table}[!htbp]\centering\footnotesize\setlength{\tabcolsep}{2pt}\caption{Transformer graph and stage studies; each panel changes its stated factor.}\label{tab:reported_GS01}\label{tab:reported_GS02}\label{tab:reported_KS02}
\begin{minipage}[t]{.485\linewidth}\vspace{0pt}
\centering\footnotesize
\setlength{\tabcolsep}{3pt}
\textbf{A. Token relations.}\par\smallskip

\begin{tabularx}{\linewidth}{@{}Lcrrrr@{}}
\toprule
\makecell{Token\\relation} & \makecell{Uses\\attention} & MN40 & Scan & Part & Hours\\
\midrule
Attention-affinity & Yes & 93.8 & 89.7 & 86.4 & 31.6\\
Feature kNN & No & 94.0 & 90.0 & 86.6 & 31.3\\
Radius & No & 94.1 & 90.1 & 86.7 & 30.8\\
Center kNN & No & 94.3 & 90.4 & 86.9 & 30.4\\
\bottomrule
\end{tabularx}
\end{minipage}\hfill\begin{minipage}[t]{.485\linewidth}\vspace{0pt}
\begin{minipage}[t]{\linewidth}
\centering\footnotesize
\setlength{\tabcolsep}{3pt}
\textbf{B. Depth selection.}\par\smallskip
\begin{tabularx}{\linewidth}{@{}Lrrrr@{}}
\toprule
Sources & MN40 & Scan & Part & Hours\\
\midrule
Final & 93.2 & 88.3 & 85.7 & 27.4\\
Intermediate + final & 93.8 & 89.4 & 86.3 & 28.9\\
Three sparse & 94.3 & 90.4 & 86.9 & 30.4\\
All blocks & 94.2 & 90.3 & 86.9 & 33.8\\
\bottomrule
\end{tabularx}
\end{minipage}\par\medskip
\begin{minipage}[t]{\linewidth}
\centering\footnotesize
\setlength{\tabcolsep}{3pt}
\textbf{C. Neighborhood sensitivity.}\par\smallskip
\begin{tabularx}{\linewidth}{@{}Lrrr@{}}
\toprule
Graph & MN40 & Scan & Part\\
\midrule
$k=8$ & 94.1 & 90.1 & 86.7\\
$k=16$ & 94.3 & 90.4 & 86.9\\
$k=24$ & 94.2 & 90.3 & 86.8\\
Radius matched degree & 94.2 & 90.2 & 86.8\\
\bottomrule
\end{tabularx}
\end{minipage}
\end{minipage}\end{table}

\paragraph{Transformer depth selection}
Three selected levels and all blocks give the same part mIoU; all blocks require 3.4 more hours.

\paragraph{Transformer neighborhood sensitivity}
Accuracy changes little around the selected neighborhood size. A radius graph and fixed-degree nearest neighbors nevertheless define different supports.

\paragraph{Point-to-image branch mapping}
The default $C_3/C_5/C_5$ mapping exceeds the reversed mapping by 0.9 points on ModelNet40 and 1.6 on ScanObjectNN (Table~\ref{tab:reported_GS04}A). Using $C_3$ for the shallow branch also reduces its projection input from 4,096 to 1,024 dimensions and raises ScanObjectNN accuracy from 88.6\% to 89.6\% relative to $C_5/C_5/C_5$.

\begin{table}[!tbp]\caption{Branch mapping and source/neighborhood settings.}\label{tab:reported_GS04}\label{tab:reported_KS04}\centering\small\setlength{\tabcolsep}{3pt}\renewcommand{\arraystretch}{1.04}
\begin{minipage}[t]{.485\linewidth}\vspace{0pt}\centering\footnotesize\setlength{\tabcolsep}{2pt}
\textbf{A. Point-to-image target mapping.}\par\smallskip
\begin{tabularx}{\linewidth}{@{}LLrrr@{}}
\toprule
\makecell{Map\\for l2/l4/g} & Mapping & MN40 & Scan & Part\\
\midrule
C5/C5/C5 & High-level & 93.1 & 88.6 & 85.7\\
C3/C4/C5 & Strict stage & 93.4 & 89.1 & 85.9\\
C3/C5/C5 & Default & 93.6 & 89.6 & 86.1\\
C5/C3/C3 & Reversed & 92.7 & 88.0 & 85.3\\
\bottomrule
\end{tabularx}

\end{minipage}\hfill\begin{minipage}[t]{.485\linewidth}\vspace{0pt}\centering\footnotesize\setlength{\tabcolsep}{2pt}
\textbf{B. Source and neighborhood settings.}\par\smallskip
\begin{tabularx}{\linewidth}{@{}Lrrr@{}}
\toprule
\makecell{Controlled\\change} & MN40 & Scan & Part\\
\midrule
l4/g & 93.2 & 88.9 & 85.7\\
l2/l4/g (default) & 93.6 & 89.6 & 86.1\\
$k=8$ & 93.4 & 89.3 & 86.0\\
$k=16$ & 93.6 & 89.6 & 86.1\\
$k=24$ & 93.5 & 89.5 & 86.1\\
$w_{\mathrm{band}}=0.02$ & 93.5 & 89.4 & 86.0\\
$w_{\mathrm{band}}=0.05$ & 93.6 & 89.6 & 86.1\\
$w_{\mathrm{band}}=0.10$ & 93.5 & 89.5 & 86.0\\
\bottomrule
\end{tabularx}
\par\smallskip{\footnotesize Each row changes one setting relative to the default.}

\end{minipage}
\end{table}

\paragraph{Shared projection dimension}
Widths 256 and 512 both give 93.6\% on ModelNet40 and 86.1\% part mIoU. The 256-dimensional projection gives the higher ScanObjectNN accuracy (89.6\% versus 89.5\%).

\begin{table}[!tbp]\caption{Projection-width and weight-transfer studies.}\label{tab:reported_GS05}\label{tab:reported_NS05}\label{tab:weight_transfer_part_complete}\centering\small\setlength{\tabcolsep}{3pt}\renewcommand{\arraystretch}{1.04}
\begin{minipage}[t]{.485\linewidth}\vspace{0pt}\centering\footnotesize\setlength{\tabcolsep}{2pt}
\textbf{A. Projection width.}\par\smallskip
\begin{tabularx}{\linewidth}{@{}Lrrrr@{}}
\toprule
Width & Params. (M) & MN40 & Scan & Part\\
\midrule
128 & 72.62 & 93.3 & 89.1 & 85.8\\
256 & 73.37 & 93.6 & 89.6 & 86.1\\
512 & 75.10 & 93.6 & 89.5 & 86.1\\
\bottomrule
\end{tabularx}

\end{minipage}\hfill\begin{minipage}[t]{.485\linewidth}\vspace{0pt}\centering\footnotesize\setlength{\tabcolsep}{2pt}
\textbf{B. Cross-backbone and point-count retuning.}\par\smallskip
\begin{tabularx}{\linewidth}{@{}LLrrr@{}}
\toprule
Encoder / points & Weights & MN40 & Scan & Part\\
\midrule

Transformer / 1,024 & Default & 94.3 & 90.4 & 86.9\\
Transformer / 1,024 & Retuned & 94.4 & 90.7 & 87.0\\
DGCNN / 2,048 & Default & 93.8 & 90.0 & 86.4\\
DGCNN / 2,048 & Retuned & 93.9 & 90.2 & 86.5\\
\bottomrule
\end{tabularx}

\end{minipage}\par\smallskip
{\footnotesize\textbf{C. Independent validation search, DGCNN with 1,024 points (default $\to$ tuned).}\par\smallskip
\begin{tabularx}{\linewidth}{@{}Lrrrr@{}}\toprule
Model & MN40 & Scan & Part & H100 h (search / final run)\\\midrule
H & $93.1\to93.3$ & $88.7\to89.1$ & $85.7\to85.9$ & 72.0 / 26.5\\
H + adaptive fusion & $93.2\to93.5$ & $89.0\to89.5$ & $85.8\to86.1$ & 72.0 / 27.3\\
MS-Point & $93.6\to93.8$ & $89.6\to90.4$ & $86.1\to86.4$ & 72.0 / 30.5\\\midrule
$\Delta$ Full $-$ H & \textbf{+0.5} & \textbf{+1.3} & \textbf{+0.5} & $0.0 / +4.0$\\
$\Delta$ Full $-$ (H + fusion) & \textbf{+0.3} & \textbf{+0.9} & \textbf{+0.3} & $0.0 / +3.2$\\\bottomrule
\end{tabularx}\par\smallskip}\par
{\footnotesize\raggedright H pools and projects the three native stages ($l_2,l_4,g$), paired with $C_3,C_5,C_5$ through separate heads. Its losses are $\mathcal L_{\rm same},\mathcal L_{\rm cross},\mathcal L_{\rm level}$. H with fusion adds an adaptive gate and $\mathcal L_{\rm fused}$. Full MS-Point also supervises six neighborhood-response branches and adds $\mathcal L_{\rm band}$. Deltas compare tuned models in percentage points and H100 hours. Part is instance mIoU.\par}
{\footnotesize\raggedright \textbf{Independent validation search.}
Each model received six trials of 12 H100 hours each (72 H100 hours per model). We selected the setting with the highest frozen-feature linear-SVM accuracy on a fixed stratified 10\% subset of the ScanObjectNN training split, used it across all three downstream tasks, and averaged three paired final pretraining runs.

In $(\mathrm{lr},\tau,\lambda_{\mathrm{cross}})$ order, the six candidates were $(0.0005,0.05,0.2)$, $(0.0005,0.10,0.4)$, $(0.001,0.05,0.4)$, $(0.001,0.10,0.4)$, $(0.001,0.20,0.6)$, $(0.002,0.10,0.2)$. H, H with fusion, and full MS-Point selected the second, fourth, and third settings, respectively. \RevisionRed{On ScanObjectNN the tuned full model exceeds these two tuned baselines by 1.3 and 0.9 percentage points.}\par}
\end{table}

Figure~\ref{fig:projection_sensitivity} expresses the projection-width comparison as per-task gains over 128 dimensions.

\paragraph{Source, neighborhood, and diversity settings}
The reference setting in this neighborhood comparison uses $k=16$.

\paragraph{Parameter-accounting breakdown}
The parameter increments sum to 1.53 million, or 2.13\% of the global baseline. Deployment excludes training-only components.

\section{Graph-Shift Properties}
\label{app:further_theory}
For $H_s=PH$ and $H_d=(I-P)H$, complementarity gives
\begin{equation}
 \|H\|_F^2=\|H_s\|_F^2+\|H_d\|_F^2+2\langle H_s,H_d\rangle_F.
\end{equation}
The cross term generally remains nonzero. For a symmetric shift $S=I-L$, it equals $\sum_j(1-\lambda_j)\lambda_j\|\widehat H_j\|_2^2$. Independent nonlinear adaptors and pooling further change the relation between the two embeddings.

The normalized-Laplacian spectrum lies in $[0,2]$, where $|1-\lambda|$ is not monotone. A lazy reference $S_\eta=I-\eta L$, $0\leq\eta\leq1/2$, has nonnegative, nonincreasing propagation gain and complementary deviation $\eta LH$. This specifies a symmetric reference, separately from the adaptive directed implementation.

If $P$ is nonnegative and row-stochastic, each output channel is a convex combination of its neighbors, giving $\|PH\|_{\max}\leq\|H\|_{\max}$ and $\|(I-P)H\|_{\max}\leq2\|H\|_{\max}$. In particular, repeated averaging with row-stochastic shifts preserves the same upper bound on the maximum absolute feature value.

Point neighborhoods and image grids generally have different spectra and boundary conventions. Cross-modal branch matching aligns their training roles, without assuming a shared Fourier basis.

\section{\RevisionBlack{Replicate-Level Exploratory Results}}
\label{app:seed_results}

\RevisionBlack{Tables~\ref{tab:seed_pairs} and~\ref{tab:filter_seeds} report the replicate-level outcomes for the short-budget studies in Appendix~\ref{app:pilots}. Accuracy is in percent and paired differences are in percentage points.}

\begin{table}[!htbp]
\centering\small
\caption{All ten paired terminal outcomes. Both variants have 73,367,890 trainable parameters; the response intervention changes both point and image inputs.}
\label{tab:seed_pairs}
\begin{tabular}{rrrr}
\toprule
\RevisionBlack{Replicate} & Duplicated features & Responses & Difference\\
\midrule
\RevisionBlack{A} & 91.0454 & 91.3290 & +0.2836\\
\RevisionBlack{B} & 91.2075 & 91.6937 & +0.4862\\
\RevisionBlack{C} & 91.2480 & 91.7342 & +0.4862\\
\RevisionBlack{D} & 91.4100 & 91.6937 & +0.2836\\
\RevisionBlack{E} & 91.5721 & 91.4100 & -0.1621\\
\RevisionBlack{F} & 91.1264 & 90.9643 & -0.1621\\
\RevisionBlack{G} & 91.7342 & 91.3695 & -0.3647\\
\RevisionBlack{H} & 91.7747 & 91.4911 & -0.2836\\
\RevisionBlack{I} & 91.2480 & 91.9773 & +0.7293\\
\RevisionBlack{J} & 91.8558 & 91.8963 & +0.0405\\
\midrule
Mean & 91.4222 & 91.5559 & +0.1337\\
\bottomrule
\end{tabular}
\end{table}

\RevisionBlack{The two-sided sign-flip test enumerates all $2^{10}$ sign assignments and compares their absolute mean differences with the observed value, giving $p=0.3047$. The test assumes sign-exchangeability under the null.}

\begin{table}[!htbp]
\centering
\begin{minipage}{0.78\linewidth}
\centering\small
\caption{\RevisionBlack{Best validation accuracy for the response variants over two training replicates.}}
\label{tab:filter_seeds}
\begin{tabularx}{\linewidth}{@{}Lrrr@{}}
\toprule
Response & \RevisionBlack{Replicate A} & \RevisionBlack{Replicate B} & Mean\\
\midrule
Fixed pair & 90.7012 & 91.1585 & 90.9299\\
Learnable first order & 91.0061 & 91.1585 & 91.0823\\
Polynomial & 91.4634 & 90.9045 & 91.1839\\
Three responses & 90.8028 & 91.1077 & 90.9553\\
\bottomrule
\end{tabularx}
\end{minipage}
\end{table}

\section{Pooled Set Matching and Longer Runs}
\label{app:local_runs}

\subsection{The Implemented Objective}
\RevisionBlack{The extension contains} a pooled set-matching objective in addition to the response variants. A point-side head reads the $l_4$ map and adaptive average pooling produces 64 index bins. An image-side head reads C4 and pools it to an $8\times8$ grid. Features are normalized after these heads and pooling. This is a correspondence-light object-paired objective: consecutive bins in an arbitrary point ordering are not coordinate-defined patches.

Let $u_{ap}$ and $v_{bq}$ be normalized point-bin and image-grid descriptors from objects $a$ and $b$, and $s_{abpq}=u_{ap}^\top v_{bq}$. \RevisionBlack{The object-pair score} is
\begin{equation}
 S_{ab}=\frac12\left[
 \frac1{P_t}\sum_p\tau_m\log\sum_q e^{s_{abpq}/\tau_m}
 +\frac1{Q_t}\sum_q\tau_m\log\sum_p e^{s_{abpq}/\tau_m}\right].
 \label{eq:saved_set_score}
\end{equation}
A symmetric cross-entropy classifies the diagonal object pairs from $S/\tau_c$. The matching temperature $\tau_m$ and contrastive temperature $\tau_c$ play different roles. The log-sum-exp terms provide soft best-match aggregation without a one-to-one correspondence constraint or explicit visibility model.

\RevisionRed{The allocated-head control contains the same local heads but sets the local objective to zero. Pooled set matching trains these heads through the auxiliary matching objective. Both configurations have the same allocated parameter count; only the matching configuration activates the local training paths.}

\subsection{Two-Replicate Short Runs}
\RevisionBlack{The short-budget study uses two training replicates, five warm-start epochs, at most 2,048 objects, and batch size 32. Table~\ref{tab:local_pilot} reports the best validation accuracy from each run.}

\begin{table}[!htbp]
\centering\small
\caption{\RevisionBlack{Pooled-set pilot.} Times are means in minutes; the allocated-head control is inactive in the local objective.}
\label{tab:local_pilot}
\begin{tabularx}{\linewidth}{@{}Lrrrr@{}}
\toprule
Variant & \RevisionBlack{Replicate A} & \RevisionBlack{Replicate B} & Parameters & Minutes\\
\midrule
Inactive local-head control & 90.8537 & 90.8028 & 73,450,066 & 7.23\\
Pooled set matching & 91.0061 & 91.1077 & 73,450,066 & 7.26\\
Three responses + set matching & 91.2602 & 91.2602 & 87,656,664 & 8.76\\
\bottomrule
\end{tabularx}
\end{table}

\RevisionBlack{Pooled set matching improves the mean validation accuracy by 0.2287 percentage points over the inactive local-head control. The three-response combination also increases the response-bank capacity, as reflected in its parameter count.}

\subsection{Twenty-Five-Epoch Warm-Start Runs}
\RevisionBlack{We also evaluate a 25-epoch warm-start continuation with one training replicate, no training-object cap, and batch size 32. Separate columns give the best validation accuracy and the final SVM test accuracy.}

\begin{table}[!htbp]
\centering
\begin{minipage}{0.78\linewidth}
\centering\small
\caption{\RevisionBlack{25-epoch warm-start continuations with one training replicate. Accuracy is in percent and time in hours.}}
\label{tab:long_warmstart}
\begin{tabularx}{\linewidth}{@{}Lrrr@{}}
\toprule
Variant & Best validation & Terminal test & Hours\\
\midrule
Fixed responses & 91.8191 & 90.9238 & 9.63\\
Polynomial responses & 91.3618 & 90.4781 & 12.55\\
Pooled set matching & 91.5142 & 91.8152 & 9.67\\
Three responses + set matching & 91.4126 & 91.2480 & 12.91\\
\bottomrule
\end{tabularx}
\end{minipage}
\end{table}

\RevisionBlack{Fixed responses give the highest best-validation accuracy, while pooled set matching gives the highest terminal-test accuracy in this continuation. The two columns retain their respective selection rules.}

\begin{figure}[!htbp]
\centering
\includegraphics[width=1.0\linewidth,height=.78\textheight,keepaspectratio]{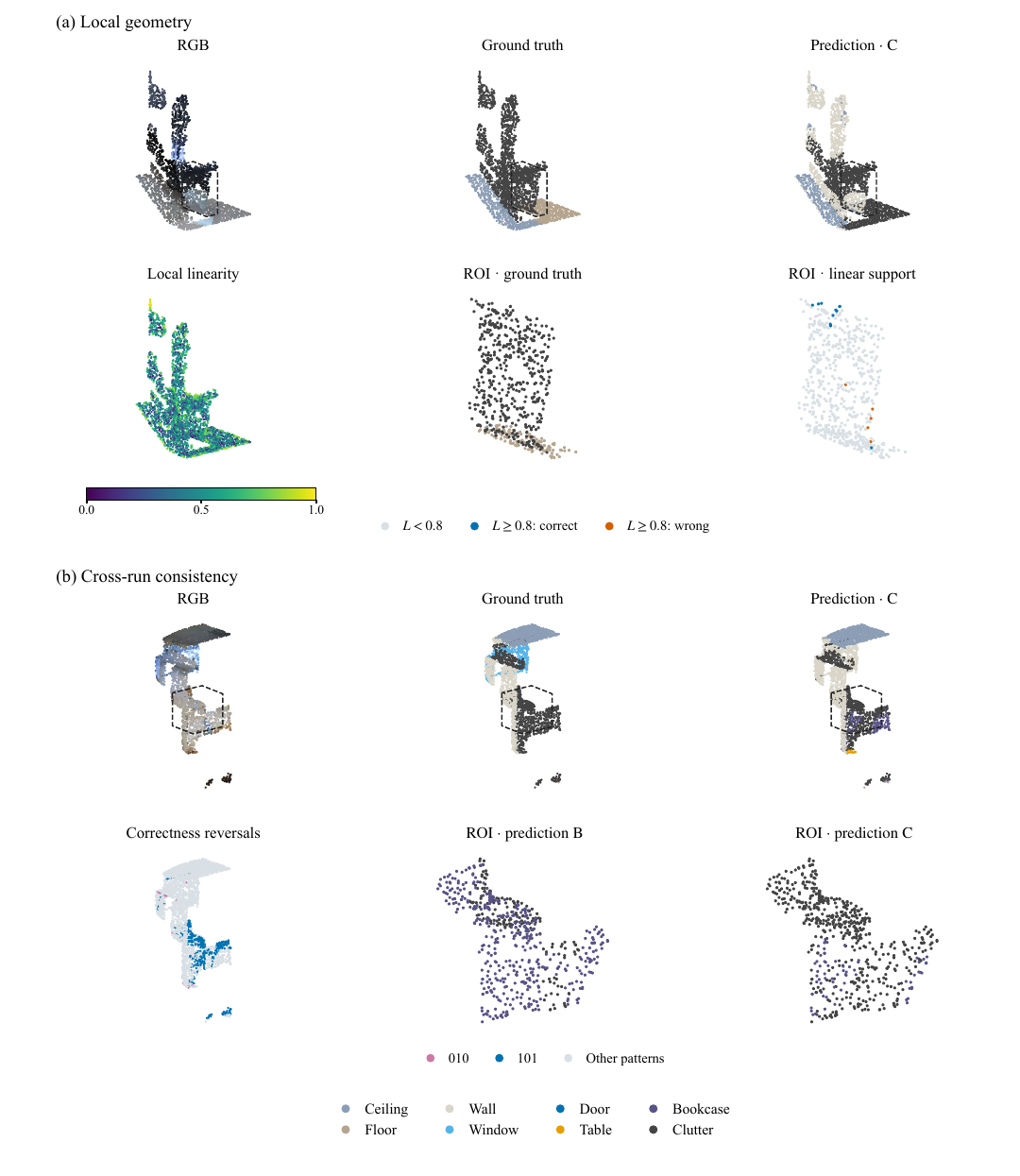}
\caption{\textbf{\RevisionBlack{Geometry and segmentation across DGCNN runs.}} Local linearity is $(\lambda_1-\lambda_2)/\lambda_1$ for the covariance of 16 spatial neighbors. Projected boxes locate the displayed regions. \RevisionBlack{The lower case was selected for the largest number of correctness reversals. Codes 010 and 101 indicate correctness across three training replicates (A/B/C), with 1 for correct and 0 for incorrect.}}
\label{fig:v8_geometry_checkpoint_diagnostics_no_tables}
\end{figure}

\paragraph{Geometry and variation across segmentation runs.}
Figure~\ref{fig:v8_geometry_checkpoint_diagnostics_no_tables} complements the classification records with spatial diagnostics from segmentation models. The upper panels locate errors relative to local geometric support; the lower panels show points whose correctness differs across runs.

\section{Token-Encoder Interface}
\label{app:token_visibility}
\subsection{A Backbone Interface with Explicit Geometric Support}
\RevisionBlack{MS-Point uses a common supervision interface for graph encoders and point Transformers.} For selected stage $d$, the point encoder supplies $(H^{(d)},C^{(d)},\mathcal I^{(d)})$: node-major features $H^{(d)}\in\mathbb R^{M_d\times C_d}$, centers $C^{(d)}\in\mathbb R^{M_d\times3}$, and point memberships $\mathcal I_i^{(d)}$. Centers define a neighborhood; memberships identify the region represented by each token. A class token has no local geometric support and is excluded from the graph, although it can provide a separate object branch.

For a center-space graph, define
\begin{align}
 W_{ij}^{(d)}=\mathbf1\{j\in\operatorname{kNN}(c_i^{(d)};C^{(d)})\},\quad
 P_{ij}^{(d)}=\frac{W_{ij}^{(d)}}{\sum_jW_{ij}^{(d)}},\\
 H_s^{(d)}=P^{(d)}H^{(d)},\quad H_r^{(d)}=H^{(d)}-H_s^{(d)}.
 \label{eq:token_interface}
\end{align}
An implementation must specify self-neighbor inclusion and ensure nonempty neighborhoods. A radius graph can use a self-neighbor fallback. If a symmetric analysis is required, explicitly symmetrize the weights before normalization; ordinary directed $k$-NN is not symmetric. The center graph does not depend on the attention weights. Feature-space $k$-NN instead depends on learned features, even when it does not read attention matrices. \RevisionBlack{The chosen neighborhood rule defines the response construction.}

\paragraph{Semantic depth and spatial scale.}
The three stages in Table~\ref{tab:token_run_config} retain the same spatial resolution. Using three stages and using all blocks both give 86.9 part mIoU, with training costs of 30.4 and 33.8 H100 hours, respectively (Table~\ref{tab:reported_GS02}B).
\par\noindent\begin{minipage}{\linewidth}\centering\footnotesize
\intablecaption{MS-Point-T tokenization and supervision settings.}\label{tab:token_run_config}
\begin{tabularx}{\linewidth}{@{}lL@{}}\toprule
Tokenization & 64 patch centers per view by farthest point sampling, with $k$-NN point grouping\\
Supervised blocks & Blocks 4, 8, and 12 of the 12-block point Transformer\\
Response graph & $k=16$ nearest patch centers among all 64 unmasked tokens\\\bottomrule
\end{tabularx}\end{minipage}\par

\paragraph{Projection, targets, and fusion.}
Each base or response map is pooled into a descriptor and independently projected to 256 dimensions. Image ConvNeXt stages or ViT patch grids can replace ResNet stages without modifying the underlying attention mechanism. A ViT class token cannot be reshaped into a spatial grid; only patch tokens with a known layout participate in image smoothing. The map $\rho$ records the image block/stage, response identity, and head identity. Reusing one image stage for two point stages does not imply shared projection parameters. \RevisionBlack{The target map is specified for each backbone; C3/C5/C5 is the map used by the nine-branch DGCNN variant.}

\paragraph{Masking and graph scope.}
MS-Point-T uses contrastive pretraining on all patch tokens without a reconstruction decoder. Because their centers remain fixed within a view, the three supervised blocks share the same center-$k$NN graph. Computing all pairwise center distances costs $O(M_d^2)$, and gathering neighboring features costs $O(M_d k C_d)$.

\begin{algorithm}[!htbp]
\caption{Structured pretraining with point tokens (interface specification)}
\label{alg:token_interface}
\begin{algorithmic}[1]
\REQUIRE Paired object point cloud and image; selected token stages; map $\rho$
\STATE Construct two point augmentations and the selected encoded image view(s)
\STATE Run each encoder once per input view and retain selected feature maps
\FOR{each point view and selected token stage $d$}
\STATE Obtain $H^{(d)}$, centers, and memberships; construct or reuse the specified graph
\STATE Form $H_s^{(d)}=P^{(d)}H^{(d)}$ and $H_r^{(d)}=H^{(d)}-H_s^{(d)}$
\STATE Adapt, pool, and project each active base/response branch
\ENDFOR
\STATE Construct image-stage base/response targets and apply independent heads using $\rho$
\STATE Compute branch gates and fused embeddings in each modality
\STATE Evaluate the five objectives with the selected branch sets and reduction conventions
\STATE Backpropagate their weighted sum; update trainable modules
\end{algorithmic}
\end{algorithm}

\section{Extended Results, Protocols, and Resource Details}
\label{app:expanded_protocols}
We report further comparisons across backbones, graph constructions, local supervision, and label budgets. Each study fixes the comparison conditions stated in its table. We distinguish published baselines, controlled full-training experiments, and shorter exploratory runs throughout.

\subsection{Published Results and Controlled Reruns}
Table~\ref{tab:literature_expanded} compares frozen representations under the evaluation settings used in the respective publications. The ScanObjectNN split and feature evaluator remain part of each result's protocol. We use Point-PQAE, PointGPT, PointDif, PCP-MAE, OpenShape, and ULIP-2 to contextualize prediction targets and supervision sources; our controlled objective comparisons use the common evaluator in Table~\ref{tab:completed_baselines}.

\par\noindent\begin{minipage}[t]{.60\linewidth}\vspace{0pt}
\centering\footnotesize\setlength{\tabcolsep}{1.5pt}\renewcommand{\arraystretch}{1.02}
\intablecaption{Literature results with explicit evaluation provenance.}
\label{tab:literature_expanded}\label{tab:source_protocol_labels}
\textbf{A. Frozen representations under native evaluation settings.}\par\smallskip
\begin{tabularx}{\linewidth}{@{}LLLrrc@{}}
\toprule
Method & Pretraining & Backbone & MN40 & Scan & Protocol\\
\midrule
Point-BERT & Masked token & Transf. & 87.4 & -- & P0\\
Point-MAE & Recon\-struction & Transf. & 91.0 & 77.7 & P1\\
Point-M2AE & Multiscale\newline MAE & Transf. & 92.9 & 84.1 & P0\\
CrossPoint & Cross-modal & DGCNN & 91.2 & 81.7 & P1\\
I2P-MAE & Image-guided MAE & Transf. & 93.4 & 87.1 & P0\\
MM-Point & Multi-view\newline contrast & DGCNN & 92.4 & 87.8 & P2\\
\makecell[l]{Point-MAE\\+ TPM} & Multiple\newline masks & Transf. & 91.2 & 78.2 & P4\\
\makecell[l]{Point-M2AE\\+ TPM} & Multiple\newline masks & Transf. & 93.2 & 84.0 & P4\\
MS-Point & Structured\newline contrast & DGCNN & 93.6 & 89.6 & P3\\
\bottomrule
\end{tabularx}
\par\smallskip\textbf{B. Full fine-tuning reference (separate evaluator and split).}\par\smallskip
\begin{tabularx}{\linewidth}{@{}LLLrr@{}}
\toprule
Method & Evaluator & Scan setting & MN40 & Scan\\
\midrule
Point-MAE & Full fine-tuning & PB-T50-RS & 93.8 & 85.18\\
\bottomrule
\end{tabularx}

\end{minipage}\hfill\begin{minipage}[t]{.37\linewidth}\vspace{0pt}
\par\smallskip{\footnotesize P1: frozen SVM, source OBJ-BG setting (CrossPoint and Point-MAE). P2: frozen SVM, published setting (MM-Point). P3: frozen SVM, benchmark setting (MS-Point). P4: frozen SVM, PB-T50-RS, TPM arXiv v1, as detailed below. \RevisionBlack{P0: native frozen result with evaluator/split details unreported. Panels A and B retain their respective protocols; controlled reruns are in Table~\ref{tab:completed_baselines}.}}

{\footnotesize Transf.: Transformer.}

\end{minipage}\par\medskip

\paragraph{Evaluation Protocol of TPM.}
TPM arXiv v1 (2409.17547, Table 2) evaluates frozen linear SVM on ScanObjectNN PB-T50-RS. Its Point-MAE and Point-MAE+TPM results are 90.8/77.1 and 91.2/78.2 on ModelNet40/ScanObjectNN; Point-M2AE and Point-M2AE+TPM obtain 92.9/83.1 and 93.2/84.0. We keep these native results separate from our controlled OBJ-BG reruns, where Point-MAE+TPM obtains 91.8/79.2.

The references are Point-BERT~\citep{yu2022pointbert}, Point-MAE~\citep{pang2022masked}, Point-M2AE~\citep{zhang2022pointm2ae}, CrossPoint~\citep{afham2022crosspoint}, I2P-MAE~\citep{zhang2023i2p}, MM-Point~\citep{yu2024mmpoint}, and TPM~\citep{liu2025tpm}. \RevisionRed{TPM obtains 91.8/79.2 in our controlled evaluation (Table~\ref{tab:completed_baselines}) and 91.2/78.2 under its published protocol.}

\subsection{Cross-Backbone and Image-Stage Studies}

\par\noindent\begin{minipage}[t]{.52\linewidth}\vspace{0pt}

\centering\footnotesize
\setlength{\tabcolsep}{1.5pt}
\intablecaption{Full structured objective with different feature providers. Part is instance mIoU.}
\label{tab:crossbackbone_complete}
\begin{tabularx}{\linewidth}{@{}LLrrrr@{}}
\toprule
3D encoder & 2D encoder & MN40 & Scan & Part & H100 h\\
\midrule
DGCNN & ResNet-50 & 93.6 & 89.6 & 86.1 & 30.5\\
Point\newline Transformer & ResNet-50 & 94.3 & 90.4 & 86.9 & 30.4\\
DGCNN & ConvNeXt-B & 93.9 & 90.0 & 86.5 & 32.1\\
Point\newline Transformer & ConvNeXt-B & 94.6 & 90.8 & 87.2 & 32.3\\
\bottomrule
\end{tabularx}

\end{minipage}\hfill\begin{minipage}[t]{.45\linewidth}\vspace{0pt}
The same-Transformer global/structured contrast in the main text isolates the objective more directly than comparing different encoders. The additional matrix tests whether stronger feature providers can coexist with the structured objective.
\end{minipage}\par\medskip

\par\noindent\textbf{ResNet-50 stage convention.} For a $224\times224$ input, C2--C5 denote PyTorch \texttt{layer1}--\texttt{layer4} (\texttt{conv2\_x}--\texttt{conv5\_x}), not literal convolution indices. Their (stride, channels, grid) tuples are $(4,256,56\times56)$, $(8,512,28\times28)$, $(16,1024,14\times14)$, and $(32,2048,7\times7)$, respectively.\par

\begin{table}[tbp]
\begin{minipage}[t]{0.485\linewidth}
\centering\footnotesize
\setlength{\tabcolsep}{3pt}
\caption{Image-feature selection study. \RevisionBlack{Relative cost measures image-feature computation.}}
\label{tab:image_stages_complete}
\begin{tabularx}{\linewidth}{@{}Lrrr@{}}
\toprule
Image features & MN40 & Scan & Rel. cost\\
\midrule
C5 only & 92.9 & 88.1 & 0.82\\
C3 + C5 & 93.4 & 89.0 & 0.91\\
C2 + C3 + C4 + C5 & 93.6 & 89.6 & 1.00\\
Unaligned blocks (10/15/20) & 93.1 & 88.5 & 0.96\\
\bottomrule
\end{tabularx}
\end{minipage}\hfill
\begin{minipage}[t]{0.485\linewidth}
\centering\footnotesize
\setlength{\tabcolsep}{2pt}
\caption{Coordination-weight sensitivity and a validation-selected configuration; Part is instance mIoU.}
\label{tab:coordination_weights_complete}
\begin{tabularx}{\linewidth}{@{}Lrrr@{}}
\toprule
Setting & MN40 & Scan & Part\\
\midrule
No coordination ($w=0$) & 93.1 & 88.7 & 85.6\\
$w=0.2$ & 93.4 & 89.2 & 85.9\\
Default ($w=0.4$) & 93.6 & 89.6 & 86.1\\
$w=0.6$ & 93.5 & 89.4 & 86.0\\
Validation-selected & 93.8 & 90.4 & 86.4\\
\bottomrule
\end{tabularx}
\end{minipage}
\end{table}

The default nine-branch model pairs $l_2,l_4,g$ with $C_3,C_5,C_5$, respectively (Table~\ref{tab:reported_GS04}A). Table~\ref{tab:image_stages_complete} compares image-stage selections, including the four stages $C_2$--$C_5$ shown in Figure~\ref{fig:eleven_branch_flow}.

\subsection{Loss Roles, Weights, and Extended Task Metrics}
\RevisionBlack{The five loss terms jointly train the shared encoder through augmentation consistency, cross-level coordination, branch alignment, fused alignment, and response diversity. Appendix~\ref{sec:interface_analysis} analyzes how these training signals interact.}

\par\noindent\textbf{Objective weights.} Table~\ref{tab:app_training_config} lists the default coefficients with the training configuration. The five roles above describe jointly optimized terms; their exact averaging conventions determine effective regularization.\par

The combined loss and retuning results are in Tables~\ref{tab:losses_part_complete} and~\ref{tab:weight_transfer_part_complete}. Removing branch-level alignment has the largest classification reduction in the loss-deletion table. This supports its role within the complete system; the deletion still leaves image supervision through the fused target. \RevisionBlack{The default weights support reuse across tasks; validation selection improves ScanObjectNN accuracy by 0.8 points.}

\subsection{Fusion, Factorial Effects, and Geometry Controls}

\begin{table}[!tbp]\centering\small\setlength{\tabcolsep}{3pt}\renewcommand{\arraystretch}{1.04}
\begin{minipage}[t]{.485\linewidth}\vspace{0pt}\centering\footnotesize\setlength{\tabcolsep}{2pt}
\caption{Fusion study.}\label{tab:fusion_complete}
\begin{tabularx}{\linewidth}{@{}Lrrrr@{}}
\toprule
Fusion & MN40 & Scan & Part & Rel. cost\\
\midrule
Mean & 93.3 & 89.0 & 85.8 & 0.94\\
Concatenate + MLP & 93.4 & 89.2 & 85.9 & 1.12\\
Adaptive gates & 93.6 & 89.6 & 86.1 & 1.00\\
\bottomrule
\end{tabularx}
\par\smallskip{\footnotesize Relative cost is the fusion-cost measure.}

\end{minipage}\hfill\begin{minipage}[t]{.485\linewidth}\vspace{0pt}\centering\footnotesize\setlength{\tabcolsep}{2pt}
\caption{Accuracy at H100 time budgets.}\label{tab:convergence_complete}
\begin{tabularx}{\linewidth}{@{}Lrrrr@{}}
\toprule
Method & 12 h & 24 h & Final & Final h\\
\midrule
Point-MAE & 88.6 & 90.0 & 91.4 & 72.0\\
Inter-MAE & 89.1 & 91.0 & 92.1 & 76.5\\
MM-Point & 91.0 & 92.1 & 92.5 & 36.9\\
MS-Point & 92.2 & 93.3 & 93.6 & 30.5\\
MS-Point-T & 92.8 & 94.0 & 94.3 & 30.4\\
\bottomrule
\end{tabularx}
\par\smallskip{\footnotesize MN40 accuracy; final time in hours. \RevisionBlack{Accuracy is evaluated at the listed training-time budgets.}}

\end{minipage}
\end{table}

\RevisionBlack{Table~\ref{tab:factorial_part_complete} evaluates hierarchy and responses jointly. Their combined gains overlap, with interaction contrasts of $-0.1$ and $-0.5$ percentage points on ModelNet40 and ScanObjectNN. The fusion study compares the resulting representations under mean, concatenation, and adaptive combination. See the response ablations in Table~\ref{tab:reported_GS03}.}

\subsection{View Counts, Timing, and Optimization Progress}
At each iteration, standard MS-Point encodes two augmented point views and one image per object, drawing the image from two offline renderings. MM-Point encodes four images per object from 24 candidates (Table~\ref{tab:cost_full_complete}). The dense-task variant encodes a second image.

\RevisionRed{We give total training time for complete systems and per-epoch time for the control runs. MS-Point uses 30.5 GPU hours, compared with 72.0/76.5 hours for Point-MAE/Inter-MAE, a reduction of 57.64\%/60.13\% for these complete systems. The 50.0-minute epoch timing characterizes the control runs.}

\subsection{Paired Confidence Intervals}

Table~\ref{tab:paired_complete} places the paired intervals beside the corresponding marginal summaries. \RevisionBlack{The paired intervals summarize full-training differences across matched runs. Appendix~\ref{app:seed_results} reports the ten-pair short-budget pilot separately.}

\subsection{Cross-Modal Perturbations and Outdoor Class Groups}

\begin{table}[!htbp]
\begin{minipage}[t]{.485\linewidth}\vspace{0pt}
\centering\footnotesize
\setlength{\tabcolsep}{2pt}
\caption{Our observation perturbations; Part is instance mIoU. No exact calibration is used for object-level rows.}
\label{tab:perturbation_complete}
\begin{tabularx}{\linewidth}{@{}LLrrr@{}}
\toprule
Condition & Calibration & MN40 & Scan & Part\\
\midrule
Object identity & No & 93.6 & 89.6 & 86.1\\
Viewpoint $\pm15^\circ$ & No & 93.5 & 89.4 & 86.0\\
30\% random occlusion & No & 93.3 & 89.2 & 85.8\\
Asynchronous same object & No & 93.2 & 89.0 & 85.7\\
Region soft matching & Coarse & 93.8 & 90.1 & 86.7\\
\bottomrule
\end{tabularx}
\end{minipage}\hfill\begin{minipage}[t]{.485\linewidth}\vspace{0pt}
\centering\footnotesize
\setlength{\tabcolsep}{1.5pt}
\caption{\RevisionBlack{Scene results with 1\% labels, MinkUNet-34C, sequence 08. Camera denotes pretraining access.}}
\label{tab:outdoor_reference_complete}
\begin{tabularx}{\linewidth}{@{}LLrrr@{}}
\toprule
Initialization & Camera & mIoU & Gain & GPU h\\
\midrule
Random initialization & No & 34.2 & -- & 26.0\\
Scene-only SSL & No & 40.8 & +6.6 & 35.0\\
Global MS-Point & Yes & 44.6 & +10.4 & 42.0\\
MS-Point + local & Yes & 47.3 & +13.1 & 50.0\\
\bottomrule
\end{tabularx}
\end{minipage}
\end{table}
\FloatBarrier

\par\noindent\begin{minipage}[t]{.49\linewidth}\vspace{0pt}

\centering\footnotesize
\setlength{\tabcolsep}{1.5pt}
\intablecaption{Class-frequency groups at 1\% labels: eight frequent, six medium-frequency, and five rare classes.}
\label{tab:outdoor_groups_complete}
\begin{tabularx}{\linewidth}{@{}Lrrrr@{}}
\toprule
Initialization & Freq. (8) & Mid. (6) & Rare (5) & Overall\\
\midrule
Random initialization & 45.8 & 31.1 & 19.4 & 34.2\\
Scene-only SSL & 52.4 & 38.8 & 24.7 & 40.8\\
Global MS-Point & 56.0 & 42.7 & 28.6 & 44.6\\
MS-Point + local & 58.2 & 45.1 & 32.3 & 47.3\\
\bottomrule
\end{tabularx}

\end{minipage}\hfill\begin{minipage}[t]{.48\linewidth}\vspace{0pt}
Local alignment improves mIoU in the frequent, medium-frequency, and rare class groups by 2.2, 2.4, and 3.7 points over global MS-Point. Gains over random initialization are 12.4, 14.0, and 12.9 points, respectively, a 66.5\% relative improvement for rare classes.
\end{minipage}\par\medskip

\FloatBarrier

\begin{figure}[!htbp]
\centering
\includegraphics[width=1.0\linewidth,height=.78\textheight,keepaspectratio]{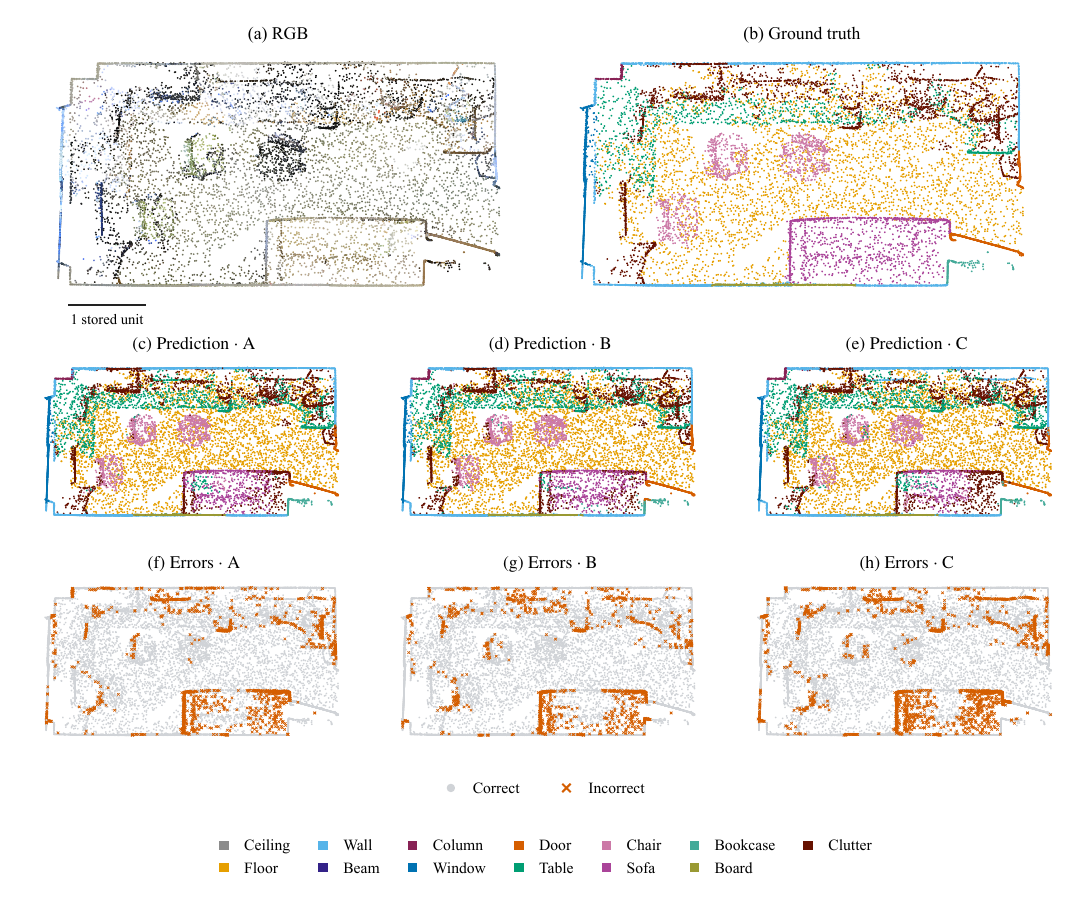}
\caption{\textbf{\RevisionBlack{Indoor segmentation across three DGCNN training replicates.}} The visualization combines 55 stored blocks and displays the same 13,334 coordinate cells in all panels, using a geometry-based sample and the lower 62\% height slice. Orange crosses mark incorrect labels. The scale bar is in stored coordinate units. \RevisionBlack{The block mosaic compares replicates A, B and C.}}
\label{fig:v8_office_mosaic_comparison}
\end{figure}

\paragraph{Spatial context across indoor blocks.}
Figure~\ref{fig:v8_office_mosaic_comparison} expands the indoor examples to an office mosaic. The shared view makes it easier to locate differences between the three sets of predictions and to compare their spatial error patterns.


\begin{thebibliography}{44}
\providecommand{\natexlab}[1]{#1}
\providecommand{\url}[1]{\texttt{#1}}
\expandafter\ifx\csname urlstyle\endcsname\relax
  \providecommand{\doi}[1]{doi: #1}\else
  \providecommand{\doi}{doi: \begingroup \urlstyle{rm}\Url}\fi

\bibitem[Afham et~al.(2022)Afham, Dissanayake, Dissanayake, Dharmasiri,
  Thilakarathna, and Rodrigo]{afham2022crosspoint}
Mohamed Afham, Isuru Dissanayake, Dinithi Dissanayake, Amaya Dharmasiri,
  Kanchana Thilakarathna, and Ranga Rodrigo.
\newblock {CrossPoint: Self-Supervised Cross-Modal Contrastive Learning for 3D
  Point Cloud Understanding}.
\newblock In \emph{Proceedings of the IEEE/CVF Conference on Computer Vision
  and Pattern Recognition (CVPR)}, pp.\  9902--9912, 2022.

\bibitem[Armeni et~al.(2016)Armeni, Sener, Zamir, Jiang, Brilakis, Fischer, and
  Savarese]{armeni20163d}
Iro Armeni, Ozan Sener, Amir~R. Zamir, Helen Jiang, Ioannis Brilakis, Martin
  Fischer, and Silvio Savarese.
\newblock {3D Semantic Parsing of Large-Scale Indoor Spaces}.
\newblock In \emph{Proceedings of the IEEE Conference on Computer Vision and
  Pattern Recognition (CVPR)}, pp.\  1534--1543, 2016.

\bibitem[Behley et~al.(2019)Behley, Garbade, Milioto, Quenzel, Behnke,
  Stachniss, and Gall]{behley2019semantickitti}
Jens Behley, Martin Garbade, Andres Milioto, Jan Quenzel, Sven Behnke, Cyrill
  Stachniss, and Jurgen Gall.
\newblock {SemanticKITTI: A Dataset for Semantic Scene Understanding of LiDAR
  Sequences}.
\newblock In \emph{Proceedings of the IEEE/CVF International Conference on
  Computer Vision (ICCV)}, 2019.

\bibitem[Caron et~al.(2020)Caron, Misra, Mairal, Goyal, Bojanowski, and
  Joulin]{caron2020unsupervised}
Mathilde Caron, Ishan Misra, Julien Mairal, Priya Goyal, Piotr Bojanowski, and
  Armand Joulin.
\newblock {Unsupervised Learning of Visual Features by Contrasting Cluster
  Assignments}.
\newblock In \emph{Advances in Neural Information Processing Systems},
  volume~33, pp.\  9912--9924. Curran Associates, Inc., 2020.

\bibitem[Chang et~al.(2015)Chang, Funkhouser, Guibas, Hanrahan, Huang, Li,
  Savarese, Savva, Song, Su, Xiao, Yi, and Yu]{chang2015shapenet}
Angel~X. Chang, Thomas Funkhouser, Leonidas Guibas, Pat Hanrahan, Qixing Huang,
  Zimo Li, Silvio Savarese, Manolis Savva, Shuran Song, Hao Su, Jianxiong Xiao,
  Li~Yi, and Fisher Yu.
\newblock {ShapeNet: An Information-Rich 3D Model Repository}.
\newblock \emph{arXiv preprint arXiv:1512.03012}, 2015.
\newblock URL \url{https://arxiv.org/abs/1512.03012}.

\bibitem[Chen et~al.(2023)Chen, Wang, Yang, Yu, Yuan, and
  Yue]{chen2023pointgpt}
Guangyan Chen, Meiling Wang, Yi~Yang, Kai Yu, Li~Yuan, and Yufeng Yue.
\newblock {PointGPT: Auto-regressively Generative Pre-training from Point
  Clouds}.
\newblock In \emph{Advances in Neural Information Processing Systems},
  volume~36, pp.\  29667--29679. Curran Associates, Inc., 2023.
\newblock \doi{10.52202/075280-1291}.

\bibitem[Chen et~al.(2020)Chen, Kornblith, Norouzi, and Hinton]{chen2020simple}
Ting Chen, Simon Kornblith, Mohammad Norouzi, and Geoffrey Hinton.
\newblock {A Simple Framework for Contrastive Learning of Visual
  Representations}.
\newblock In \emph{Proceedings of the 37th International Conference on Machine
  Learning}, volume 119 of \emph{Proceedings of Machine Learning Research},
  pp.\  1597--1607. PMLR, 2020.

\bibitem[Defferrard et~al.(2016)Defferrard, Bresson, and
  Vandergheynst]{defferrard2016convolutional}
Micha\"{e}l Defferrard, Xavier Bresson, and Pierre Vandergheynst.
\newblock {Convolutional Neural Networks on Graphs with Fast Localized Spectral
  Filtering}.
\newblock In \emph{Advances in Neural Information Processing Systems},
  volume~29. Curran Associates, Inc., 2016.

\bibitem[He et~al.(2016)He, Zhang, Ren, and Sun]{he2016deep}
Kaiming He, Xiangyu Zhang, Shaoqing Ren, and Jian Sun.
\newblock {Deep Residual Learning for Image Recognition}.
\newblock In \emph{Proceedings of the IEEE Conference on Computer Vision and
  Pattern Recognition (CVPR)}, pp.\  770--778, 2016.

\bibitem[Huang et~al.(2021)Huang, Xie, Zhu, and Zhu]{liu2021strl}
Siyuan Huang, Yichen Xie, Song-Chun Zhu, and Yixin Zhu.
\newblock {Spatio-Temporal Self-Supervised Representation Learning for 3D Point
  Clouds}.
\newblock In \emph{Proceedings of the IEEE/CVF International Conference on
  Computer Vision (ICCV)}, pp.\  6535--6545, 2021.

\bibitem[Kingma \& Ba(2015)Kingma and Ba]{kingma2014adam}
Diederik~P. Kingma and Jimmy Ba.
\newblock {Adam: A Method for Stochastic Optimization}.
\newblock In \emph{International Conference on Learning Representations}, 2015.

\bibitem[Li et~al.(2018)Li, Bu, Sun, Wu, Di, and Chen]{li2018pointcnn}
Yangyan Li, Rui Bu, Mingchao Sun, Wei Wu, Xinhan Di, and Baoquan Chen.
\newblock {PointCNN: Convolution On X-Transformed Points}.
\newblock In \emph{Advances in Neural Information Processing Systems},
  volume~31. Curran Associates, Inc., 2018.

\bibitem[Liu et~al.(2024)Liu, Wu, Gong, Liu, Miao, and Ma]{liu2024intermae}
Jiaming Liu, Yue Wu, Maoguo Gong, Zhixiao Liu, Qiguang Miao, and Wenping Ma.
\newblock {Inter-Modal Masked Autoencoder for Self-Supervised Learning on Point
  Clouds}.
\newblock \emph{IEEE Transactions on Multimedia}, 26:\penalty0 3897--3908,
  2024.
\newblock \doi{10.1109/tmm.2023.3317998}.

\bibitem[Liu et~al.(2025)Liu, Kong, Wu, Gong, Li, Miao, Ma, and
  Qin]{liu2025tpm}
Jiaming Liu, Linghe Kong, Yue Wu, Maoguo Gong, Hao Li, Qiguang Miao, Wenping
  Ma, and Can Qin.
\newblock {Triple Point Masking}.
\newblock \emph{IEEE Transactions on Circuits and Systems for Video
  Technology}, 35\penalty0 (9):\penalty0 8466--8477, 2025.
\newblock \doi{10.1109/tcsvt.2025.3549786}.

\bibitem[Liu et~al.(2023{\natexlab{a}})Liu, Shi, Kuang, Zhu, Li, Han, Cai,
  Porikli, and Su]{liu2023openshape}
Minghua Liu, Ruoxi Shi, Kaiming Kuang, Yinhao Zhu, Xuanlin Li, Shizhong Han,
  Hong Cai, Fatih Porikli, and Hao Su.
\newblock {OpenShape: Scaling Up 3D Shape Representation Towards Open-World
  Understanding}.
\newblock In \emph{Advances in Neural Information Processing Systems},
  volume~36, pp.\  44860--44879. Curran Associates, Inc., 2023{\natexlab{a}}.
\newblock \doi{10.52202/075280-1944}.

\bibitem[Liu et~al.(2019)Liu, Fan, Xiang, and Pan]{liu2019relation}
Yongcheng Liu, Bin Fan, Shiming Xiang, and Chunhong Pan.
\newblock {Relation-Shape Convolutional Neural Network for Point Cloud
  Analysis}.
\newblock In \emph{Proceedings of the IEEE/CVF Conference on Computer Vision
  and Pattern Recognition (CVPR)}, pp.\  8895--8904, 2019.

\bibitem[Liu et~al.(2023{\natexlab{b}})Liu, Kong, CEN, Chen, Zhang, Pan, Chen,
  and Liu]{liu2023seal}
Youquan Liu, Lingdong Kong, Jun CEN, Runnan Chen, Wenwei Zhang, Liang Pan, Kai
  Chen, and Ziwei Liu.
\newblock {Segment Any Point Cloud Sequences by Distilling Vision Foundation
  Models}.
\newblock In \emph{Advances in Neural Information Processing Systems},
  volume~36, pp.\  37193--37229. Curran Associates, Inc., 2023{\natexlab{b}}.
\newblock \doi{10.52202/075280-1617}.

\bibitem[Liu et~al.(2022{\natexlab{b}})Liu, Mao, Wu, Feichtenhofer, Darrell,
  and Xie]{liu2022convnext}
Zhuang Liu, Hanzi Mao, Chao-Yuan Wu, Christoph Feichtenhofer, Trevor Darrell,
  and Saining Xie.
\newblock {A ConvNet for the 2020s}.
\newblock In \emph{Proceedings of the IEEE/CVF Conference on Computer Vision
  and Pattern Recognition (CVPR)}, pp.\  11976--11986, 2022{\natexlab{b}}.

\bibitem[Pang et~al.(2022)Pang, Wang, Tay, Liu, Tian, and Yuan]{pang2022masked}
Yatian Pang, Wenxiao Wang, Francis E.~H. Tay, Wei Liu, Yonghong Tian, and
  Li~Yuan.
\newblock {Masked Autoencoders for Point Cloud Self-supervised Learning}.
\newblock In \emph{European Conference on Computer Vision}, pp.\  604--621,
  2022.
\newblock \doi{10.1007/978-3-031-20086-1_35}.

\bibitem[Qi et~al.(2017{\natexlab{a}})Qi, Su, Mo, and Guibas]{qi2017pointnet}
Charles~R. Qi, Hao Su, Kaichun Mo, and Leonidas~J. Guibas.
\newblock {PointNet: Deep Learning on Point Sets for 3D Classification and
  Segmentation}.
\newblock In \emph{Proceedings of the IEEE Conference on Computer Vision and
  Pattern Recognition (CVPR)}, pp.\  652--660, 2017{\natexlab{a}}.

\bibitem[Qi et~al.(2017{\natexlab{b}})Qi, Yi, Su, and
  Guibas]{qi2017pointnetplusplus}
Charles~Ruizhongtai Qi, Li~Yi, Hao Su, and Leonidas Guibas.
\newblock {PointNet++: Deep Hierarchical Feature Learning on Point Sets in a
  Metric Space}.
\newblock In \emph{Advances in Neural Information Processing Systems},
  volume~30. Curran Associates, Inc., 2017{\natexlab{b}}.

\bibitem[Qi et~al.(2023)Qi, Dong, Fan, Ge, Zhang, Ma, and Yi]{qi2023recon}
Zekun Qi, Runpei Dong, Guofan Fan, Zheng Ge, Xiangyu Zhang, Kaisheng Ma, and
  Li~Yi.
\newblock {Contrast with Reconstruct: Contrastive 3{D} Representation Learning
  Guided by Generative Pretraining}.
\newblock In \emph{Proceedings of the 40th International Conference on Machine
  Learning}, volume 202 of \emph{Proceedings of Machine Learning Research},
  pp.\  28223--28243. PMLR, 2023.

\bibitem[Sauder \& Sievers(2019)Sauder and Sievers]{sauder2019self}
Jonathan Sauder and Bjarne Sievers.
\newblock {Self-Supervised Deep Learning on Point Clouds by Reconstructing
  Space}.
\newblock In \emph{Advances in Neural Information Processing Systems},
  volume~32. Curran Associates, Inc., 2019.

\bibitem[Sautier et~al.(2022)Sautier, Puy, Gidaris, Boulch, Bursuc, and
  Marlet]{sautier2022slidr}
Corentin Sautier, Gilles Puy, Spyros Gidaris, Alexandre Boulch, Andrei Bursuc,
  and Renaud Marlet.
\newblock {Image-to-Lidar Self-Supervised Distillation for Autonomous Driving
  Data}.
\newblock In \emph{Proceedings of the IEEE/CVF Conference on Computer Vision
  and Pattern Recognition (CVPR)}, pp.\  9891--9901, 2022.

\bibitem[Shuman et~al.(2013)Shuman, Narang, Frossard, Ortega, and
  Vandergheynst]{shuman2013emerging}
D.~I. Shuman, S.~K. Narang, P.~Frossard, A.~Ortega, and P.~Vandergheynst.
\newblock {The emerging field of signal processing on graphs: Extending
  high-dimensional data analysis to networks and other irregular domains}.
\newblock \emph{IEEE Signal Processing Magazine}, 30\penalty0 (3):\penalty0
  83--98, 2013.
\newblock \doi{10.1109/msp.2012.2235192}.

\bibitem[Uy et~al.(2019)Uy, Pham, Hua, Nguyen, and Yeung]{uy2019revisiting}
Mikaela~Angelina Uy, Quang-Hieu Pham, Binh-Son Hua, Thanh Nguyen, and Sai-Kit
  Yeung.
\newblock {Revisiting Point Cloud Classification: A New Benchmark Dataset and
  Classification Model on Real-World Data}.
\newblock In \emph{Proceedings of the IEEE/CVF International Conference on
  Computer Vision (ICCV)}, pp.\  1588--1597, 2019.

\bibitem[Wang et~al.(2024)Wang, Wu, Lam, Ning, Yu, Wang, Li, and
  Srikanthan]{wang2024gpsformer}
\RevisionRed{Changshuo Wang, Meiqing Wu, Siew-Kei Lam, Xin Ning, Shangshu Yu, Ruiping Wang,
  Weijun Li, and Thambipillai Srikanthan.
\newblock {GPSFormer: A Global Perception and Local Structure Fitting-Based
  Transformer for Point Cloud Understanding}.
\newblock In \emph{{Computer Vision -- ECCV 2024}}, pp.\  75--92. Springer
  Nature Switzerland, October 2024.
\newblock \doi{10.1007/978-3-031-73242-3_5}.}

\bibitem[Wang et~al.(2021)Wang, Liu, Yue, Lasenby, and
  Kusner]{wang2021unsupervised}
Hanchen Wang, Qi~Liu, Xiangyu Yue, Joan Lasenby, and Matt~J. Kusner.
\newblock {Unsupervised Point Cloud Pre-Training via Occlusion Completion}.
\newblock In \emph{Proceedings of the IEEE/CVF International Conference on
  Computer Vision (ICCV)}, pp.\  9782--9792, 2021.

\bibitem[Wang et~al.(2019)Wang, Sun, Liu, Sarma, Bronstein, and
  Solomon]{wang2019dynamic}
Yue Wang, Yongbin Sun, Ziwei Liu, Sanjay~E. Sarma, Michael~M. Bronstein, and
  Justin~M. Solomon.
\newblock {Dynamic Graph CNN for Learning on Point Clouds}.
\newblock \emph{ACM Transactions on Graphics (TOG)}, 38\penalty0 (5):\penalty0
  1--12, 2019.
\newblock \doi{10.1145/3326362}.

\bibitem[Wu et~al.(2016)Wu, Zhang, Xue, Freeman, and Tenenbaum]{wu2016learning}
Jiajun Wu, Chengkai Zhang, Tianfan Xue, Bill Freeman, and Josh Tenenbaum.
\newblock {Learning a Probabilistic Latent Space of Object Shapes via 3D
  Generative-Adversarial Modeling}.
\newblock In \emph{Advances in Neural Information Processing Systems},
  volume~29. Curran Associates, Inc., 2016.

\bibitem[Wu et~al.(2015)Wu, Song, Khosla, Yu, Zhang, Tang, and Xiao]{wu20153d}
Zhirong Wu, Shuran Song, Aditya Khosla, Fisher Yu, Linguang Zhang, Xiaoou Tang,
  and Jianxiong Xiao.
\newblock {3D ShapeNets: A Deep Representation for Volumetric Shapes}.
\newblock In \emph{Proceedings of the IEEE Conference on Computer Vision and
  Pattern Recognition (CVPR)}, pp.\  1912--1920, 2015.

\bibitem[Xie et~al.(2020)Xie, Gu, Guo, Qi, Guibas, and
  Litany]{xie2020pointcontrast}
Saining Xie, Jiatao Gu, Demi Guo, Charles~R Qi, Leonidas Guibas, and Or~Litany.
\newblock {{PointContrast}: Unsupervised Pre-training for 3D Point Cloud
  Understanding}.
\newblock In \emph{European Conference on Computer Vision (ECCV)}, pp.\
  574--591, 2020.
\newblock \doi{10.1007/978-3-030-58580-8_34}.

\bibitem[Xue et~al.(2024)Xue, Yu, Zhang, Panagopoulou, Li,
  Mart{\'\i}n-Mart{\'\i}n, Wu, Xiong, Xu, Niebles, and Savarese]{xue2024ulip2}
Le~Xue, Ning Yu, Shu Zhang, Artemis Panagopoulou, Junnan Li, Roberto
  Mart{\'\i}n-Mart{\'\i}n, Jiajun Wu, Caiming Xiong, Ran Xu, Juan~Carlos
  Niebles, and Silvio Savarese.
\newblock {ULIP-2: Towards Scalable Multimodal Pre-training for 3D
  Understanding}.
\newblock In \emph{Proceedings of the IEEE/CVF Conference on Computer Vision
  and Pattern Recognition (CVPR)}, pp.\  27091--27101, 2024.

\bibitem[Yi et~al.(2016)Yi, Kim, Ceylan, Shen, Yan, Su, Lu, Huang, Sheffer, and
  Guibas]{yi2016scalable}
Li~Yi, Vladimir~G. Kim, Duygu Ceylan, I-Chao Shen, Mengyan Yan, Hao Su, Cewu
  Lu, Qixing Huang, Alla Sheffer, and Leonidas Guibas.
\newblock {A scalable active framework for region annotation in 3D shape
  collections}.
\newblock \emph{ACM Transactions on Graphics (TOG)}, 35\penalty0 (6):\penalty0
  1--12, 2016.
\newblock \doi{10.1145/2980179.2980238}.

\bibitem[Yin et~al.(2024)Yin, Yang, Liu, Wang, and Gao]{yin2024pdnet}
\RevisionRed{Xingyilang Yin, Xi~Yang, Liangchen Liu, Nannan Wang, and Xinbo Gao.
\newblock {Point Deformable Network with Enhanced Normal Embedding for Point
  Cloud Analysis}.
\newblock \emph{Proceedings of the AAAI Conference on Artificial Intelligence},
  38\penalty0 (7):\penalty0 6738--6746, March 2024.
\newblock \doi{10.1609/aaai.v38i7.28497}.}

\bibitem[Yu \& Song(2024)Yu and Song]{yu2024mmpoint}
Hai-Tao Yu and Mofei Song.
\newblock {MM-Point: Multi-View Information-Enhanced Multi-Modal
  Self-Supervised 3D Point Cloud Understanding}.
\newblock \emph{Proceedings of the AAAI Conference on Artificial Intelligence},
  38\penalty0 (7):\penalty0 6773--6781, 2024.
\newblock \doi{10.1609/aaai.v38i7.28501}.

\bibitem[Yu et~al.(2022)Yu, Tang, Rao, Huang, Zhou, and Lu]{yu2022pointbert}
Xumin Yu, Lulu Tang, Yongming Rao, Tiejun Huang, Jie Zhou, and Jiwen Lu.
\newblock {Point-BERT: Pre-Training 3D Point Cloud Transformers With Masked
  Point Modeling}.
\newblock In \emph{Proceedings of the IEEE/CVF Conference on Computer Vision
  and Pattern Recognition (CVPR)}, pp.\  19313--19322, 2022.

\bibitem[Zeng et~al.(2025)Zeng, Dong, Zhou, Qiu, Dong, Luo, and
  Li]{zeng2025deepla}
\RevisionRed{Ziyin Zeng, Mingyue Dong, Jian Zhou, Huan Qiu, Zhen Dong, Man Luo, and Bijun
  Li.
\newblock {DeepLA-Net: Very Deep Local Aggregation Networks for Point Cloud
  Analysis}.
\newblock In \emph{{Proceedings of the IEEE/CVF Conference on Computer Vision
  and Pattern Recognition (CVPR)}}, pp.\  1330--1341, June 2025.}

\bibitem[Zhang et~al.(2022)Zhang, Guo, Gao, Fang, Zhao, Wang, Qiao, and
  Li]{zhang2022pointm2ae}
Renrui Zhang, Ziyu Guo, Peng Gao, Rongyao Fang, Bin Zhao, Dong Wang, Yu~Qiao,
  and Hongsheng Li.
\newblock {Point-M2AE: Multi-scale Masked Autoencoders for Hierarchical Point
  Cloud Pre-training}.
\newblock In \emph{Advances in Neural Information Processing Systems},
  volume~35, pp.\  27061--27074. Curran Associates, Inc., 2022.
\newblock \doi{10.52202/068431-1962}.

\bibitem[Zhang et~al.(2023)Zhang, Wang, Qiao, Gao, and Li]{zhang2023i2p}
Renrui Zhang, Liuhui Wang, Yu~Qiao, Peng Gao, and Hongsheng Li.
\newblock {Learning 3D Representations From 2D Pre-Trained Models via
  Image-to-Point Masked Autoencoders}.
\newblock In \emph{Proceedings of the IEEE/CVF Conference on Computer Vision
  and Pattern Recognition (CVPR)}, pp.\  21769--21780, 2023.

\bibitem[Zhang et~al.(2024)Zhang, Zhang, and Yan]{zhang2024pcpmae}
Xiangdong Zhang, Shaofeng Zhang, and Junchi Yan.
\newblock {PCP-MAE: Learning to Predict Centers for Point Masked Autoencoders}.
\newblock In \emph{Advances in Neural Information Processing Systems},
  volume~37, pp.\  80303--80327. Curran Associates, Inc., 2024.
\newblock \doi{10.52202/079017-2553}.

\bibitem[Zhang et~al.(2025{\natexlab{a}})Zhang, Yuan, Qi, Zhang, Zhou, Ji, Yan, and
  Li]{zhang2025pcm}
\RevisionRed{Tao Zhang, Haobo Yuan, Lu~Qi, Jiangning Zhang, Qianyu Zhou, Shunping Ji,
  Shuicheng Yan, and Xiangtai Li.
\newblock {Point Cloud Mamba: Point Cloud Learning via State Space Model}.
\newblock \emph{Proceedings of the AAAI Conference on Artificial Intelligence},
  39\penalty0 (10):\penalty0 10121--10130, April 2025{\natexlab{a}}.
\newblock \doi{10.1609/aaai.v39i10.33098}.}

\bibitem[Zhang et~al.(2025{\natexlab{b}})Zhang, Zhang, and Yan]{zhang2025pointpqae}
Xiangdong Zhang, Shaofeng Zhang, and Junchi Yan.
\newblock {Towards More Diverse and Challenging Pre-training for Point Cloud
  Learning: Self-Supervised Cross Reconstruction with Decoupled Views}.
\newblock In \emph{Proceedings of the IEEE/CVF International Conference on
  Computer Vision (ICCV)}, pp.\  28696--28706, 2025{\natexlab{b}}.

\bibitem[Zheng et~al.(2024)Zheng, Huang, Mei, Hou, Lyu, Dai, Ouyang, and
  Gong]{zheng2024pointdif}
Xiao Zheng, Xiaoshui Huang, Guofeng Mei, Yuenan Hou, Zhaoyang Lyu, Bo~Dai,
  Wanli Ouyang, and Yongshun Gong.
\newblock {Point Cloud Pre-training with Diffusion Models}.
\newblock In \emph{Proceedings of the IEEE/CVF Conference on Computer Vision
  and Pattern Recognition (CVPR)}, pp.\  22935--22945, 2024.

\end{thebibliography}
\end{document}